\documentclass{article}
\addtocontents{toc}{\protect\setcounter{tocdepth}{-10}} 

\usepackage[main, final]{neurips_2026}

\usepackage[utf8]{inputenc} 
\usepackage[T1]{fontenc}    
\usepackage{hyperref}       
\usepackage{url}            
\usepackage{booktabs}       
\usepackage{amsfonts}       
\usepackage{amsmath}
\usepackage{nicefrac}       
\usepackage{microtype}      
\usepackage{xcolor}         
\usepackage{makecell}
\usepackage{graphicx}
\usepackage{diagbox}
\usepackage{subfigure}
\usepackage{rotating}
\usepackage[dvipsnames]{xcolor}
\usepackage{dsfont}
\usepackage{wrapfig}
\usepackage{longtable}
\usepackage{enumitem}
\usepackage{etoc}
\usepackage[normalem]{ulem}

\usepackage{tcolorbox,multicol}
\usepackage{multirow}

\usepackage{listings}

\lstdefinestyle{wrong}{
  backgroundcolor=\color{red!20},  
  frame=single,                       
}

\lstdefinestyle{correct}{
  backgroundcolor=\color{green!20},  
  frame=single,                       
}

\newcommand{\method}{\textit{MABGrader }}
\newcommand{\nospacemethod}{\textit{MABGrader}}

\newcommand{\synthmethod}{\textit{TROJAN }}
\newcommand{\nospacesynthmethod}{\textit{TROJAN}}

\newcommand{\humands}{\textit{Trojan-Bench(H) }}
\newcommand{\nospacehumands}{\textit{Trojan-Bench(H)}}

\newcommand{\synthds}{\textit{Trojan-Bench(S) }}
\newcommand{\nospacesynthds}{\textit{Trojan-Bench(S)}}

\title{Can LLMs Reliably Grade Olympiad Proofs? A Study of Mathematical Verification with LLMs}

\author{%
Azim Ospanov$^{1,2}$\thanks{First author: azimospanov@outlook.com}
\quad Zijin Feng$^{1}$\thanks{Corresponding authors: fengzijinn@gmail.com, farnia@cse.cuhk.edu.hk}
\quad Ding Ding$^{1,3}$ 
\quad Chengwu Liu$^1$ 
\quad Haoli Bai$^1$ \\
\quad \textbf{Jiacheng Sun}$^1$ 
\quad \textbf{Lifeng Shang$^{1}$} 
\quad \textbf{Farzan Farnia$^{2,\dagger}$} \\
    $^1$Huawei Foundation Model Department \\
    $^2$The Chinese University of Hong Kong \\
    $^3$The Hong Kong University of Science and Technology \\
}

\begin{document}

\maketitle

\vspace{-0.35cm}
\begin{abstract}
  Large Language Models (LLMs) have made remarkable progress on solving competition-level mathematics, yet their ability to verify natural-language mathematical proofs remains relatively underexplored. This gap is especially pronounced for Olympiad-level proofs, which require meticulous assessment of each claim in a long reasoning chain. Current proof verification thus primarily relies on expert inspection, which is costly to scale. In this paper, we present a systematic study of LLM-based verification in this setting. Specifically, we introduce a human-curated dataset of USAMO 2025 candidate solutions with expert grading, and conduct a broad evaluation of existing LLM verifiers across frontier models. We also propose \nospacesynthmethod, an iterative self-critique pipeline for generating high-fidelity adversarial proofs at scale. Our comprehensive evaluation covers 20 models spanning seven metrics, two prompt styles, three inference methods, and three prompt templates, showing that state-of-the-art models still leave considerable room for improvement. To narrow this gap, we propose \nospacemethod, a Multi-Armed Bandit framework that reframes grading as arm selection over the discrete score space. \method outperforms the SOTA ProofGrader by 26.4\% in Quadratic Weighted Kappa and reduces grading error by 14\%. Code and dataset are available at \url{https://github.com/aziksh-ospanov/TROJAN}.
\end{abstract}

\vspace{-0.15cm}
\section{Introduction}
\vspace{-0.1cm}
\begin{figure}[ht]
\vspace{-0.5cm}
    \centering
    \includegraphics[width=0.95\textwidth]{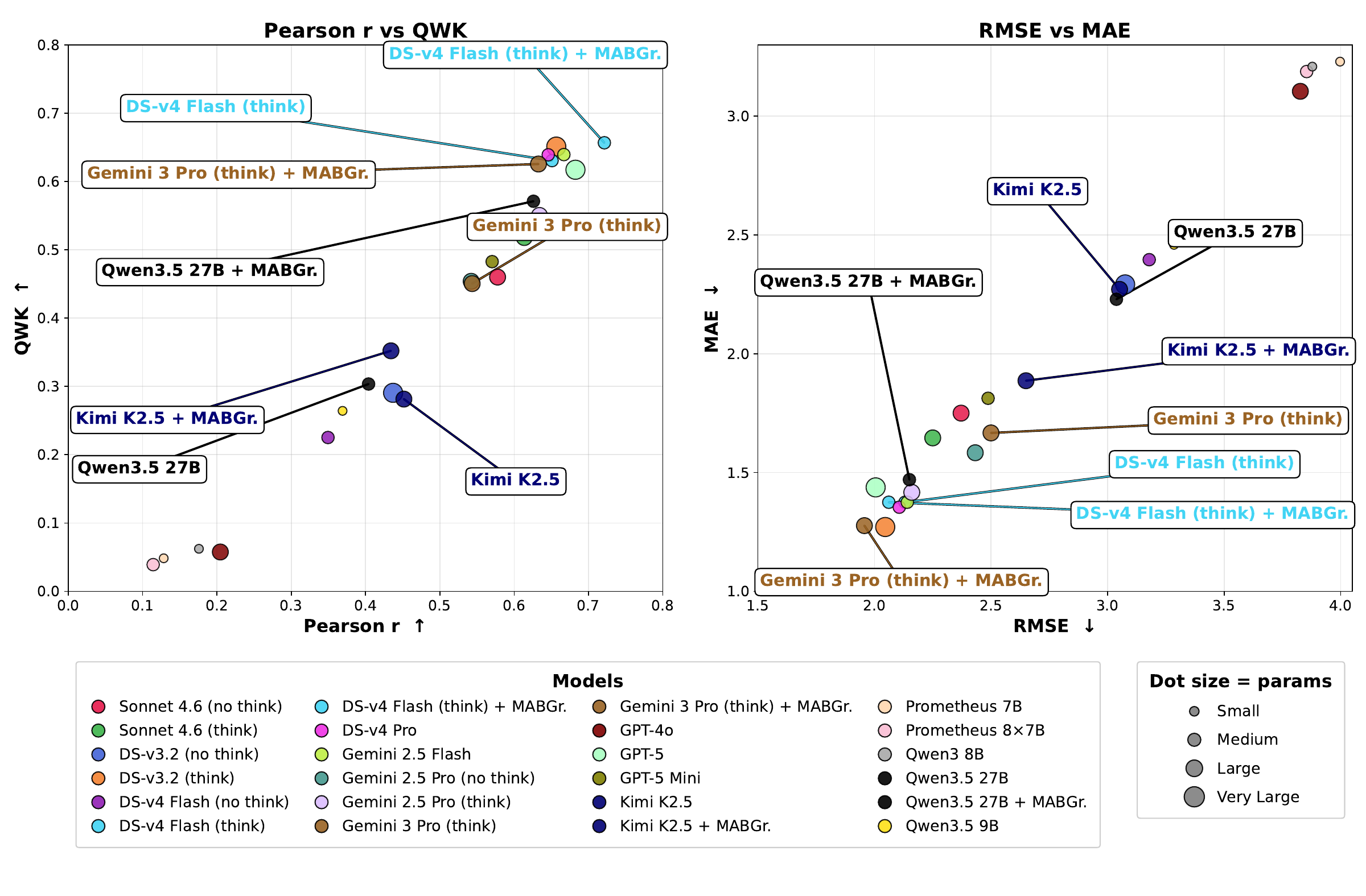}
    \vspace{-5mm}
    \caption{Performance of models on \humands, comparing Majority@16 with our \texttt{MABGr.} inference method across 4 metrics. Color and marker size denotes the model and parameter scale, resp. The spread indicates considerable room for improvement, while \texttt{MABGr.} show consistent gains.}
    \label{fig:scatter-plot-all-models}
    \vspace{-5mm}
\end{figure}

Mathematical problem-solving is a core capability of large language models (LLMs). Recent frontier models have demonstrated remarkable proficiency in competition-level mathematics, including the International Mathematical Olympiad (IMO) and the United States of America Mathematical Olympiad (USAMO)~\cite{wei2022chain, huang2025winning}. Much of this progress, however, is measured by final-answer correctness, which is a limited proxy for complex mathematical reasoning since a solution may reach the correct answer while containing logical gaps, unjustified claims, or subtle errors~\cite{DBLP:journals/corr/abs-2511-22570,DBLP:journals/corr/abs-2510-13888}. Moreover, many proving problems do not admit a final answer, requiring costly expert verification.

These motivate scalable methods for evaluating natural-language mathematical solutions.  In this work, we study the verification of natural-language mathematical solutions. Specifically, an LLM-based verifier receives a mathematical problem, a candidate solution, and optionally a reference solution and a marking scheme; it must then identify reasoning errors and assign a fair grade~\cite{DBLP:journals/corr/abs-2510-13888,DBLP:journals/corr/abs-2505-23281}.  Recent work has approached this task by constructing proof-verification benchmarks and designing LLM-based graders: MathArena~\cite{DBLP:journals/corr/abs-2505-23281} evaluates LLMs on competition proof problems using expert graders to assess model-generated solutions, ProofAutoGrader~\cite{DBLP:conf/emnlp/LuongHNGCSKBLMZHMKABSTL25} constructs IMO-GradingBench and studies automatic grading with reference solutions and guidelines, and ProofGrader~\cite{DBLP:journals/corr/abs-2510-13888} analyzes how model choice, grading schemes, reference solutions, and ensembling affect agreement with expert.

Despite this progress, two challenges remain for Olympiad-level mathematical verification. First, existing datasets often collect LLM-generated candidate solutions and rely on human experts to grade them, which is costly, difficult to scale, and provides limited control over the error distribution~\cite{DBLP:journals/corr/abs-2505-23281,DBLP:journals/corr/abs-2510-13888}. This makes it difficult to diagnose which types of mistakes LLM can or cannot detect. Second, verification itself remains difficult. Competition proofs are long, multi-step arguments in which a single mistake or omission can invalidate later conclusions or alter the appropriate partial credit. A verifier must thus go beyond surface plausibility by localizing the failure point and assessing its effect on the final grade. Figure~\ref{fig:scatter-plot-all-models} illustrates this gap. Existing approaches, including prompt engineering, PRM/ORM-based selection~\cite{lightman2024lets, DBLP:conf/acl/WangLSXDLCWS24}, and LLM-judge ensembles~\cite{chen2026harnessingmultiplelargelanguage, gu2025surveyllmasajudge}, can improve verification performance, but often require careful finetuning and stacking of different models. These limitations motivate diagnostic benchmarks for targeted error analysis and inference methods that improve grading without modifying the underlying base model.

In this work, we present a systematic study of LLM-based verification for Olympiad-level mathematics. We introduce a human-curated benchmark dataset derived from the USAMO 2025, where rubric-aligned adversarial errors, such as imprecise notation and unjustified lemma applications, are deliberately injected into candidate solutions. This design enables targeted evaluation of whether LLM judges can detect specific reasoning flaws and assign appropriate partial credit, while providing greater control over error type and severity than spontaneous LLM-generated mistakes. To scale data generation, we propose an iterative self-critique pipeline called \synthmethod (\textbf{\underline{T}}argeted adve\textbf{\underline{R}}sarial err\textbf{\underline{O}}r in\textbf{\underline{J}}ection for m\textbf{\underline{A}}th evaluatio\textbf{\underline{N}}) that produces adversarial solutions with human-like flaws. We show that fine-tuning on this synthetic data improves agreement with expert grading. We further propose \nospacemethod, an agentic Multi-Armed Bandit framework for mathematical verification. Rather than producing a verification judgment in one pass, \method formulates verification as an optimal selection problem over candidate judgments, where each arm corresponds to a possible grade with supporting error analysis. Through iterative sampling and reward optimization, it allocates more computation to uncertain cases and improves grade selection without modifying the underlying base model. Our contributions are summarized as follows:

\vspace{-0.2cm}
\begin{itemize}
    \item We introduce \nospacehumands, a human-curated dataset of error-injected USAMO 2025 solutions for evaluating how LLM verifiers detect subtle reasoning flaws and assign grades.
    \item We propose \nospacesynthmethod, a self-critique pipeline for generating high-fidelity adversarial samples, and show that training on the resulting \synthds improves alignment with experts.
    \item We conduct a systematic evaluation of 20 models across multiple prompts, inference methods, and metrics, measuring their verification abilities on Olympiad-level mathematics.
    \item We introduce \nospacemethod, a Multi-Armed Bandit algorithm that mitigates LLM verification instability and provides a robust alternative to single- and multi-shot grading methods.
\end{itemize}

\section{Related Works}
\vspace{-0.2cm}
\paragraph{Benchmarks for Mathematical Reasoning.} The mathematical evaluation includes foundational datasets such as GSM8K \cite{cobbe2021trainingverifierssolvemath} and MATH \cite{lightman2024lets}, alongside advanced benchmarks like AIME \cite{AIME2025}, HARDMath2 \cite{DBLP:journals/corr/abs-2505-11774}, and CollegeMath \cite{DBLP:conf/icml/TangZWW24}. Recent research has focused on competition-level mathematics through specialized benchmarks including Omni-MATH~\cite{gao_omni-math_2024}, Olympiad-Bench~\cite{he2024olympiadbenchchallengingbenchmarkpromoting}, and HARP~\cite{yue2024harpchallenginghumanannotatedmath}. Modern frameworks such as MathArena~\cite{DBLP:journals/corr/abs-2505-23281} provide model rankings based on challenging problems sampled from fresh Olympiads. Beyond benchmarking, recent breakthroughs have led to systems that participate directly in the International Mathematical Olympiad (IMO) \cite{huang2025winning, alphaproof, AlphaGeometryTrinh2024}. These advancements demonstrate that models are moving toward achieving performance comparable to top human competitors.

\vspace{-0.2cm}
\paragraph{LLM-as-a-Judge in mathematics.} LLM-as-a-Judge frameworks utilize large language models to evaluate responses based on metrics such as correctness and fairness \cite{gu2025surveyllmasajudge, zheng2023judging, li2025from, thakur2025judgingjudgesevaluatingalignment}. Recent studies on LLM reliability show that while these models achieve high correlation with human experts on standard tasks \cite{tan2025judgebench, frick2025how}, they remain susceptible to issues such as sycophancy \cite{petrov2025brokenmathbenchmarksycophancytheorem}. However, when grading complex mathematical theorems, many researchers continue to rely on human experts, a practice that limits the scalability of such evaluation pipelines \cite{DBLP:journals/corr/abs-2510-13888, DBLP:conf/emnlp/LuongHNGCSKBLMZHMKABSTL25, welleck2022naturalprover, petrov2025proof}.

\vspace{-0.2cm}
\paragraph{Verification in formal mathematics.} Complementing informal evaluation, formal mathematics utilizes proof assistants such as Lean~\cite{lean4paper}, Coq~\cite{bertot2013interactive}, and Isabelle~\cite{nipkow2002isabelle} to achieve automated verification. Benchmarks in this domain rely on the underlying language kernel to confirm proof validity, offering an objective alternative to natural language assessment~\cite{zheng2022minif2fcrosssystembenchmarkformal, ospanov2026minifflean, tsoukalas2024putnambenchevaluatingneuraltheoremprovers, yousefzadeh2025leandatasetinternationalmath}. Recent research also employs compiler feedback to guide LLM generation~\cite{ospanov2025apolloautomatedllmlean, DBLP:journals/corr/abs-2508-03613} and integrates informal reasoning steps into formal proofs to enhance verifiability~\cite{ospanov2025hermesefficientverifiablemathematical, jiang2023draft, cao2025revivingdspadvancedtheorem}.

\vspace{-0.2cm}
\paragraph{Industry standards on fair grading.} According to \cite{dekoninck2026the}, in math competitions with binary grading, the agreement is approximately 90.4\%. As we expand this to a 0–7 scale, the agreement and correlation will naturally be lower. Although those figures are not directly translatable to our setting, the consensus is that prestigious math Olympiads and new mathematical proofs should have a high rate of agreement among judges and reviewers. The intended gold standard involves multiple graders, a rubric, and a reconciliation process, which is stricter than single-rater "good enough" labeling. For general educational purposes, \cite{wood2021establishing} reports an absolute $QWK_{human, LLM} \geq 0.70$ as an acceptable metric for automated scoring. This is not a universal rule of thumb, as different domains have different requirements. We use these guidelines as a conservative lower bound for model evaluation. In this work, we systematically examine the current state of LLM grading to clarify the performance of automated judging against human labels.

\vspace{-0.2cm}
\section{Problem Formulation}\label{sec:preliminaries}

\vspace{-0.3cm}
\begin{figure}[t]
\centering

\begin{tcolorbox}[colback=gray!1!white,
    colframe=gray!75!black,title={}]

\textbf{Problem:} Let $k$ and $d$ be positive integers. Prove that there exists a positive integer $N$ such that for every odd integer $n>N$, the digits in the base-$2n$ representation of $n^k$ are all greater than $d$.

\vspace{0.5em}
\hrule

\vspace{-0.3cm}
\begin{multicols}{2}
\begin{center}
\tcbox[colback=ForestGreen,
      colframe=black,
      boxrule=0.5pt,
      arc=2pt,
    left=0.5pt,right=0.5pt,top=0.5pt,bottom=0.5pt]{%
  \textbf{Full Setting}%
}
\end{center}

\columnbreak

\begin{center}
\tcbox[colback=BrickRed,
      colframe=black,
      boxrule=0.5pt,
      arc=2pt,
      left=0.5pt,right=0.5pt,top=0.5pt,bottom=0.5pt]{%
  \textbf{Blind Setting}%
}
\end{center}
\end{multicols}

\begin{multicols}{2}
\textcolor{ForestGreen}{\textbf{Reference solution:}} We claim one such $N$ is $\boxed{N=2^{k-1}(d+1)}$  ... 

\columnbreak

\vspace*{\fill} 
\begin{center}
    \textcolor{BrickRed}{\texttt{NO REFERENCE SOLUTION}}
\end{center}
\vspace*{\fill} 
\end{multicols}

\begin{multicols}{2}
\textcolor{ForestGreen}{\textbf{Marking Scheme:} }

\textit{Initial closed form expression for each digit [1 pt]}: Uses the base-2n representation...




\columnbreak

\vspace*{\fill} 
\begin{center}
    \textcolor{BrickRed}{\texttt{NO MARKING SCHEME}}
\end{center}
\vspace*{\fill} 
\end{multicols}

\begin{multicols}{2}
\textcolor{ForestGreen}{\textbf{Candidate Solution:}} Begin by letting $a_i$ be the digit at the $i$th position from the...

\columnbreak

\textcolor{BrickRed}{\textbf{Candidate Solution:}} Begin by letting $a_i$ be the digit at the $i$th position from the...
\end{multicols}


\end{tcolorbox}

\vspace{-3mm}
\caption{Example verification inputs for USAMO 2025 Problem 1 under the Full and Blind settings. In the Full setting, the verifier receives the problem, reference solution, marking scheme, and candidate solution. In the Blind setting, the verifier receives only the problem and candidate solution.}
\label{fig:usamo_example}
\vspace{-3mm}
\end{figure}


In this work,  we study the verification of natural-language mathematical solutions. Given a mathematical problem $p$, a candidate solution $\tilde{s}$, and optionally a reference solution $\texttt{Ref}$ and a marking scheme $\texttt{MS}$, the goal is for an LLM-based verifier to identify reasoning errors and assign a grade that fairly reflects the correctness and completeness of the solution. Following common practice in ~\cite{DBLP:journals/corr/abs-2505-23281,DBLP:conf/emnlp/LuongHNGCSKBLMZHMKABSTL25,DBLP:journals/corr/abs-2510-13888}, grades range from 0 to 7.

We consider two verification settings, illustrated in Figure~\ref{fig:usamo_example}. In the Full setting, corresponding to the $\texttt{Ref+MS}$ configuration of \cite{DBLP:journals/corr/abs-2510-13888}, the verifier is provided with both the reference solution and the marking scheme specifying point allocations and deductions. In the Blind setting, corresponding to the $\texttt{None}$ configuration of \cite{DBLP:journals/corr/abs-2510-13888}, the verifier receives only the problem and candidate solution, without access to either a reference solution or a marking scheme.

\vspace{-0.2cm}
\section{Dataset Construction}\label{sec:ds construction}
\vspace{-0.1cm}
To comprehensively evaluate whether LLMs can verify Olympiad-level proofs and assign appropriate grades, in this section we describe the construction of our human-curated \humands benchmark and the proposed data generation pipeline \nospacesynthmethod, which we use to obtain dataset \synthds.

\vspace{-0.1cm}
\subsection{Human curated dataset}
\vspace{-0.1cm}

We construct \humands to support controlled evaluation of LLM-based proof verification. Unlike existing datasets that collect LLM-generated solutions and annotate them with human grades, we use a backward construction procedure: starting from golden reference solutions, we map proof components to marking-scheme criteria and systematically inject errors that correspond to targeted point deductions. We build our dataset \humands from the USAMO 2025 competition set, which contains six proof-based problems across multiple areas of mathematics. Grades range from 0 to 7 according to the marking schemes. We use the USAMO 2025 problems, reference solutions, and marking schemes provided by MathArena~\cite{DBLP:journals/corr/abs-2505-23281}.

Concretely, we construct candidate solutions by injecting subtle errors into the golden solution, disrupting the proof structure and leading to specific point deductions. For each problem, we randomly sample a target grade and introduce errors designed to match the corresponding marking-scheme deductions.  Furthermore, when constructing faulty solutions, we applied heavier penalties to fundamental parts of the proof, such as operating in base-$2n$ in Problem 1 or establishing the correct definition and structure of cupcake segments in Problem 6. A logical gap in a fundamental step typically causes a cascade of point losses across other criteria. The 2026 version of the USAMO in \cite{DBLP:journals/corr/abs-2505-23281} follows a similar logic by specifying overarching deductions.

To account for the inherent variance in individual grading, we utilized two independent annotators for the final dataset. We followed the data creation pipeline from \cite{rein2024gpqa} to increase the validity and reliability of the final set. However, we omitted the final step of non-expert validation due to the specialized nature of mathematical proofs. The result is a set of 48 human-verified USAMO 2025 candidate solutions with a corresponding human grade and a justification. We provide a detailed description of the creation pipeline and aggregated metrics in the following sections, and a full datasheet can be found in the Appendix Section~\ref{app:datasheet}.

\begin{table}[ht]
\centering
\vspace{-0.1cm}
\caption{Inter-annotator agreement on our human-curated benchmark \humands}
\label{tab:human-ds-info}
\begin{tabular}{ccccccccc}
\toprule
Metric & QWK & Pearson $r$ & RMSE & MAE & WTA$_{\leq 1}$ & WTA$_{\leq 2}$ & Exact Match & Bias\\
\midrule
Value & 0.76 & 0.89 & 1.09 & 0.62 & 0.85 & 0.94 & 0.58 & $\pm 0.08$\\
\bottomrule
\end{tabular}
\vspace{-0.2cm}
\end{table}

\paragraph{Expert Inter-Rater Agreement. } Table~\ref{tab:human-ds-info} provides the agreement metrics for the human-curated dataset following the procedure in \cite{rein2024gpqa}. The results show a Quadratic Weighted Kappa (QWK) of 0.76 and a Pearson correlation of 0.89, which exceed the performance of any of the evaluated LLMs and indicate strong expert consensus. Scoring consistency is further reflected in the Within Tolerance Agreement, where 85\% of scores fall within 1 point ($WTA_{\leq 1}$) and 94\% fall within 2 points ($WTA_{\leq 2}$). Combined with a low systematic bias of $\pm 0.08$, these metrics establish the human-curated set as a reliable gold standard for assessing automated grading systems.

\vspace{-0.1cm}
\subsection{Synthetic dataset generation}\label{sec:synthetic data generation}

\begin{figure}[t]
    \centering
    \includegraphics[width=\linewidth]{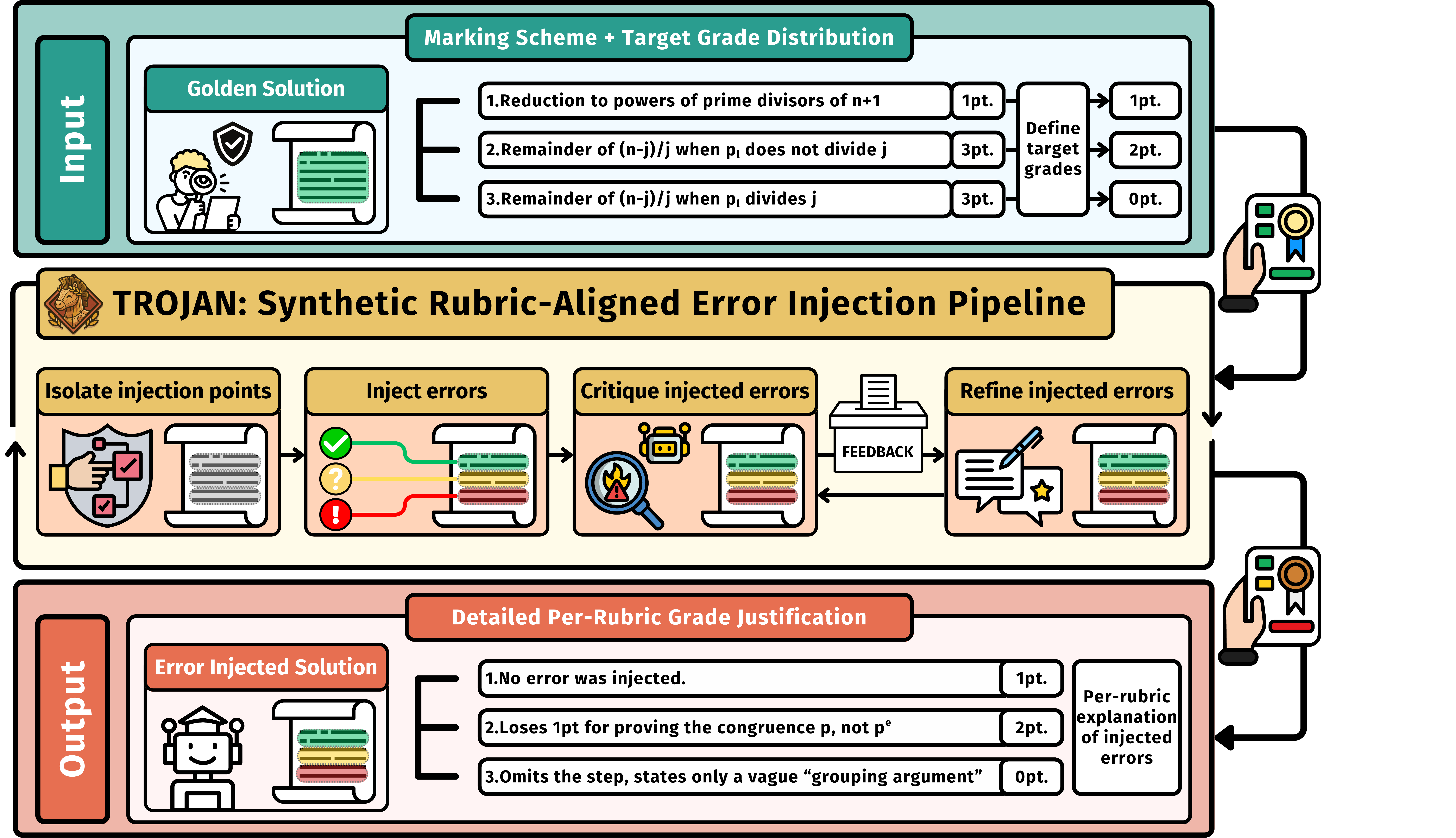}
    \caption{Overview of our synthetic data generation pipeline \nospacesynthmethod. Given a problem, reference solution, marking scheme, and target score profile, the pipeline injects rubric-aligned errors, performs iterative self-critique and refinement, and outputs an adversarial proof with a grade justification.}
    \label{fig:synth-pipeline}
    \vspace{-5mm}
\end{figure}

To scale benchmark construction beyond manual error injection, we introduce \nospacesynthmethod, an LLM-based pipeline for generating adversarial mathematical proofs. Given a mathematical problem, a reference golden solution, a marking scheme, and a target score profile, \nospacesynthmethod \ outputs a proof whose errors are calibrated to be consistent with the specified score profile, together with a structured grade justification. Figure~\ref{fig:synth-pipeline} provides an overview of the pipeline. 

The score profile specifies the target point allocation across marking-scheme criteria. The generator LLM is given one human-authored error-injection example for one-shot prompting and then analyzes the marking scheme. For criteria assigned full marks, it is allowed to closely paraphrase the corresponding part of the reference solution; for each criterion with points losses, it is tasked with injecting errors calibrated to the target score. The same LLM then performs self-critique using a separate prompt. The goal is to isolate the parts that contain mistakes and produce an explicit self-judgment. Its output is an alignment judgment together with revision guidance. In the refinement stage, the model applies the suggested revisions to the corresponding proof blocks. This critique-refinement loop terminates once the self-critique model accepts the current candidate $n_{\mathrm{alignment}}$ consecutive rounds; if alignment is broken before then, the streak is discarded and refinement continues with the latest critique. The maximum number of critique-refinement rounds is bounded by $n_{\mathrm{iter}}$.

The pipeline is general and can be applied to any problem set with valid reference solutions and marking schemes. We construct a large synthetic dataset \synthds from four competitions: USAMO 2025, USAMO 2026, IMO 2025, and IMC 2025, with problem statements, reference solutions, with data sourced from the MathArena repository~\cite{DBLP:journals/corr/abs-2505-23281}. When evaluating the synthetic data, we treat this grade and justification as ground-truth labels. Experiments on the subset of \synthds show that it closely correlates with \nospacehumands, suggesting good alignment.

\vspace{-0.2cm}

\begin{table}[ht]
\centering
\caption{Correlation and agreement of the synthetically-generated benchmark \synthds with expert graders.}
\label{tab:synth-ds-human-audit}
\begin{tabular}{ccccccccc}
\toprule
Metric & QWK & Pearson $r$ & RMSE & MAE & WTA$_{\leq 1}$ & WTA$_{\leq 2}$ & Exact Match & Bias\\
\midrule
Value & 0.78 & 0.87 & 1.20 & 0.77 & 0.83 & 0.90 & 0.67 & 0.11\\
\bottomrule
\end{tabular}
\vspace{-0.2cm}
\end{table}

\begin{table}[ht]
\centering
\caption{Inter-annotator agreement on the synthetically-generated benchmark \nospacesynthds.}
\label{tab:synth-ds-iaa}
\begin{tabular}{ccccccccc}
\toprule
Metric & QWK & Pearson $r$ & RMSE & MAE & WTA$_{\leq 1}$ & WTA$_{\leq 2}$ & Exact Match & Bias\\
\midrule
Value & 0.78 & 0.94 & 0.70 & 0.49 & 1.00 & 1.00 & 0.51 & $\pm 0.00$\\
\bottomrule
\end{tabular}
\vspace{-0.2cm}
\end{table}

\paragraph{Audit of \nospacesynthds. } To further validate the quality of the proposed \nospacesynthmethod, we performed a manual audit of the \synthds subset. We sampled 50 problems and annotated using similar procedures as in the construction and audit of \nospacehumands. Each problem was verified by two experts and we present the correlation, error and bias metrics between LLM and human assigned grades in Table~\ref{tab:synth-ds-human-audit}. We observe that generally the correlation is high and $\sim83\%$ of problems fall within one grading points. This generally indicates that the proposed rubric-based error injection helps LLMs inject relevant and appropriate synthetic errors into the correct solutions. We also report the inter-annotation agreement in Table~\ref{tab:synth-ds-iaa} between the experts in this audit and show that the consensus was high when grading, with all graded problems falling into $WTA_{\leq 1}$. Combined, this shows robustness and overall high quality of synthetically injected errors via \nospacesynthmethod.

\section{Our Approach}\label{sec:approach}
To address the difficulty of assigning fair grades to Olympiad-level problems, we implemented a Multi-Armed Bandit (MAB) algorithm for grade selection. Even with access to a mark scheme and a reference solution, producing a grade outright is often hard. Prior work also suggests that evaluating a pre-assigned grade together with its justification is easier than scoring from scratch, since the candidate grade provides a concrete anchor for assessment.~\cite{cobbe2021trainingverifierssolvemath, lightman2024lets}. Independently, evidence from the LLM-as-a-judge and educational assessment literatures shows that aggregating multiple independent evaluators of the same candidate reduces individual-judge bias and improves agreement with expert ratings~\citep{verga2024replacingjudgesjuriesevaluating, zheng2023judging, thakur2025judgingjudgesevaluatingalignment, e98ec698-e30f-319f-8f04-9cae556cb352}.

Motivated by this, we first considered an A/B-style voting scheme that, given a grade range, argues for one grade against another. This approach is token-hungry: its token cost grows linearly with the number of comparisons, and it wastes pulls on clearly sub-optimal arms whose exploration cost is high. We therefore adopt UCB~\cite{LAI19854, Auer2002Finitetime}, a well-studied bandit algorithm which, to our knowledge, has not previously been applied to mathematical grading. We compare them in the Appendix section~\ref{app:effect-of-mab}.

At the core of our algorithm is a judgment-defense setup. For each candidate grade $g \in \{0, \dots, 7\}$, we prompt an LLM to justify the grade and record its justification along with a self-reported confidence. In the next phase, the same model, now prompted as a judging oracle, selects arms and assigns rewards. This recovers the classic stochastic bandit setting. After each round, \method aggregates the oracle's rewards and updates the arm statistics, balancing exploration and exploitation to converge on the grade closest to the underlying true grade.

At round $t$, let $N_g(t)$ denote the number of times arm $g$ has been pulled prior to round $t$, let $r_s \in [0,1]$ denote the reward observed at pull $s$, and let
$$
\hat{\mu}_g(t) \;=\; \frac{1}{N_g(t)} \sum_{s=1}^{t-1} r_s \,\mathds{1}\!\left[g_s = g\right],
\qquad
N_g(t) \;=\; \sum_{s=1}^{t-1} \mathds{1}\!\left[g_s = g\right]
$$
denote the empirical mean and pull count of arm $g$. UCB first plays each arm once, then at every subsequent round selects
$$
g_t \;=\; \arg\max_{g \in \{0,\dots,7\}} \Big[\, \underbrace{\hat{\mu}_g(t)}_{\text{exploitation}} \;+\; \underbrace{c\sqrt{\tfrac{2\ln t}{N_g(t)}}}_{\text{exploration}} \,\Big],
$$
breaking ties uniformly at random. The default $c=1$ recovers the classical UCB1 bound of \cite{Auer2002Finitetime}, which yields $\mathcal{O}(\sqrt{KT\ln T})$ regret for $K=8$ arms with rewards in $[0,1]$; the oracle's 1-5 fairness Likert is rescaled to $[0,1]$ via $r = (\tilde r - 1)/4$. The parameter $c$ is user-controlled and can be adapted to different problems. The design is compatible with any single-arm sampling bandit, including Exp3, $\epsilon$-greedy, and Thompson Sampling. For this study we chose a single, well-established algorithm and leave a comparison with alternatives to future work. We provide more details in the Appendix Section~\ref{app:mab-ablation}.

\section{Experiments}\vspace{-2mm}

\subsection{Experimental Setup}

\vspace{-0.1cm}
\paragraph{Inference Settings.} In all experiments, we fix the sampling parameters at $top_p=0.95$ and temperature $t=0.1$ to minimize LLM-induced variance during the grading process. Unless otherwise noted, we employ the "thinking" versions of the evaluated models. For models that support adjustable reasoning effort, the setting was fixed to "high". Maximum token limits vary by model and are detailed in the Appendix. The default testing environment follows the MathArena template~\cite{DBLP:journals/corr/abs-2505-23281}. Every run was initialized with a seed of 42. We explicitly indicate where an alternative template was used for comparison. 

All multi-sample experiments are performed at @16, where we sample 16 independent attempts from scratch. To evaluate Best-of-N methods, we utilize two Outcome Reward Models (ORMs): Skywork-Reward-Llama-3.1-8Bv0.2~\cite{DBLP:journals/corr/abs-2410-18451} and ArmoRM-Llama3-8Bv0.1~\cite{DBLP:conf/emnlp/00030X0024}. We also incorporate two Process Reward Models (PRMs): math-shepherd-mistral-7b-prm~\cite{DBLP:conf/acl/WangLSXDLCWS24} and RLHFlow/Llama3.1-8B-PRMDeepseek-Data~\cite{dong2024rlhf}. For our main \method results we used 16 number of iterations and $c=0.25$ guided by our ablation result. 

\vspace{-0.1cm}
\paragraph{Evaluation Metrics.} We use seven metrics to measure alignment between verifier-generated grades and ground truth grades.
{\footnotesize
\[
\hspace{-0.1cm}
\begin{aligned}
\mathrm{QWK} &= 1-\frac{\sum_{i,j}w_{i,j}O_{i,j}}{\sum_{i,j}w_{i,j}E_{i,j}},
&
\mathrm{WTA}_{\tau} &= \frac{1}{n}\sum_{i=1}^{n}\mathbf{1}\{|y_i-\hat{y}_i|\le \tau\},
&
\mathrm{RMSE} &= \sqrt{\frac{1}{n}\sum_{i=1}^{n}(y_i-\hat{y}_i)^2},
\\
\mathrm{MAE} &= \frac{1}{n}\sum_{i=1}^{n}|y_i-\hat{y}_i|,
&
\mathrm{Pearson}_r &= \frac{\sum_i(y_i-\bar{y})(\hat{y}_i-\bar{\hat{y}})}
{\sqrt{\sum_i(y_i-\bar{y})^2}\sqrt{\sum_i(\hat{y}_i-\bar{\hat{y}})^2}},
&
\mathrm{EM} &= \frac{1}{n}\sum_{i=1}^{n}\mathbf{1}\{y_i=\hat{y}_i\}.
\end{aligned}
\]
}
Here, $n$ is the number of grading samples; $y_i$ and $\hat{y}_i$ the ground-truth and verifier-generated grades for the $i$-th sample, respectively; and $\bar{y}$ and $\bar{\hat{y}}$ their corresponding means. QWK measures ordinal agreement while penalizing larger grade discrepancies, with $O$ and $E$ denoting the observed and expected agreement matrices, respectively, and weights $w_{i,j}=(i-j)^2/(N-1)^2$ over the $N=8$ possible grades. RMSE measures overall grading error, WTA$_\tau$ measures the fraction of predictions within tolerance $\tau$, Pearson $r$ measures linear correlation, MAE measures average absolute grading error, and EM measures exact grade agreement. We also report Bias, defined as $\frac{1}{n}\sum_{i=1}^{n}(\hat{y}_i-y_i)$, with positive and negative values indicating over- and under-grading, respectively.

\vspace{-0.1cm}
\paragraph{Data Generation.} For synthetic data generation, we utilized Claude Opus 4.6 (Think)~\cite{anthropic2026claude46}. Because existing literature indicates a high self-validation bias~\cite{chen2025llmevaluatorspreferreason}, we exclude this model from our evaluation suite. We use a temperature of $t=0.7$ for data generation and refinement, while the critic module employs a more conservative $t=0.3$ to reduce variance in the feedback. We also finetuned two models, Qwen3.5-9B and 27B SFT, on \nospacesynthds. The model cards are specified in Appendix.

\vspace{-2mm}
\subsection{Analysis of \synthds}

\begin{figure}
    \centering
    \includegraphics[width=\linewidth]{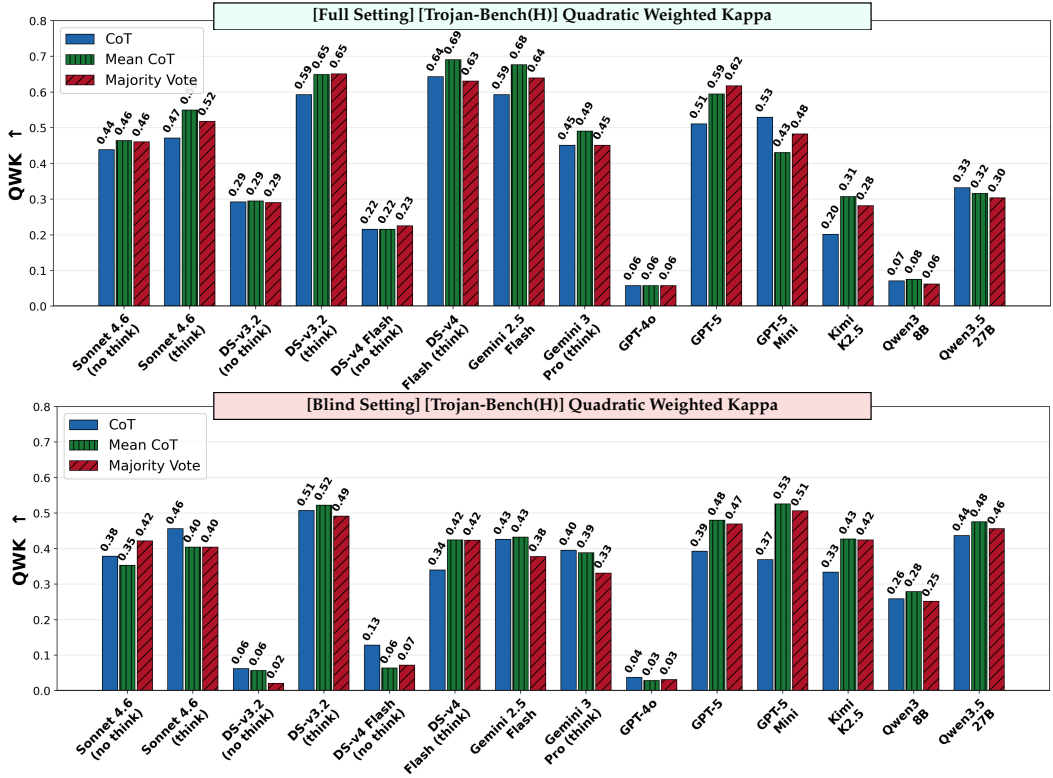}
    \caption{Model correlation with human labels using QWK across Full and Blind settings. The bar plots report CoT@1, MeanCoT@16, and MajorityVote@16 for each tested model.}
    \label{fig:qwk-barplots-main-text}
    \vspace{-2mm}
\end{figure}

\paragraph{LLM performance on the human-annotated benchmark.} We present the performance of 14 models on the human-annotated benchmark using the QWK metric in Figure~\ref{fig:qwk-barplots-main-text}. In this section, we focus primarily on the correlation metric and provide a detailed breakdown of all other metrics in the Appendix Sections~\ref{app:extended-eval}, \ref{app:barplots} and \ref{app:heatmaps}. Our evaluation spans two scenarios, Blind and Full, and incorporates three sampling methods: CoT@1 (a single inference), MeanCoT@16 (the averaged grade rounded to the nearest integer), and MajorityVote@16 (selecting the grade with the highest frequency).

Consistent with the findings in \cite{DBLP:journals/corr/abs-2510-13888}, the Full setting is, on average, a less challenging task than the Blind setting. We observe that models achieve a 0.06 higher absolute correlation and a 24.2\% relative improvement in performance when using MajorityVote@16. Interestingly, smaller models such as GPT-5-mini or Qwen3-8B achieve higher correlation when not provided with additional information. Upon further inspection, we found that these models are often unable to follow the gold solution and instead produce hallucinated responses. A deeper case analysis is provided in the Appendix Section~\ref{app:case_blind_vs_full}. Overall, the models remain below the 0.70 threshold even when supplied with full grading information and instructions, whereas expert annotators achieve a 0.76 agreement rate. Furthermore, LLM performance collapses to a maximum of 0.53 in the Blind setting. This scenario is particularly difficult as the models must both derive a correct solution and grade the attempt based on that derivation. We observe similar trends in RMSE, Pearson correlation, and Bias, with the full suite of results detailed in the Appendix Sections~\ref{app:extended-eval}, \ref{app:barplots}, and \ref{app:heatmaps}.

Additionally, we conducted a comparative analysis of the different sampling methods. Because we utilized a lower temperature, the performance gap between single and multi-sampling is relatively small. However, consistent trends remain present in our evaluation. On average, MajorityVote outperforms other sampling techniques, though the primary benefit serves to stabilize the grading output. Specifically, under Full inference, MajorityVote scores 5.46\% higher than CoT and 2.36\% higher than MeanCoT. 

\begin{figure}
    \centering
    \includegraphics[width=\linewidth]{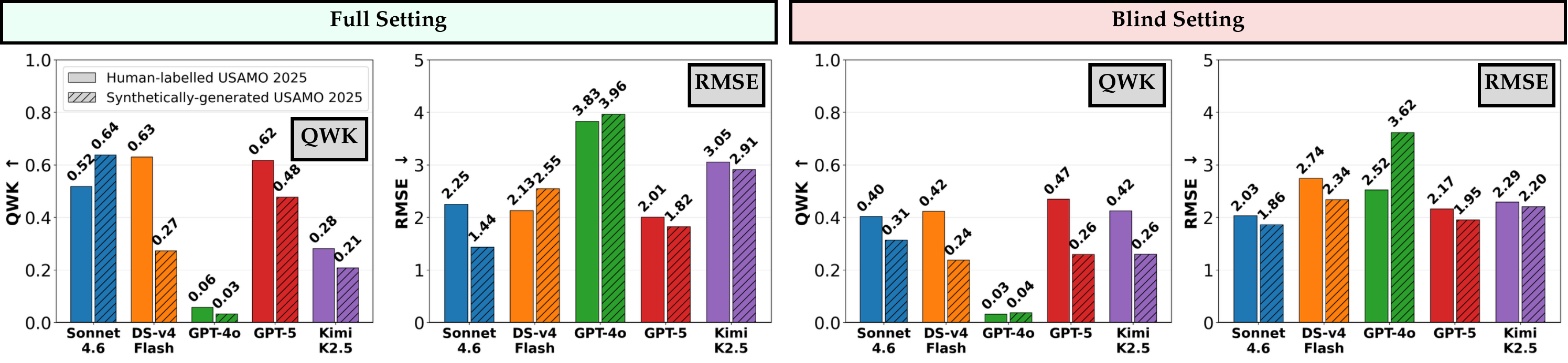}
    \caption{Grading performance on human and synthetic USAMO 2025 subsets ($N=48$). Reported values represent the outcome of MajorityVote@16 across QWK, RMSE, and Bias metrics.}
    \label{fig:human-synth-alignment-usamo-2025}
\end{figure}

\begin{table*}[t!]
    \centering
    \caption{Comparative performance of base and fine-tuned models on \synthmethod synthetic samples.}
    \label{tab:sft-experiment-main}
\begin{footnotesize}
\begin{tabular}{lc*{7}{c}}
  \toprule
  & \multicolumn{3}{c}{\textbf{Full Setting}} & \multicolumn{3}{c}{\textbf{Blind Setting}} \\
  \cmidrule(lr){2-4}\cmidrule(lr){5-7}
  & QWK ($\uparrow$) & RMSE ($\downarrow$) & Bias ($\approx 0$) & QWK ($\uparrow$) & RMSE ($\downarrow$) & Bias ($\approx 0$)\\
  \midrule
  Qwen3.5-9B (Base) & 0.02 & 4.00 & 3.23 & 0.06 & 2.52 & 0.63 \\
  Qwen3.5-9B (SFT) & 
  \textbf{0.35} & \textbf{2.08} & \textbf{0.27} & \textbf{0.07} & \textbf{2.40} & \textbf{0.39} \\
  \midrule
  Qwen3.5-27B (Base) & 0.36 & 2.98 & 1.54 & -0.11 & 4.12 & -3.40 \\
  Qwen3.5-27B (SFT) & \textbf{0.47} & \textbf{2.10} & \textbf{0.77} & \textbf{0.06} & \textbf{2.50} & \textbf{0.21} \\
  \bottomrule
\end{tabular}
\end{footnotesize}
\vspace{-1mm}
\end{table*}


\begin{table}[ht]
\centering
\caption{Sensitivity of QWK, RMSE, and Bias to $n_{\text{iter}}$ and $n_{\text{align}}$.}
\label{tab:synth-generation-ablation}
\vspace{-1mm}
\begin{footnotesize}
\setlength{\tabcolsep}{4pt}
\renewcommand{\arraystretch}{1.05}
\begin{tabular}{c@{\hskip 4pt}c ccc ccc ccc}
\toprule
& & \multicolumn{3}{c}{QWK $\uparrow$} & \multicolumn{3}{c}{RMSE $\downarrow$} & \multicolumn{3}{c}{Bias} \\
\cmidrule(lr){3-5}\cmidrule(lr){6-8}\cmidrule(lr){9-11}
& $n_{\text{align}}\!\rightarrow$ & 1 & 2 & 3 & 1 & 2 & 3 & 1 & 2 & 3 \\
\midrule
\multirow{5}{*}{$n_{\text{iter}}$}
 & 1 & 0.583 & —     & —     & 1.414 & —     & —     &  0.400 & —     & —     \\
 & 3 & 0.579 & 0.833 & 0.776 & 1.265 & 0.894 & 1.049 &  0.400 & 0.000 & 0.100 \\
 & 5 & 0.757 & \textbf{0.884} & 0.818 & 0.949 & \textbf{0.707} & 0.894 & 0.300 & 0.100 & 0.000 \\
 & 7 & 0.694 & 0.683 & 0.717 & 1.225 & 1.140 & 1.342 &  0.700 & 0.500 & 0.600 \\
 & 9 & 0.571 & 0.735 & 0.622 & 1.225 & 1.140 & 1.304 & -0.300 & 0.100 & 0.700 \\
\bottomrule
\end{tabular}
\end{footnotesize}
\vspace{-2mm}
\end{table}

\paragraph{Quality of the Synthetic Dataset.} Furthermore, to evaluate the quality of the synthetic dataset, we provide two points of analysis. First, we measured the QWK, RMSE, and Bias on the USAMO 2025 subset of the synthetic data. We kept all sampling parameters consistent across both sets, with each containing 48 samples. We present these results in Figure~\ref{fig:human-synth-alignment-usamo-2025}. Our evaluation suggests that on the human-curated benchmark, models achieve a 0.07 lower absolute QWK, a 0.12 higher absolute RMSE, and a 0.04 higher absolute Bias compared to their synthetic counterparts. These results indicate that the human-crafted dataset presents a higher level of difficulty. While most models exhibit similar trends, we observe that DeepSeek-V4-Flash-Thinking shows a significantly lower correlation with synthetic data compared to the human-annotated version. When measuring the correlation of model responses between the expert-labeled and synthetic sets, we observe a Pearson $r$ of 0.89. 

\paragraph{Fine-Tuning on \nospacesynthds.} Additionally, we performed SFT on the synthetic data and present our findings in Table~\ref{tab:sft-experiment-main}. We supervised fine-tuned Qwen3.5-9B and 27B models~\cite{qwen3.5} using LLaMA-Factory~\cite{zheng2024llamafactory} on the full set of our synthetic data derived from four competition problems (see Section~\ref{sec:synthetic data generation}). The training mixture includes data from both the full and blind verification settings. The data were split into training and validation sets using a 95/5 ratio. The maximum context length was set to 16K tokens, which is sufficient for our task. The 9B model was fully fine-tuned, while the 27B model was fine-tuned with LoRA using rank-8 adapters. Both models were trained for 2 epochs with an effective global batch size of 16, bfloat16 precision, and a cosine learning-rate schedule with 10\% warmup. The peak learning rate was 1e-5 for the full fine-tuning run and 1e-4 for the LoRA run. All participating models were run in "non-thinking" mode, which excludes the reasoning phase typically enclosed in \texttt{<think>} tokens.

We observe that after fine-tuning, smaller models demonstrate improved performance across the three primary metrics. Specifically, when trained on synthetic data, models achieve a higher correlation with human grading, a result that is consistent across both small and large model scales. Taken together, these findings suggest that synthetic data provides a high degree of correlation with human-generated labels and that training on this data effectively improves agreement with expert judgment.

\paragraph{\synthmethod hyperparameter selection.} To choose parameters for our synthetic pipeline generation, we ablated two main parameters. For each pair of parameters we asked human experts to grade 10 outputs and aggregated LLM-assigned score (according to a score profile) and human assigned scores. We report results in Table~\ref{tab:synth-generation-ablation}. We selected $n_{iter}=5$ and $n_{alig.}=2$ for the \synthmethod pipeline as this configuration represents the global optimum across all primary evaluation metrics. This setting achieves the highest correlation with human labels (QWK = 0.884) and the lowest error (RMSE = 0.707). While certain alternative configurations provide a zero bias, they do so at the expense of significantly lower correlation and higher error rates. 

\begin{table*}[t!]
    \centering
    \caption{Performance on the manually annotated USAMO 2025 dataset in the Full setting, where models are given a reference solution and marking scheme. We report QWK, RMSE, and Bias for four base models: DeepSeek-V4-Flash, Kimi-K2.5, Gemini-3-Pro-Preview, and Qwen3.5-27B.}
    \label{tab:mab-usamo-human-full}
\begin{footnotesize}
\begin{tabular}{lc*{11}{c}}
  \toprule
  & \multicolumn{4}{c}{\textbf{QWK ($\uparrow$)}} & \multicolumn{4}{c}{\textbf{RMSE ($\downarrow$)}} & \multicolumn{4}{c}{\textbf{Bias ($\approx 0$)}} \\
  \cmidrule(lr){2-5}\cmidrule(lr){6-9}\cmidrule(lr){10-13}
  & \makecell{DS\\V4F} & \makecell{Kimi \\ K2.5} & \makecell{Gem.\\3P} & \makecell{Q3.5\\27B} & \makecell{DS-\\V4F} & \makecell{Kimi \\ K2.5} & \makecell{Gem.\\3P} & \makecell{Q3.5\\27B} & \makecell{DS-\\V4F} & \makecell{Kimi \\ K2.5} & \makecell{Gem.\\3P} & \makecell{Q3.5\\27B} \\
  \midrule 
  \multicolumn{12}{l}{\textit{Chain-of-Thought (CoT@1)}} \\
  \midrule 
  MathArena~\cite{DBLP:journals/corr/abs-2505-23281} & 0.59 & 0.20 & 0.45 & 0.33 & 2.28 & 3.29 & 2.50 & 3.37 & -0.69 & 2.24 & 1.35 & 1.73 \\
  ProofGr.~\cite{DBLP:journals/corr/abs-2510-13888} & 0.64 & 0.24 & 0.46 & 0.36 & 2.15 & 3.06 & 2.51 & 3.10 & -0.81 & 2.14 & 1.59 & 2.02 \\
  ProofAGr.~\cite{DBLP:conf/emnlp/LuongHNGCSKBLMZHMKABSTL25} & 0.52 & 0.19 & 0.42 & 0.32 & 2.47 & 3.47 & 2.84 & 3.60 & -1.15 & 2.86 & 1.78 & 1.97 \\
  \midrule 
  \multicolumn{12}{l}{\textit{Chain-of-Thought with Majority Vote Selection (MajorityVote@16)}} \\
  \midrule 
  MathArena~\cite{DBLP:journals/corr/abs-2505-23281} & 0.64 & 0.28 & 0.45 & 0.30 & 2.13 & 3.05 & 2.50 & 3.04 & -0.62 & 2.15 & 1.38 & 1.81 \\
  ProofGr.~\cite{DBLP:journals/corr/abs-2510-13888} & \textbf{\underline{0.66}} & 0.27 & 0.46 & 0.41 & 2.14 & 2.84 & 2.41 & 2.99 & -0.71 & 2.21 & 1.33 & 1.89 \\
  ProofAGr.~\cite{DBLP:conf/emnlp/LuongHNGCSKBLMZHMKABSTL25} & 0.55 & 0.26 & 0.44 & 0.38 & 2.40 & 3.21 & 2.70 & 3.38 & -1.06 & 2.34 & 1.49 & 1.93 \\
  \midrule 
  \multicolumn{12}{l}{\textit{Chain-of-Thought with Best-of-N Selection (BoN@16)}} \\
  \midrule 
  Skywork~\cite{DBLP:journals/corr/abs-2410-18451} & 0.61 & 0.20 & 0.41 & 0.29 & 2.16 & 3.33 & 2.64 & 3.11 & -0.22 & 2.31 & 1.44 & 1.98 \\
  ArmoRM~\cite{DBLP:conf/emnlp/00030X0024} & 0.59 & 0.15 & 0.41 & 0.29 & 2.26 & 3.46 & 2.66 & 3.16 & \textbf{\underline{-0.19}} & 2.35 & 1.58 & 2.04 \\
  RLHFlow~\cite{dong2024rlhf} & 0.60 & 0.15 & 0.50 & 0.27 & \textbf{\underline{2.02}} & 3.19 & 2.41 & 3.18 & -0.42 & 1.63 & 1.42 & 1.88 \\
  Shepherd~\cite{DBLP:conf/acl/WangLSXDLCWS24} & 0.61 & 0.12 & 0.45 & 0.31 & 2.14 & 3.40 & 2.49 & 2.95 & -0.10 & 2.13 & 1.52 & 2.04 \\
  \midrule
  \multicolumn{12}{l}{\textit{Proposed Method (\nospacemethod @16)}} \\
  \midrule
  \method & \textbf{\underline{0.66}} & \textbf{\underline{0.35}} & \textbf{\underline{0.63}} & \textbf{\underline{0.57}} & 2.06 & \textbf{\underline{2.65}} & \textbf{\underline{1.96}} & \textbf{\underline{2.15}} & -0.60 & \textbf{\underline{0.20}} & \textbf{\underline{0.34}} & \textbf{\underline{-0.98}} \\
  \bottomrule
\end{tabular}
\end{footnotesize}
\end{table*}

\subsection{Performance of \method}

\paragraph{Comparison with baselines. } Table~\ref{tab:mab-usamo-human-full} presents the performance of four models from different providers across the QWK, RMSE, and Bias metrics. To ensure a comprehensive evaluation, we compare \method against three primary baselines: zero-shot CoT@1, MajorityVote@16 using three distinct prompt templates, and Best-of-N selection utilizing four reward models applied to the MathArena template outputs. Our results indicate that \method consistently achieves the highest correlation with human experts, as evidenced by higher QWK scores across all tested models. While \method significantly improves RMSE and Bias for Kimi-K2.5, Gemini-3-Pro, and Qwen3.5-27B, we observe that specific reward models such as RLHFlow and ArmoRM yield slightly better results for DeepSeek-V4-Flash in terms of error magnitude and systematic bias. Overall, the proposed method provides a more stable alignment with human grading across diverse model architectures.

\section{Limitations and Future Work}
Our evaluation pipeline depends on the availability of both a markscheme and a golden reference solution. This assumption is reasonable for Olympiad-level grading, where they are produced alongside the competition, but it does not hold for most real-world applications such as peer review of new theorems or open-ended scientific verification. Writing a high-quality markscheme without access to a reference solution is itself a difficult task and remains an open problem that our work does not address.

The same dependency limits \method directly. Both the per-arm justifications and the oracle reward are framed against the reference proof, so the algorithm requires a markscheme that maps cleanly onto a golden solution. \method therefore does not apply to the Blind setting, where there is no point of comparison for either the candidate proof or the per-arm justifications. Extending the algorithm to reference-free regimes, for instance by having the grader first synthesize an outline of the expected proof that the oracle can then evaluate against, is a natural direction for future work.

\section{Conclusion}\vspace{-1mm}
We conducted a systematic study of LLM-based verification for Olympiad-level mathematics. We introduced a human-curated benchmark derived from USAMO with markscheme-aligned injected errors and expert grades, and used it to evaluate 20 frontier models across prompting and inference settings. Our results show that current LLM verifiers still fall short of expert-level agreement. We proposed \synthmethod for scalable adversarial proof generation to support evaluation and SFT and to improve grading alignment. Finally, we introduced \nospacemethod, a Multi-Armed Bandit method for verification that achieves the best overall performance, outperforming ProofGrader with $26.4\%$ higher QWK and $14\%$ less error. Our work suggests paths toward more reliable automated verification.

\goodbreak\newpage

\section*{Acknowledgments}
The work of Farzan Farnia is supported by a grant from the Research Grants Council of the Hong Kong Special Administrative Region, China, Project 14210725, and is also supported by CUHK Direct Research Grant with CUHK Project No. 4055164. Moreover, the work is partially supported by a grant under 1+1+1 CUHK-CUHK(SZ)-GDSTC Joint Collaboration Fund. Finally, the authors sincerely thank the anonymous reviewers and metareviewer for their helpful feedback and constructive suggestions.

\bibliographystyle{plain} 
\bibliography{Bibliography} 

\begin{thebibliography}{10}

\bibitem{anthropic2026claude46}
Anthropic.
\newblock Claude 4.6 opus.
\newblock \url{https://www.anthropic.com/news/claude-opus-4-6}, 2026.

\bibitem{anthropic2026sonnet46}
{Anthropic}.
\newblock Introducing {Claude Sonnet 4.6}, February 2026.

\bibitem{Auer2002Finitetime}
Peter Auer, Nicol{\`o} Cesa-Bianchi, and Paul Fischer.
\newblock Finite-time analysis of the multi-armed bandit problem.
\newblock {\em Machine Learning}, 2002.

\bibitem{DBLP:journals/corr/abs-2505-23281}
Mislav Balunovic, Jasper Dekoninck, Ivo Petrov, Nikola Jovanovic, and Martin~T. Vechev.
\newblock Matharena: Evaluating llms on uncontaminated math competitions.
\newblock {\em CoRR}, abs/2505.23281, 2025.

\bibitem{bertot2013interactive}
Yves Bertot and Pierre Cast{\'e}ran.
\newblock {\em Interactive theorem proving and program development: Coq’Art: the calculus of inductive constructions}.
\newblock Springer Science \& Business Media, 2013.

\bibitem{cao2025revivingdspadvancedtheorem}
Chenrui Cao, Liangcheng Song, Zenan Li, Xinyi Le, Xian Zhang, Hui Xue, and Fan Yang.
\newblock Reviving dsp for advanced theorem proving in the era of reasoning models.
\newblock In {\em The Thirty-ninth Annual Conference on Neural Information Processing Systems}, 2025.

\bibitem{chen2025llmevaluatorspreferreason}
Wei-Lin Chen, Zhepei Wei, Xinyu Zhu, Shi Feng, and Yu~Meng.
\newblock Do llm evaluators prefer themselves for a reason?, 2025.

\bibitem{chen2026harnessingmultiplelargelanguage}
Zhijun Chen, Xiaodong Lu, Jingzheng Li, Pengpeng Chen, Zhuoran Li, Kai Sun, Yuankai Luo, Qianren Mao, Ming Li, Likang Xiao, Dingqi Yang, Xiao Huang, Yikun Ban, Hailong Sun, and Philip~S. Yu.
\newblock Harnessing multiple large language models: A survey on llm ensemble, 2026.

\bibitem{cobbe2021trainingverifierssolvemath}
Karl Cobbe, Vineet Kosaraju, Mohammad Bavarian, Mark Chen, Heewoo Jun, Lukasz Kaiser, Matthias Plappert, Jerry Tworek, Jacob Hilton, Reiichiro Nakano, Christopher Hesse, and John Schulman.
\newblock Training verifiers to solve math word problems, 2021.

\bibitem{AIME2025}
American~Mathematics Competitions and Art of~Problem~Solving.
\newblock Aime problems and solutions — 2025.
\newblock \url{https://artofproblemsolving.com/wiki/index.php/AIME_Problems_and_Solutions}, 2025.

\bibitem{deepseekv4}
{DeepSeek-AI}.
\newblock Deepseek-v4: Towards highly efficient million-token context intelligence, April 2026.

\bibitem{deepseekai2025deepseekv32pushingfrontieropen}
DeepSeek-AI, Aixin Liu, Aoxue Mei, Bangcai Lin, Bing Xue, Bingxuan Wang, Bingzheng Xu, Bochao Wu, Bowei Zhang, Chaofan Lin, Chen Dong, Chengda Lu, Chenggang Zhao, Chengqi Deng, Chenhao Xu, Chong Ruan, Damai Dai, Daya Guo, Dejian Yang, Deli Chen, Erhang Li, Fangqi Zhou, Fangyun Lin, Fucong Dai, Guangbo Hao, Guanting Chen, Guowei Li, H.~Zhang, Hanwei Xu, Hao Li, Haofen Liang, Haoran Wei, Haowei Zhang, Haowen Luo, Haozhe Ji, Honghui Ding, Hongxuan Tang, Huanqi Cao, Huazuo Gao, Hui Qu, Hui Zeng, Jialiang Huang, Jiashi Li, Jiaxin Xu, Jiewen Hu, Jingchang Chen, Jingting Xiang, Jingyang Yuan, Jingyuan Cheng, Jinhua Zhu, Jun Ran, Junguang Jiang, Junjie Qiu, Junlong Li, Junxiao Song, Kai Dong, Kaige Gao, Kang Guan, Kexin Huang, Kexing Zhou, Kezhao Huang, Kuai Yu, Lean Wang, Lecong Zhang, Lei Wang, Liang Zhao, Liangsheng Yin, Lihua Guo, Lingxiao Luo, Linwang Ma, Litong Wang, Liyue Zhang, M.~S. Di, M.~Y Xu, Mingchuan Zhang, Minghua Zhang, Minghui Tang, Mingxu Zhou, Panpan Huang, Peixin Cong, Peiyi Wang, Qiancheng Wang,
  Qihao Zhu, Qingyang Li, Qinyu Chen, Qiushi Du, Ruiling Xu, Ruiqi Ge, Ruisong Zhang, Ruizhe Pan, Runji Wang, Runqiu Yin, Runxin Xu, Ruomeng Shen, Ruoyu Zhang, S.~H. Liu, Shanghao Lu, Shangyan Zhou, Shanhuang Chen, Shaofei Cai, Shaoyuan Chen, Shengding Hu, Shengyu Liu, Shiqiang Hu, Shirong Ma, Shiyu Wang, Shuiping Yu, Shunfeng Zhou, Shuting Pan, Songyang Zhou, Tao Ni, Tao Yun, Tian Pei, Tian Ye, Tianyuan Yue, Wangding Zeng, Wen Liu, Wenfeng Liang, Wenjie Pang, Wenjing Luo, Wenjun Gao, Wentao Zhang, Xi~Gao, Xiangwen Wang, Xiao Bi, Xiaodong Liu, Xiaohan Wang, Xiaokang Chen, Xiaokang Zhang, Xiaotao Nie, Xin Cheng, Xin Liu, Xin Xie, Xingchao Liu, Xingkai Yu, Xingyou Li, Xinyu Yang, Xinyuan Li, Xu~Chen, Xuecheng Su, Xuehai Pan, Xuheng Lin, Xuwei Fu, Y.~Q. Wang, Yang Zhang, Yanhong Xu, Yanru Ma, Yao Li, Yao Li, Yao Zhao, Yaofeng Sun, Yaohui Wang, Yi~Qian, Yi~Yu, Yichao Zhang, Yifan Ding, Yifan Shi, Yiliang Xiong, Ying He, Ying Zhou, Yinmin Zhong, Yishi Piao, Yisong Wang, Yixiao Chen, Yixuan Tan, Yixuan Wei, Yiyang
  Ma, Yiyuan Liu, Yonglun Yang, Yongqiang Guo, Yongtong Wu, Yu~Wu, Yuan Cheng, Yuan Ou, Yuanfan Xu, Yuduan Wang, Yue Gong, Yuhan Wu, Yuheng Zou, Yukun Li, Yunfan Xiong, Yuxiang Luo, Yuxiang You, Yuxuan Liu, Yuyang Zhou, Z.~F. Wu, Z.~Z. Ren, Zehua Zhao, Zehui Ren, Zhangli Sha, Zhe Fu, Zhean Xu, Zhenda Xie, Zhengyan Zhang, Zhewen Hao, Zhibin Gou, Zhicheng Ma, Zhigang Yan, Zhihong Shao, Zhixian Huang, Zhiyu Wu, Zhuoshu Li, Zhuping Zhang, Zian Xu, Zihao Wang, Zihui Gu, Zijia Zhu, Zilin Li, Zipeng Zhang, Ziwei Xie, Ziyi Gao, Zizheng Pan, Zongqing Yao, Bei Feng, Hui Li, J.~L. Cai, Jiaqi Ni, Lei Xu, Meng Li, Ning Tian, R.~J. Chen, R.~L. Jin, S.~S. Li, Shuang Zhou, Tianyu Sun, X.~Q. Li, Xiangyue Jin, Xiaojin Shen, Xiaosha Chen, Xinnan Song, Xinyi Zhou, Y.~X. Zhu, Yanping Huang, Yaohui Li, Yi~Zheng, Yuchen Zhu, Yunxian Ma, Zhen Huang, Zhipeng Xu, Zhongyu Zhang, Dongjie Ji, Jian Liang, Jianzhong Guo, Jin Chen, Leyi Xia, Miaojun Wang, Mingming Li, Peng Zhang, Ruyi Chen, Shangmian Sun, Shaoqing Wu, Shengfeng Ye, T.~Wang,
  W.~L. Xiao, Wei An, Xianzu Wang, Xiaowen Sun, Xiaoxiang Wang, Ying Tang, Yukun Zha, Zekai Zhang, Zhe Ju, Zhen Zhang, and Zihua Qu.
\newblock Deepseek-v3.2: Pushing the frontier of open large language models, 2025.

\bibitem{dekoninck2026the}
Jasper Dekoninck, Ivo Petrov, Kristian Minchev, Miroslav Marinov, Maria Drencheva, Lyuba Konova, Milen~Milenov Shumanov, Kaloyan Tsvetkov, Nikolay Drenchev, Lazar~D. Todorov, Kalina Nikolova, Nikolay Georgiev, Vanesa Kalinkova, Margulan Ismoldayev, Mislav Balunovic, and Martin Vechev.
\newblock The open proof corpus: A large-scale study of {LLM}-generated mathematical proofs.
\newblock In {\em The Fourteenth International Conference on Learning Representations}, 2026.

\bibitem{dong2024rlhf}
Hanze Dong, Wei Xiong, Bo~Pang, Haoxiang Wang, Han Zhao, Yingbo Zhou, Nan Jiang, Doyen Sahoo, Caiming Xiong, and Tong Zhang.
\newblock Rlhf workflow: From reward modeling to online rlhf.
\newblock {\em arXiv preprint arXiv:2405.07863}, 2024.

\bibitem{comanici2025gemini25pushingfrontier}
Gheorghe~Comanici et~al.
\newblock Gemini 2.5: Pushing the frontier with advanced reasoning, multimodality, long context, and next generation agentic capabilities, 2025.

\bibitem{e98ec698-e30f-319f-8f04-9cae556cb352}
Nancy Falchikov and Judy Goldfinch.
\newblock Student peer assessment in higher education: A meta-analysis comparing peer and teacher marks.
\newblock {\em Review of Educational Research}, 70(3):287--322, 2000.

\bibitem{frick2025how}
Evan Frick, Tianle Li, Connor Chen, Wei-Lin Chiang, Anastasios~Nikolas Angelopoulos, Jiantao Jiao, Banghua Zhu, Joseph~E. Gonzalez, and Ion Stoica.
\newblock How to evaluate reward models for {RLHF}.
\newblock In {\em The Thirteenth International Conference on Learning Representations}, 2025.

\bibitem{gao_omni-math_2024}
Bofei Gao, Feifan Song, Zhe Yang, Zefan Cai, Yibo Miao, Qingxiu Dong, Lei Li, Chenghao Ma, Liang Chen, Runxin Xu, Zhengyang Tang, Benyou Wang, Daoguang Zan, Shanghaoran Quan, Ge~Zhang, Lei Sha, Yichang Zhang, Xuancheng Ren, Tianyu Liu, and Baobao Chang.
\newblock Omni-{MATH}: {A} {Universal} {Olympiad} {Level} {Mathematic} {Benchmark} {For} {Large} {Language} {Models}, October 2024.
\newblock arXiv:2410.07985.

\bibitem{gebru2021datasheets}
Timnit Gebru, Jamie Morgenstern, Briana Vecchione, Jennifer~Wortman Vaughan, Hanna Wallach, Hal~Daum{\'e} III, and Kate Crawford.
\newblock Datasheets for datasets.
\newblock {\em Communications of the ACM}, 64(12):86--92, 2021.

\bibitem{google2025gemini3}
{Google}.
\newblock A new era of intelligence with gemini 3, November 2025.

\bibitem{alphaproof}
{Google DeepMind}.
\newblock {AI} achieves silver-medal standard solving international mathematical olympiad problems.
\newblock \url{https://deepmind.google/discover/blog/ai-solves-imo-problems-at-silver-medal-level/}, 2024.
\newblock Accessed: 2025-05-08.

\bibitem{gu2025surveyllmasajudge}
Jiawei Gu, Xuhui Jiang, Zhichao Shi, Hexiang Tan, Xuehao Zhai, Chengjin Xu, Wei Li, Yinghan Shen, Shengjie Ma, Honghao Liu, Saizhuo Wang, Kun Zhang, Yuanzhuo Wang, Wen Gao, Lionel Ni, and Jian Guo.
\newblock A survey on llm-as-a-judge, 2025.

\bibitem{he2024olympiadbenchchallengingbenchmarkpromoting}
Chaoqun He, Renjie Luo, Yuzhuo Bai, Shengding Hu, Zhen~Leng Thai, Junhao Shen, Jinyi Hu, Xu~Han, Yujie Huang, Yuxiang Zhang, Jie Liu, Lei Qi, Zhiyuan Liu, and Maosong Sun.
\newblock Olympiadbench: A challenging benchmark for promoting agi with olympiad-level bilingual multimodal scientific problems, 2024.

\bibitem{huang2025winning}
Yichen Huang and Lin~F. Yang.
\newblock Winning gold at {IMO} 2025 with a model-agnostic verification-and-refinement pipeline.
\newblock In {\em The 5th Workshop on Mathematical Reasoning and AI at NeurIPS 2025}, 2025.

\bibitem{jiang2023draft}
Albert~Qiaochu Jiang, Sean Welleck, Jin~Peng Zhou, Wenda Li, Jiacheng Liu, Mateja Jamnik, Timoth{'{e}}e Lacroix, Yuhuai Wu, and Guillaume Lample.
\newblock {Draft}, {Sketch}, and {Prove}: {Guiding} formal theorem provers with informal proofs.
\newblock In {\em International Conference on Learning Representations}, 2023.

\bibitem{kim2024prometheus2opensource}
Seungone Kim, Juyoung Suk, Shayne Longpre, Bill~Yuchen Lin, Jamin Shin, Sean Welleck, Graham Neubig, Moontae Lee, Kyungjae Lee, and Minjoon Seo.
\newblock Prometheus 2: An open source language model specialized in evaluating other language models, 2024.

\bibitem{LAI19854}
T.L Lai and Herbert Robbins.
\newblock Asymptotically efficient adaptive allocation rules.
\newblock {\em Advances in Applied Mathematics}, 6(1):4--22, 1985.

\bibitem{li2025from}
Tianle Li, Wei-Lin Chiang, Evan Frick, Lisa Dunlap, Tianhao Wu, Banghua Zhu, Joseph~E. Gonzalez, and Ion Stoica.
\newblock From crowdsourced data to high-quality benchmarks: Arena-hard and benchbuilder pipeline.
\newblock In {\em Forty-second International Conference on Machine Learning}, 2025.

\bibitem{lightman2024lets}
Hunter Lightman, Vineet Kosaraju, Yuri Burda, Harrison Edwards, Bowen Baker, Teddy Lee, Jan Leike, John Schulman, Ilya Sutskever, and Karl Cobbe.
\newblock Let's verify step by step.
\newblock In {\em The Twelfth International Conference on Learning Representations}, 2024.

\bibitem{DBLP:journals/corr/abs-2508-03613}
Yong Lin, Shange Tang, Bohan Lyu, Ziran Yang, Jui{-}Hui Chung, Haoyu Zhao, Lai Jiang, Yihan Geng, Jiawei Ge, Jingruo Sun, Jiayun Wu, Jiri Gesi, Ximing Lu, David Acuna, Kaiyu Yang, Hongzhou Lin, Yejin Choi, Danqi Chen, Sanjeev Arora, and Chi Jin.
\newblock Goedel-prover-v2: Scaling formal theorem proving with scaffolded data synthesis and self-correction.
\newblock {\em CoRR}, abs/2508.03613, 2025.

\bibitem{DBLP:journals/corr/abs-2410-18451}
Chris~Yuhao Liu, Liang Zeng, Jiacai Liu, Rui Yan, Jujie He, Chaojie Wang, Shuicheng Yan, Yang Liu, and Yahui Zhou.
\newblock Skywork-reward: Bag of tricks for reward modeling in llms.
\newblock {\em CoRR}, abs/2410.18451, 2024.

\bibitem{DBLP:conf/emnlp/LuongHNGCSKBLMZHMKABSTL25}
Thang Luong, Dawsen Hwang, Hoang~H. Nguyen, Golnaz Ghiasi, Yuri Chervonyi, Insuk Seo, Junsu Kim, Garrett Bingham, Jonathan Lee, Swaroop Mishra, Alex Zhai, Clara~Huiyi Hu, Henryk Michalewski, Jimin Kim, Jeonghyun Ahn, Junhwi Bae, Xingyou Song, Trieu~H. Trinh, Quoc~V. Le, and Junehyuk Jung.
\newblock Towards robust mathematical reasoning.
\newblock In Christos Christodoulopoulos, Tanmoy Chakraborty, Carolyn Rose, and Violet Peng, editors, {\em Proceedings of the 2025 Conference on Empirical Methods in Natural Language Processing, {EMNLP} 2025, Suzhou, China, November 4-9, 2025}, pages 35418--35442. Association for Computational Linguistics, 2025.

\bibitem{DBLP:journals/corr/abs-2510-13888}
Wenjie Ma, Andrei Cojocaru, Neel Kolhe, Bradley Louie, Robin~Said Sharif, Haihan Zhang, Vincent Zhuang, Matei Zaharia, and Sewon Min.
\newblock Reliable fine-grained evaluation of natural language math proofs.
\newblock {\em CoRR}, abs/2510.13888, 2025.

\bibitem{lean4paper}
Leonardo~de Moura and Sebastian Ullrich.
\newblock The lean 4 theorem prover and programming language.
\newblock In {\em Automated Deduction – CADE 28: 28th International Conference on Automated Deduction, Virtual Event, July 12–15, 2021, Proceedings}, page 625–635, Berlin, Heidelberg, 2021. Springer-Verlag.

\bibitem{nipkow2002isabelle}
Tobias Nipkow, Lawrence~C Paulson, and Markus Wenzel.
\newblock {\em Isabelle/HOL: a proof assistant for higher-order logic}, volume 2283.
\newblock Springer Science \& Business Media, 2002.

\bibitem{openai2024gpt4ocard}
OpenAI.
\newblock Gpt-4o system card, 2024.

\bibitem{singh2026openaigpt5card}
OpenAI.
\newblock Openai gpt-5 system card, 2026.

\bibitem{ospanov2025apolloautomatedllmlean}
Azim Ospanov, Farzan Farnia, and Roozbeh Yousefzadeh.
\newblock Apollo: Automated llm and lean collaboration for advanced formal reasoning.
\newblock In {\em The Thirty-ninth Annual Conference on Neural Information Processing Systems}, 2025.

\bibitem{ospanov2026minifflean}
Azim Ospanov, Farzan Farnia, and Roozbeh Yousefzadeh.
\newblock minif2f-lean revisited: Reviewing limitations and charting a path forward.
\newblock In {\em The Thirty-ninth Annual Conference on Neural Information Processing Systems}, 2026.

\bibitem{ospanov2025hermesefficientverifiablemathematical}
Azim Ospanov, Zijin Feng, Jiacheng Sun, Haoli Bai, Xin Shen, and Farzan Farnia.
\newblock Hermes: Towards efficient and verifiable mathematical reasoning in llms, 2025.

\bibitem{paster2023openwebmath}
Keiran Paster, Marco Dos~Santos, Zhangir Azerbayev, and Jimmy Ba.
\newblock {OpenWebMath}: An open dataset of high-quality mathematical web text.
\newblock In {\em The Twelfth International Conference on Learning Representations}, 2023.

\bibitem{petrov2025proof}
Ivo Petrov, Jasper Dekoninck, Lyuben Baltadzhiev, Maria Drencheva, Kristian Minchev, Mislav Balunovic, Nikola Jovanovi{\'c}, and Martin Vechev.
\newblock Proof or bluff? evaluating {LLM}s on 2025 {USA} math olympiad.
\newblock In {\em 2nd AI for Math Workshop @ ICML 2025}, 2025.

\bibitem{petrov2025brokenmathbenchmarksycophancytheorem}
Ivo Petrov, Jasper Dekoninck, and Martin Vechev.
\newblock Brokenmath: A benchmark for sycophancy in theorem proving with llms, 2025.

\bibitem{qwen3.5}
{Qwen Team}.
\newblock {Qwen3.5}: Towards native multimodal agents, February 2026.

\bibitem{rein2024gpqa}
David Rein, Betty~Li Hou, Asa~Cooper Stickland, Jackson Petty, Richard~Yuanzhe Pang, Julien Dirani, Julian Michael, and Samuel~R. Bowman.
\newblock {GPQA}: A graduate-level google-proof q\&a benchmark.
\newblock In {\em First Conference on Language Modeling}, 2024.

\bibitem{DBLP:journals/corr/abs-2505-11774}
James~V. Roggeveen, Erik~Y. Wang, Will Flintoft, Peter Donets, Lucy~S. Nathwani, Nickholas Gutierrez, David Ettel, Anton~Marius Graf, Siddharth Dandavate, Arjun Nageswaran, Raglan Ward, Ava Williamson, Anne Mykland, Kacper~K. Migacz, Yijun Wang, Egemen Bostan, Duy~Thuc Nguyen, Zhe He, Marc~L. Descoteaux, Felix Yeung, Shida Liu, Jorge~Garc{\'{\i}}a Ponce, Luke Zhu, Yuyang Chen, Ekaterina~S. Ivshina, Miguel Fernandez, Minjae Kim, Kennan Gumbs, Matthew~Scott Tan, Russell Yang, Mai Hoang, David Brown, Isabella~A. Silveira, Lavon Sykes, Ahmed Roman, William Fredenberg, Yiming Chen, Lucas Martin, Yixing Tang, Kelly~Werker Smith, Hongyu Liao, Logan~G. Wilson, Alexander~Dazhen Cai, Andrea~Elizabeth Biju, and Michael~P. Brenner.
\newblock Hardmath2: {A} benchmark for applied mathematics built by students as part of a graduate class.
\newblock In {\em The Thirty-ninth Annual Conference on Neural Information Processing Systems}, 2025.

\bibitem{DBLP:journals/corr/abs-2511-22570}
Zhihong Shao, Yuxiang Luo, Chengda Lu, Z.~Z. Ren, Jiewen Hu, Tian Ye, Zhibin Gou, Shirong Ma, and Xiaokang Zhang.
\newblock Deepseekmath-v2: Towards self-verifiable mathematical reasoning.
\newblock {\em CoRR}, abs/2511.22570, 2025.

\bibitem{tan2025judgebench}
Sijun Tan, Siyuan Zhuang, Kyle Montgomery, William~Yuan Tang, Alejandro Cuadron, Chenguang Wang, Raluca Popa, and Ion Stoica.
\newblock Judgebench: A benchmark for evaluating {LLM}-based judges.
\newblock In {\em The Thirteenth International Conference on Learning Representations}, 2025.

\bibitem{DBLP:conf/icml/TangZWW24}
Zhengyang Tang, Xingxing Zhang, Benyou Wang, and Furu Wei.
\newblock Mathscale: Scaling instruction tuning for mathematical reasoning.
\newblock In {\em Forty-first International Conference on Machine Learning, {ICML} 2024, Vienna, Austria, July 21-27, 2024}. OpenReview.net, 2024.

\bibitem{kimiteam2026kimik25visualagentic}
Kimi Team.
\newblock Kimi k2.5: Visual agentic intelligence, 2026.

\bibitem{qwen35blog}
Qwen Team.
\newblock Qwen3.5: Accelerating productivity with native multimodal agents, February 2026.

\bibitem{thakur2025judgingjudgesevaluatingalignment}
Aman~Singh Thakur, Kartik Choudhary, Venkat~Srinik Ramayapally, Sankaran Vaidyanathan, and Dieuwke Hupkes.
\newblock Judging the judges: Evaluating alignment and vulnerabilities in llms-as-judges, 2025.

\bibitem{AlphaGeometryTrinh2024}
Trieu Trinh, Yuhuai Wu, Quoc Le, He~He, and Thang Luong.
\newblock Solving olympiad geometry without human demonstrations.
\newblock {\em Nature}, 2024.

\bibitem{tsoukalas2024putnambenchevaluatingneuraltheoremprovers}
George Tsoukalas, Jasper Lee, John Jennings, Jimmy Xin, Michelle Ding, Michael Jennings, Amitayush Thakur, and Swarat Chaudhuri.
\newblock Putnambench: Evaluating neural theorem-provers on the putnam mathematical competition, 2024.

\bibitem{verga2024replacingjudgesjuriesevaluating}
Pat Verga, Sebastian Hofstatter, Sophia Althammer, Yixuan Su, Aleksandra Piktus, Arkady Arkhangorodsky, Minjie Xu, Naomi White, and Patrick Lewis.
\newblock Replacing judges with juries: Evaluating llm generations with a panel of diverse models, 2024.

\bibitem{DBLP:conf/emnlp/00030X0024}
Haoxiang Wang, Wei Xiong, Tengyang Xie, Han Zhao, and Tong Zhang.
\newblock Interpretable preferences via multi-objective reward modeling and mixture-of-experts.
\newblock In Yaser Al{-}Onaizan, Mohit Bansal, and Yun{-}Nung Chen, editors, {\em Findings of the Association for Computational Linguistics: {EMNLP} 2024, Miami, Florida, USA, November 12-16, 2024}, pages 10582--10592. Association for Computational Linguistics, 2024.

\bibitem{DBLP:conf/acl/WangLSXDLCWS24}
Peiyi Wang, Lei Li, Zhihong Shao, Runxin Xu, Damai Dai, Yifei Li, Deli Chen, Yu~Wu, and Zhifang Sui.
\newblock Math-shepherd: Verify and reinforce llms step-by-step without human annotations.
\newblock In {\em Proceedings of the 62nd Annual Meeting of the Association for Computational Linguistics (Volume 1: Long Papers), {ACL} 2024, Bangkok, Thailand, August 11-16, 2024}, pages 9426--9439. Association for Computational Linguistics, 2024.

\bibitem{wei2022chain}
Jason Wei, Xuezhi Wang, Dale Schuurmans, Maarten Bosma, brian ichter, Fei Xia, Ed~H. Chi, Quoc~V Le, and Denny Zhou.
\newblock Chain of thought prompting elicits reasoning in large language models.
\newblock In Alice~H. Oh, Alekh Agarwal, Danielle Belgrave, and Kyunghyun Cho, editors, {\em Advances in Neural Information Processing Systems}, 2022.

\bibitem{welleck2022naturalprover}
Sean Welleck, Jiacheng Liu, Ximing Lu, Hannaneh Hajishirzi, and Yejin Choi.
\newblock Naturalprover: Grounded mathematical proof generation with language models.
\newblock In Alice~H. Oh, Alekh Agarwal, Danielle Belgrave, and Kyunghyun Cho, editors, {\em Advances in Neural Information Processing Systems}, 2022.

\bibitem{wood2021establishing}
Scott Wood, Erin Yao, Lisa Haisfield, and Susan Lottridge.
\newblock Establishing standards of best practice in automated scoring.
\newblock Technical report, ACT, Inc., July 2021.

\bibitem{yang2025qwen3technicalreport}
An~Yang, Anfeng Li, Baosong Yang, Beichen Zhang, Binyuan Hui, Bo~Zheng, Bowen Yu, Chang Gao, Chengen Huang, Chenxu Lv, Chujie Zheng, Dayiheng Liu, Fan Zhou, Fei Huang, Feng Hu, Hao Ge, Haoran Wei, Huan Lin, Jialong Tang, Jian Yang, Jianhong Tu, Jianwei Zhang, Jianxin Yang, Jiaxi Yang, Jing Zhou, Jingren Zhou, Junyang Lin, Kai Dang, Keqin Bao, Kexin Yang, Le~Yu, Lianghao Deng, Mei Li, Mingfeng Xue, Mingze Li, Pei Zhang, Peng Wang, Qin Zhu, Rui Men, Ruize Gao, Shixuan Liu, Shuang Luo, Tianhao Li, Tianyi Tang, Wenbiao Yin, Xingzhang Ren, Xinyu Wang, Xinyu Zhang, Xuancheng Ren, Yang Fan, Yang Su, Yichang Zhang, Yinger Zhang, Yu~Wan, Yuqiong Liu, Zekun Wang, Zeyu Cui, Zhenru Zhang, Zhipeng Zhou, and Zihan Qiu.
\newblock Qwen3 technical report, 2025.

\bibitem{yousefzadeh2025leandatasetinternationalmath}
Roozbeh Yousefzadeh, Xuenan Cao, and Azim Ospanov.
\newblock A lean dataset for international math olympiad: Small steps towards writing math proofs for hard problems, 2025.

\bibitem{yue2024harpchallenginghumanannotatedmath}
Albert~S. Yue, Lovish Madaan, Ted Moskovitz, DJ~Strouse, and Aaditya~K. Singh.
\newblock Harp: A challenging human-annotated math reasoning benchmark, 2024.

\bibitem{zheng2022minif2fcrosssystembenchmarkformal}
Kunhao Zheng, Jesse~Michael Han, and Stanislas Polu.
\newblock {miniF2F}: a cross-system benchmark for formal {Olympiad}-level mathematics, 2022.

\bibitem{zheng2023judging}
Lianmin Zheng, Wei-Lin Chiang, Ying Sheng, Siyuan Zhuang, Zhanghao Wu, Yonghao Zhuang, Zi~Lin, Zhuohan Li, Dacheng Li, Eric Xing, Hao Zhang, Joseph~E. Gonzalez, and Ion Stoica.
\newblock Judging {LLM}-as-a-judge with {MT}-bench and chatbot arena.
\newblock In {\em Thirty-seventh Conference on Neural Information Processing Systems Datasets and Benchmarks Track}, 2023.

\bibitem{zheng2024llamafactory}
Yaowei Zheng, Richong Zhang, Junhao Zhang, Yanhan Ye, Zheyan Luo, Zhangchi Feng, and Yongqiang Ma.
\newblock Llamafactory: Unified efficient fine-tuning of 100+ language models.
\newblock In {\em Proceedings of the 62nd Annual Meeting of the Association for Computational Linguistics (Volume 3: System Demonstrations)}, Bangkok, Thailand, 2024. Association for Computational Linguistics.

\end{thebibliography}


\newpage\appendix

\addtocontents{toc}{\protect\setcounter{tocdepth}{2}}
\etocsetnexttocdepth{1}
\etocsettocstyle{\section*{Contents}\medskip}{}
\tableofcontents

\goodbreak\newpage

\section{Datasheets} \label{app:datasheet}
Following the framework of \cite{gebru2021datasheets} and \cite{paster2023openwebmath}, Table~\ref{tab:datasheet-human} provides the additional information about our expert-annotated dataset and Table~\ref{tab:datasheet-synth} provides additional information about our synthetically generated dataset.

\renewcommand{\arraystretch}{1.25}

\begin{center}
\begin{longtable}{p{6.5cm} | p{6.5cm} }
\caption[Datasheet for our dataset]{Datasheet for our expert annotated dataset} \label{tab:datasheet-human} \\

\hline \multicolumn{1}{c}{\textbf{Questions}} & \multicolumn{1}{|c}{\textbf{Answers}} \\ \hline 
\endfirsthead

\hline \multicolumn{2}{r}{{Continued on next page}} \\ \hline
\endfoot

\hline \hline
\endlastfoot

    \multicolumn{2}{c}{\textbf{Motivation}}                   \\
    \hline
    For what purpose was the dataset created?  &  The dataset was developed to benchmark the performance of current Large Language Models in grading and evaluating Olympiad-level mathematical proofs.    \\
    \hline
    Who created the dataset and on behalf of which entity?  &  The authors of this paper.    \\
    \hline
    Who funded the creation of the dataset?  &  The company where the authors work.    \\
    \hline
    Any other comment?  &  None.    \\
    \hline
    \multicolumn{2}{c}{\textbf{Composition}}  \\
    \hline
    What do the instances that comprise the dataset represent?  &  USAMO 2025 problems that consist of problem statements, golden solutions, corresponding markschemes, candidate proofs with minor errors and corresponding expert assigned grades.    \\
    \hline
    How many instances are there in total?  &  48 instances    \\
    \hline
    Does the dataset contain all possible instances or is it a sample of instances from a larger set?  &  Yes, we present all instances.    \\
    \hline
    What data does each instance consist of?  &  A problem statement, golden solution, corresponding markscheme, candidate solution with logical errors and corresponding expert grades.    \\
    \hline
    Is there a label or target associated with each instance?  &  Only human assigned grades.    \\
    \hline
    Is any information missing from individual instances?  &  No.    \\
    \hline
    Are relationships between individual instances made explicit?  &  Not applicable. \\
    \hline
    Are there recommended data splits?  &  No recommended split.   \\
    \hline
    Are there any errors, sources of noise, or redundancies in the dataset?  &  No, there are no errors in the proposed dataset to the best of our knowledge. Verification was done by two independent annotators.  \\
    \hline
    Is the dataset self-contained, or does it link to or otherwise rely on external resources?  &  The dataset is self-contained.    \\
    \hline
    Does the dataset contain data that might be considered confidential?  &  No.    \\
    \hline
    Does the dataset contain data that, if viewed directly, might be offensive, insulting, threatening, or might otherwise cause anxiety?  &  No.    \\
    \hline
    \multicolumn{2}{c}{\textbf{Collection Process}}  \\
    \hline
    How was the data associated with each instance acquired?  &  All 48 instances originate from our team and further verified using two-step collection described in \cite{rein2024gpqa}.   \\
    \hline
    What mechanisms or procedures were used to collect the data?  & The seed problems were taken from USAMO 2025 subset of the MathArena repository~\cite{DBLP:journals/corr/abs-2505-23281}. Based on corresponding markschemes, we injected mistakes that would result in a particular loss of points.  \\
    \hline
    If the dataset is a sample from a larger set, what was the sampling strategy?  &  No, there is no larger version.    \\
    \hline
    Who was involved in the data collection process and how were they compensated?  &  Authors of this work are responsible for data collection.    \\
    \hline
    Over what time frame was the data collected?  &  The dataset was collected over a four-month period, from January 2026 through April 2026.    \\
    \hline
    Were any ethical review processes conducted?  &  No.    \\
    \hline
    \multicolumn{2}{c}{\textbf{Preprocessing/cleaning/labeling}}  \\
    \hline
    Was any preprocessing/cleaning/labeling of the data done?  &  Yes, we manually assigned grades to each sample and preprocessed the data to follow MathArena~\cite{DBLP:journals/corr/abs-2505-23281} format.  \\
    \hline
    Was the “raw” data saved in addition to the preprocessed/cleaned/labeled data?  &  No, all raw data was converted to preprocessed samples.    \\
    \hline
    Is the software that was used to preprocess/clean/label the data available?  &  We used open-source text editors to perform annotation.  \\
    \hline
    Any other comments?  &  No.    \\
    \hline
    \multicolumn{2}{c}{\textbf{Uses}}  \\
    \hline
    Has the dataset been used for any tasks already?  &  We evaluated a large number of LLMs and measured their correlation with human labels. Moreover, we tested our proposed \method agent.  \\
    \hline
    Is there a repository that links to any or all papers or systems that use the dataset?  &  Public GitHub and HuggingFace links will be provided at a later date.    \\
    \hline
    What (other) tasks could the dataset be used for?  &  The dataset may be used for benchmarking alignment with human grades in LLM-as-a-Judge scenarios.   \\
    \hline
    Is there anything about the composition of the dataset or the way it was collected and preprocessed/cleaned/labeled that might impact future uses?  &  We do not anticipate future issues emerging from the proposed benchmark dataset.   \\
    \hline
    Are there tasks for which the dataset should not be used?  &  The data is primarily for benchmarking, so we discourage others from training their models on this dataset.    \\
    \hline
    Any other comments?  &  No.    \\
    \hline
    \multicolumn{2}{c}{\textbf{Distribution}}  \\
    \hline
    Will the dataset be distributed to third parties outside of the entity on behalf of which the dataset was created?  &  Yes, we will release the dataset on public platforms such as GitHub and HuggingFace at a later date.    \\
    \hline
    How will the dataset will be distributed?  &  GitHub, HuggingFace.    \\
    \hline
    When will the dataset be distributed?  &  The dataset will be distributed along with the camera-ready version of the submission.    \\
    \hline
    Will the dataset be distributed under a copyright or other intellectual property license, and/or under applicable terms of use?  &  Yes, the dataset will be released under the MIT license.   \\
    \hline
    Have any third parties imposed IP-based or other restrictions on the data associated with the instances?  &  No.   \\
    \hline
    Do any export controls or other regulatory restrictions apply to the dataset or to individual instances?  &  No.    \\
    \hline
    Any other comments?  &  No.    \\
    \hline
    \multicolumn{2}{c}{\textbf{Maintenance}}  \\
    \hline
    Who will be supporting/hosting/maintaining the dataset?  &  The first author of the submission.    \\
    \hline
    How can the owner/curator/manager of the dataset be contacted?  &  The curator of the dataset may be reached through an email or any other public means, such as GitHub profile.    \\
    \hline
    Is there an erratum?  &  At this stage there is no associated erratum.  \\
    \hline
    Will the dataset be updated?  &  Yes, we plan to periodically update the dataset if mistakes are found.  \\ 
    \hline
    If the dataset relates to people, are there applicable limits on the retention of the data associated with the instances?  &  Not applicable.    \\
    \hline
    Will older versions of the dataset continue to be supported/hosted/maintained?  &  Yes.    \\
    \hline
    If others want to extend/augment/build on/contribute to the dataset, is there a mechanism for them to do so?  &  Yes, since we release all accompanied information, others may use this data for other uses.   \\
    \hline
    Any other comments?  &  No.  
\end{longtable}
\end{center}

\goodbreak\newpage

\renewcommand{\arraystretch}{1.25}

\begin{center}
\begin{longtable}{p{6.5cm} | p{6.5cm} }
\caption[Datasheet for our dataset]{Datasheet for our synthetically generated dataset using \nospacesynthmethod.} \label{tab:datasheet-synth} \\

\hline \multicolumn{1}{c}{\textbf{Questions}} & \multicolumn{1}{|c}{\textbf{Answers}} \\ \hline 
\endfirsthead

\hline \multicolumn{2}{r}{{Continued on next page}} \\ \hline
\endfoot

\hline \hline
\endlastfoot

    \multicolumn{2}{c}{\textbf{Motivation}}                   \\
    \hline
    For what purpose was the dataset created?  &  The dataset was developed to benchmark the performance of current Large Language Models in grading and evaluating Olympiad-level mathematical proofs.    \\
    \hline
    Who created the dataset and on behalf of which entity?  &  The authors of this paper.    \\
    \hline
    Who funded the creation of the dataset?  &  The company where the authors work.    \\
    \hline
    Any other comment?  &  None.    \\
    \hline
    \multicolumn{2}{c}{\textbf{Composition}}  \\
    \hline
    What do the instances that comprise the dataset represent?  &  USAMO 2025/USAMO 2026/IMO 2025/IMC 2025 problems that consist of problem statements, golden solutions, corresponding markschemes, candidate proofs with minor errors and corresponding expert assigned grades.    \\
    \hline
    How many instances are there in total?  &  200 instances    \\
    \hline
    Does the dataset contain all possible instances or is it a sample of instances from a larger set?  &  Yes, we present all instances.    \\
    \hline
    What data does each instance consist of?  &  A problem statement, golden solution, corresponding markscheme, candidate solution with logical errors and corresponding expert grades.    \\
    \hline
    Is there a label or target associated with each instance?  &  Only LLM assigned target grades as described in the score profile.    \\
    \hline
    Is any information missing from individual instances?  &  No.    \\
    \hline
    Are relationships between individual instances made explicit?  &  Not applicable. \\
    \hline
    Are there recommended data splits?  &  No recommended split.   \\
    \hline
    Are there any errors, sources of noise, or redundancies in the dataset?  &  No, there are no errors in the proposed dataset to the best of our knowledge. Verification was done by a critic LLM, namely Claude-Opus-4.6-Think.  \\
    \hline
    Is the dataset self-contained, or does it link to or otherwise rely on external resources?  &  The dataset is self-contained.    \\
    \hline
    Does the dataset contain data that might be considered confidential?  &  No.    \\
    \hline
    Does the dataset contain data that, if viewed directly, might be offensive, insulting, threatening, or might otherwise cause anxiety?  &  No.    \\
    \hline
    \multicolumn{2}{c}{\textbf{Collection Process}}  \\
    \hline
    How was the data associated with each instance acquired?  &  All 200 instances originate from our team and further verified using critique-alignment process of our \synthmethod pipeline.   \\
    \hline
    What mechanisms or procedures were used to collect the data?  & The seed problems were taken from USAMO 2025/USAMO 2026/IMO 2025/IMC 2025 subsets of the MathArena repository~\cite{DBLP:journals/corr/abs-2505-23281}. Based on corresponding markschemes, LLM injected mistakes that would result in a particular loss of points as described in a given score profile.  \\
    \hline
    If the dataset is a sample from a larger set, what was the sampling strategy?  &  No, there is no larger version.    \\
    \hline
    Who was involved in the data collection process and how were they compensated?  &  Authors of this work are responsible for data collection.    \\
    \hline
    Over what time frame was the data collected?  &  The dataset was collected over a four-month period, from January 2026 through April 2026.    \\
    \hline
    Were any ethical review processes conducted?  &  No.    \\
    \hline
    \multicolumn{2}{c}{\textbf{Preprocessing/cleaning/labeling}}  \\
    \hline
    Was any preprocessing/cleaning/labeling of the data done?  &  Yes, we manually assigned grades to each sample and preprocessed the data to follow MathArena~\cite{DBLP:journals/corr/abs-2505-23281} format.  \\
    \hline
    Was the “raw” data saved in addition to the preprocessed/cleaned/labeled data?  &  No, all raw data was converted to preprocessed samples.    \\
    \hline
    Is the software that was used to preprocess/clean/label the data available?  &  We used open-source text editors to perform annotation.  \\
    \hline
    Any other comments?  &  No.    \\
    \hline
    \multicolumn{2}{c}{\textbf{Uses}}  \\
    \hline
    Has the dataset been used for any tasks already?  &  We evaluated a large number of LLMs and measured their correlation with LLM-assigned labels. Moreover, we tested our proposed \method agent.  \\
    \hline
    Is there a repository that links to any or all papers or systems that use the dataset?  &  Public GitHub and HuggingFace links will be provided at a later date.    \\
    \hline
    What (other) tasks could the dataset be used for?  &  The dataset may be used for benchmarking alignment with human grades in LLM-as-a-Judge scenarios.   \\
    \hline
    Is there anything about the composition of the dataset or the way it was collected and preprocessed/cleaned/labeled that might impact future uses?  &  We do not anticipate future issues emerging from the proposed benchmark dataset.   \\
    \hline
    Are there tasks for which the dataset should not be used?  &  The data is for benchmarking and may be used for training. At this stage we do not see scenarios for which we discourage others to use this data.    \\
    \hline
    Any other comments?  &  No.    \\
    \hline
    \multicolumn{2}{c}{\textbf{Distribution}}  \\
    \hline
    Will the dataset be distributed to third parties outside of the entity on behalf of which the dataset was created?  &  Yes, we will release the dataset on public platforms such as GitHub and HuggingFace at a later date.    \\
    \hline
    How will the dataset will be distributed?  &  GitHub, HuggingFace.    \\
    \hline
    When will the dataset be distributed?  &  The dataset will be distributed along with the camera-ready version of the submission.    \\
    \hline
    Will the dataset be distributed under a copyright or other intellectual property license, and/or under applicable terms of use?  &  Yes, the dataset will be released under the MIT license.   \\
    \hline
    Have any third parties imposed IP-based or other restrictions on the data associated with the instances?  &  No.   \\
    \hline
    Do any export controls or other regulatory restrictions apply to the dataset or to individual instances?  &  No.    \\
    \hline
    Any other comments?  &  No.    \\
    \hline
    \multicolumn{2}{c}{\textbf{Maintenance}}  \\
    \hline
    Who will be supporting/hosting/maintaining the dataset?  &  The first author of the submission.    \\
    \hline
    How can the owner/curator/manager of the dataset be contacted?  &  The curator of the dataset may be reached through an email or any other public means, such as GitHub profile.    \\
    \hline
    Is there an erratum?  &  At this stage there is no associated erratum.  \\
    \hline
    Will the dataset be updated?  &  Yes, we plan to periodically update the dataset if mistakes are found.  \\ 
    \hline
    If the dataset relates to people, are there applicable limits on the retention of the data associated with the instances?  &  Not applicable.    \\
    \hline
    Will older versions of the dataset continue to be supported/hosted/maintained?  &  Yes.    \\
    \hline
    If others want to extend/augment/build on/contribute to the dataset, is there a mechanism for them to do so?  &  Yes, since we release all accompanied information, others may use this data for other uses.   \\
    \hline
    Any other comments?  &  No.  
\end{longtable}
\end{center}

\goodbreak\newpage

\section{Distribution of injected errors in \humands}

Table~\ref{tab:trojan-h-error-distribution} shows the distribution of injected errors in the \humands dataset. During construction, we apply logical gap errors the most, since they are easier to control and spot right away. Moreover, a logical gap is the widest error category that is difficult to define with one term, so many unjustified claims go under this broad umbrella category. "Misapplied theorem" refers to errors with incorrect preliminaries or even a completely wrong theorem used to prove the statement. Conceptual errors refer to fundamental issues with the proof, i.e., wrong constructions or even interpretations of the problem. Computational slips are very minor mistakes that do not affect the grand scheme of things, but still lead to less rigorous proofs. "Incomplete justification" is a mistake in which the proof is only partially complete and technically correct, but fails to explain all conditions/steps.

\begin{table}[ht]
\centering
\caption{Distribution of injected error types in \nospacehumands. We report them as absolute counts and percentages.}
\label{tab:trojan-h-error-distribution}
\begin{tabular}{lcc}
\toprule
Error Type & Count & Percentage\\
\midrule
Logical gap & 27 & 35.06\% \\
Misapplied theorem & 18 & 23.38\% \\
Conceptual error & 12 & 15.58\% \\
Computational slip & 8 & 10.39\% \\
Incomplete justification & 6 & 7.79\% \\
None & 6 & 7.79\% \\
Total & 77 & 100.0\% \\
\bottomrule
\end{tabular}
\end{table}

\section{Evaluated Model List}
\label{app:model-list}

Table~\ref{tab:evaluated_models} summarises every grader that participates in our evaluation suite. The list spans 20 distinct models from major providers, covering both open-weight and closed-weight families together with their thinking-enabled and non-thinking variants where available. Each row reports which (dataset, prompting style) combinations a given model was run on. The two datasets are USAMO 2025 (the human-annotated benchmark, \nospacehumands) and the synthetic set (\nospacesynthds), and the two styles are the rubric-equipped Full template and the rubric-free Blind template. The final column lists the maximum completion-token budget observed across attempts, which clarifies the inference-cost regime each model operated in. We observe that models rarely hit the maximum token limit.

In total, the table tracks 52 model$\times$(dataset, style) configurations and serves as an index for the per-model results that follow later in this appendix. Coverage is not uniform across providers. Smaller open-weight reward-style models such as the Prometheus family were evaluated on \humands Full only, since they did not produce usable scores in the rubric-free regime. Frontier thinking-enabled models were evaluated more broadly because their scoring behaviour is the primary subject of our analysis.

\begin{table}[htbp]
\centering
\small
\setlength{\tabcolsep}{4pt}
\caption{The list of all 20 evaluated models. A \checkmark{} marks the presence of a sampling\_grades.csv-producing run for the indicated (dataset, style) pair. In total 52 model$\times$(dataset, style) runs are reported. The last column reports the maximum \texttt{completion\_tokens}.}
\label{tab:evaluated_models}
\begin{tabular}{lccccr}
\toprule
 & \multicolumn{2}{c}{\humands} & \multicolumn{2}{c}{\synthds} & \\
\cmidrule(lr){2-3} \cmidrule(lr){4-5}
Model & Full & Blind & Full & Blind & Max tokens \\
\midrule
Claude-Sonnet-4.6~\cite{anthropic2026sonnet46} (\textcolor{red}{T}) & \checkmark & \checkmark & $-$ & $-$ & 65,536 \\
Claude-Sonnet-4.6~\cite{anthropic2026sonnet46} (\textcolor{green}{T}) & \checkmark & \checkmark & \checkmark & \checkmark & 23,625 \\
DeepSeek-V3.2~\cite{deepseekai2025deepseekv32pushingfrontieropen} (\textcolor{red}{T}) & \checkmark & \checkmark & \checkmark & $-$ & 8,192 \\
DeepSeek-V3.2~\cite{deepseekai2025deepseekv32pushingfrontieropen} (\textcolor{green}{T}) & \checkmark & \checkmark & \checkmark & $-$ & 65,536 \\
DeepSeek-V4-Flash~\cite{deepseekv4} (\textcolor{red}{T}) & \checkmark & \checkmark & $-$ & $-$ & 65,536 \\
DeepSeek-V4-Flash~\cite{deepseekv4} (\textcolor{green}{T}) & \checkmark & \checkmark & \checkmark & \checkmark & 65,536 \\
DeepSeek-V4-Pro~\cite{deepseekv4} (\textcolor{green}{T}) & \checkmark & $-$ & $-$ & $-$ & 65,536 \\
Gemini-2.5-Flash~\cite{comanici2025gemini25pushingfrontier} & \checkmark & \checkmark & $-$ & $-$ & 32,768 \\
Gemini-2.5-Pro~\cite{comanici2025gemini25pushingfrontier} (\textcolor{red}{T}) & \checkmark & $-$ & $-$ & $-$ & 65,536 \\
Gemini-2.5-Pro~\cite{comanici2025gemini25pushingfrontier} (\textcolor{green}{T}) & \checkmark & $-$ & $-$ & $-$ & 65,536 \\
Gemini-3-Pro-Prev.~\cite{google2025gemini3} (\textcolor{green}{T}) & \checkmark & \checkmark & $-$ & $-$ & 65,536 \\
GPT-4o~\cite{openai2024gpt4ocard} & \checkmark & \checkmark & \checkmark & \checkmark & 4,096 \\
GPT-5~\cite{singh2026openaigpt5card} & \checkmark & \checkmark & \checkmark & \checkmark & 65,536 \\
GPT-5-Mini~\cite{singh2026openaigpt5card} & \checkmark & \checkmark & \checkmark & \checkmark & 32,768 \\
Kimi-K2.5~\cite{kimiteam2026kimik25visualagentic} & \checkmark & \checkmark & \checkmark & \checkmark & 32,768 \\
Prometheus-V2-7B~\cite{kim2024prometheus2opensource} & \checkmark & $-$ & $-$ & $-$ & 1024 \\
Prometheus-V2-8$\times$7B~\cite{kim2024prometheus2opensource} & \checkmark & $-$ & $-$ & $-$ & 1024 \\
Qwen3-8B~\cite{yang2025qwen3technicalreport} & \checkmark & \checkmark & $-$ & $-$ & 16,384 \\
Qwen3.5-27B~\cite{qwen35blog} & \checkmark & \checkmark & \checkmark & \checkmark & 32,768 \\
Qwen3.5-9B~\cite{qwen35blog} & \checkmark & $-$ & \checkmark & \checkmark & 16,384 \\
\bottomrule
\end{tabular}
\end{table}

\section{Extended evaluation results}
\label{app:extended-eval}

This section reports the full set of correlation and error metrics for every grader in Table~\ref{tab:evaluated_models}, broken down by inference style (Full and Blind), dataset (\nospacehumands\ and \nospacesynthds), and aggregation method (single CoT, Mean-CoT@16, and MajorityVote@16). The first two tables of each subsection report the mean and standard deviation of every metric across the 16 sampling attempts at the per-attempt level, which characterises the within-grader variance. The remaining six tables report the aggregated grade under the three aggregation methods, which is the quantity each grader exposes to a downstream user.

\subsection{\humands}

\begin{table}[htbp]
\centering
\caption{Mean $\pm$ standard deviation of Pearson $r$, Kendall $\tau$, QWK, and MAE across individual CoT sampling attempts per problem. Thinking-enabled and non-thinking model variants are denoted by \textcolor{green}{T} and \textcolor{red}{T}, respectively. Results are evaluated on the human-labelled \humands dataset against a standard markscheme and golden solution.}
\label{tab:human_full_cot_per_attempt_summary_part1}
\begin{footnotesize}
\begin{tabular}{lrrrr}
\toprule
                                    Model &       Pearson $r$ &    Kendall $\tau$ &               QWK &               MAE \\
\midrule
   Claude-Sonnet-4.6 (\textcolor{red}{T}) & 0.554 $\pm$ 0.025 & 0.451 $\pm$ 0.026 & 0.446 $\pm$ 0.024 & 1.785 $\pm$ 0.070 \\
 Claude-Sonnet-4.6 (\textcolor{green}{T}) & 0.604 $\pm$ 0.046 & 0.506 $\pm$ 0.043 & 0.513 $\pm$ 0.043 & 1.623 $\pm$ 0.087 \\
       DeepSeek-V3.2 (\textcolor{red}{T}) & 0.435 $\pm$ 0.011 & 0.342 $\pm$ 0.009 & 0.289 $\pm$ 0.008 & 2.292 $\pm$ 0.025 \\
     DeepSeek-V3.2 (\textcolor{green}{T}) & 0.622 $\pm$ 0.065 & 0.526 $\pm$ 0.058 & 0.613 $\pm$ 0.066 & 1.357 $\pm$ 0.136 \\
             deepseek-v4-flash-nothinking & 0.347 $\pm$ 0.014 & 0.286 $\pm$ 0.020 & 0.228 $\pm$ 0.010 & 2.363 $\pm$ 0.030 \\
 DeepSeek-V4-Flash (\textcolor{green}{T}) & 0.656 $\pm$ 0.012 & 0.544 $\pm$ 0.015 & 0.631 $\pm$ 0.021 & 1.375 $\pm$ 0.029 \\
                          deepseek-v4-pro & 0.594 $\pm$ 0.078 & 0.510 $\pm$ 0.069 & 0.586 $\pm$ 0.078 & 1.469 $\pm$ 0.133 \\
                         Gemini-2.5-Flash & 0.586 $\pm$ 0.068 & 0.478 $\pm$ 0.057 & 0.572 $\pm$ 0.067 & 1.512 $\pm$ 0.153 \\
      Gemini-2.5-Pro (\textcolor{red}{T}) & 0.555 $\pm$ 0.075 & 0.449 $\pm$ 0.074 & 0.462 $\pm$ 0.083 & 1.606 $\pm$ 0.146 \\
    Gemini-2.5-Pro (\textcolor{green}{T}) & 0.599 $\pm$ 0.062 & 0.484 $\pm$ 0.057 & 0.515 $\pm$ 0.063 & 1.500 $\pm$ 0.133 \\
Gemini-3-Pro-Prev. (\textcolor{green}{T}) & 0.565 $\pm$ 0.032 & 0.463 $\pm$ 0.035 & 0.470 $\pm$ 0.032 & 1.615 $\pm$ 0.077 \\
                                   GPT-4o & 0.216 $\pm$ 0.026 & 0.178 $\pm$ 0.020 & 0.062 $\pm$ 0.012 & 3.101 $\pm$ 0.032 \\
                                    GPT-5 & 0.620 $\pm$ 0.051 & 0.522 $\pm$ 0.050 & 0.557 $\pm$ 0.056 & 1.507 $\pm$ 0.102 \\
                               GPT-5-Mini & 0.519 $\pm$ 0.049 & 0.436 $\pm$ 0.043 & 0.421 $\pm$ 0.046 & 1.918 $\pm$ 0.117 \\
                                Kimi-K2.5 & 0.413 $\pm$ 0.091 & 0.328 $\pm$ 0.079 & 0.282 $\pm$ 0.067 & 2.234 $\pm$ 0.171 \\
                         Prometheus-V2-7B & 0.020 $\pm$ 0.097 & 0.004 $\pm$ 0.083 & 0.010 $\pm$ 0.068 & 2.857 $\pm$ 0.107 \\
                Prometheus-V2-8$\times$7B & 0.063 $\pm$ 0.109 & 0.060 $\pm$ 0.076 & 0.031 $\pm$ 0.053 & 2.895 $\pm$ 0.145 \\
                                 Qwen3-8B & 0.180 $\pm$ 0.053 & 0.186 $\pm$ 0.039 & 0.076 $\pm$ 0.024 & 3.070 $\pm$ 0.080 \\
                              Qwen3.5-27B & 0.396 $\pm$ 0.036 & 0.319 $\pm$ 0.042 & 0.287 $\pm$ 0.043 & 2.278 $\pm$ 0.112 \\
                               Qwen3.5-9B & 0.427 $\pm$ 0.069 & 0.373 $\pm$ 0.056 & 0.306 $\pm$ 0.064 & 2.370 $\pm$ 0.162 \\
\bottomrule
\end{tabular}
\end{footnotesize}
\end{table}

\begin{table}[htbp]
\centering
\caption{Mean $\pm$ standard deviation of RMSE, bias, and exact-match accuracy across individual CoT sampling attempts per problem. Thinking-enabled and non-thinking model variants are denoted by \textcolor{green}{T} and \textcolor{red}{T}, respectively. Results are evaluated on the human-labelled \humands dataset against a standard markscheme and golden solution.}
\label{tab:human_full_cot_per_attempt_summary_part2}
\begin{footnotesize}
\begin{tabular}{lrrr}
\toprule
                                    Model &              RMSE &               Bias &             Exact \\
\midrule
   Claude-Sonnet-4.6 (\textcolor{red}{T}) & 2.408 $\pm$ 0.050 &  1.349 $\pm$ 0.085 & 0.228 $\pm$ 0.045 \\
 Claude-Sonnet-4.6 (\textcolor{green}{T}) & 2.227 $\pm$ 0.096 &  1.177 $\pm$ 0.110 & 0.277 $\pm$ 0.038 \\
       DeepSeek-V3.2 (\textcolor{red}{T}) & 3.068 $\pm$ 0.022 &  2.073 $\pm$ 0.035 & 0.238 $\pm$ 0.011 \\
     DeepSeek-V3.2 (\textcolor{green}{T}) & 2.117 $\pm$ 0.201 &  0.260 $\pm$ 0.207 & 0.414 $\pm$ 0.039 \\
             deepseek-v4-flash-nothinking & 3.158 $\pm$ 0.029 &  2.058 $\pm$ 0.036 & 0.232 $\pm$ 0.022 \\
 DeepSeek-V4-Flash (\textcolor{green}{T}) & 2.111 $\pm$ 0.061 & -0.615 $\pm$ 0.165 & 0.380 $\pm$ 0.020 \\
                          deepseek-v4-pro & 2.255 $\pm$ 0.225 & -0.021 $\pm$ 0.195 & 0.401 $\pm$ 0.020 \\
                         Gemini-2.5-Flash & 2.217 $\pm$ 0.183 &  0.461 $\pm$ 0.209 & 0.325 $\pm$ 0.057 \\
      Gemini-2.5-Pro (\textcolor{red}{T}) & 2.358 $\pm$ 0.199 &  1.245 $\pm$ 0.227 & 0.339 $\pm$ 0.036 \\
    Gemini-2.5-Pro (\textcolor{green}{T}) & 2.186 $\pm$ 0.145 &  1.098 $\pm$ 0.111 & 0.330 $\pm$ 0.058 \\
Gemini-3-Pro-Prev. (\textcolor{green}{T}) & 2.454 $\pm$ 0.109 &  1.365 $\pm$ 0.088 & 0.359 $\pm$ 0.020 \\
                                   GPT-4o & 3.826 $\pm$ 0.025 &  3.064 $\pm$ 0.037 & 0.166 $\pm$ 0.013 \\
                                    GPT-5 & 2.112 $\pm$ 0.123 &  0.960 $\pm$ 0.084 & 0.285 $\pm$ 0.034 \\
                               GPT-5-Mini & 2.588 $\pm$ 0.126 &  1.456 $\pm$ 0.112 & 0.227 $\pm$ 0.036 \\
                                Kimi-K2.5 & 2.998 $\pm$ 0.178 &  1.944 $\pm$ 0.189 & 0.251 $\pm$ 0.028 \\
                         Prometheus-V2-7B & 3.564 $\pm$ 0.103 &  1.895 $\pm$ 0.272 & 0.138 $\pm$ 0.036 \\
                Prometheus-V2-8$\times$7B & 3.589 $\pm$ 0.131 &  2.464 $\pm$ 0.120 & 0.150 $\pm$ 0.042 \\
                                 Qwen3-8B & 3.824 $\pm$ 0.055 &  2.908 $\pm$ 0.115 & 0.157 $\pm$ 0.030 \\
                              Qwen3.5-27B & 3.071 $\pm$ 0.110 &  1.903 $\pm$ 0.159 & 0.253 $\pm$ 0.017 \\
                               Qwen3.5-9B & 3.145 $\pm$ 0.166 &  2.012 $\pm$ 0.186 & 0.221 $\pm$ 0.029 \\
\bottomrule
\end{tabular}
\end{footnotesize}
\end{table}

\begin{table}[htbp]
\centering
\caption{Mean $\pm$ standard deviation of Pearson $r$, Kendall $\tau$, QWK, and MAE across individual CoT sampling attempts per problem. Thinking-enabled and non-thinking model variants are denoted by \textcolor{green}{T} and \textcolor{red}{T}, respectively. Results are evaluated on the human-labelled \humands dataset without access to a standard markscheme and golden solution.}
\label{tab:human_blind_cot_per_attempt_summary_part1}
\begin{footnotesize}
\begin{tabular}{lrrrr}
\toprule
                                    Model &       Pearson $r$ &    Kendall $\tau$ &               QWK &               MAE \\
\midrule
   Claude-Sonnet-4.6 (\textcolor{red}{T}) & 0.379 $\pm$ 0.043 & 0.293 $\pm$ 0.061 & 0.349 $\pm$ 0.049 & 1.797 $\pm$ 0.057 \\
 Claude-Sonnet-4.6 (\textcolor{green}{T}) & 0.584 $\pm$ 0.100 & 0.481 $\pm$ 0.131 & 0.497 $\pm$ 0.100 & 1.646 $\pm$ 0.188 \\
       DeepSeek-V3.2 (\textcolor{red}{T}) & 0.151 $\pm$ 0.115 & 0.089 $\pm$ 0.079 & 0.053 $\pm$ 0.045 & 3.012 $\pm$ 0.097 \\
     DeepSeek-V3.2 (\textcolor{green}{T}) & 0.514 $\pm$ 0.072 & 0.399 $\pm$ 0.064 & 0.456 $\pm$ 0.069 & 1.699 $\pm$ 0.111 \\
             deepseek-v4-flash-nothinking & 0.164 $\pm$ 0.125 & 0.112 $\pm$ 0.124 & 0.071 $\pm$ 0.058 & 2.776 $\pm$ 0.036 \\
 DeepSeek-V4-Flash (\textcolor{green}{T}) & 0.468 $\pm$ 0.072 & 0.361 $\pm$ 0.054 & 0.396 $\pm$ 0.058 & 2.146 $\pm$ 0.143 \\
                         Gemini-2.5-Flash & 0.339 $\pm$ 0.147 & 0.265 $\pm$ 0.119 & 0.336 $\pm$ 0.149 & 1.913 $\pm$ 0.196 \\
Gemini-3-Pro-Prev. (\textcolor{green}{T}) & 0.464 $\pm$ 0.033 & 0.376 $\pm$ 0.033 & 0.369 $\pm$ 0.049 & 2.057 $\pm$ 0.095 \\
                                   GPT-4o & 0.063 $\pm$ 0.069 & 0.031 $\pm$ 0.057 & 0.040 $\pm$ 0.043 & 2.138 $\pm$ 0.055 \\
                                    GPT-5 & 0.475 $\pm$ 0.051 & 0.351 $\pm$ 0.067 & 0.435 $\pm$ 0.054 & 1.663 $\pm$ 0.110 \\
                               GPT-5-Mini & 0.458 $\pm$ 0.065 & 0.337 $\pm$ 0.063 & 0.455 $\pm$ 0.066 & 1.770 $\pm$ 0.114 \\
                                Kimi-K2.5 & 0.434 $\pm$ 0.070 & 0.319 $\pm$ 0.066 & 0.372 $\pm$ 0.079 & 1.879 $\pm$ 0.134 \\
                                 Qwen3-8B & 0.196 $\pm$ 0.119 & 0.154 $\pm$ 0.093 & 0.175 $\pm$ 0.107 & 2.442 $\pm$ 0.173 \\
                              Qwen3.5-27B & 0.454 $\pm$ 0.022 & 0.352 $\pm$ 0.030 & 0.448 $\pm$ 0.023 & 1.703 $\pm$ 0.107 \\
\bottomrule
\end{tabular}
\end{footnotesize}
\end{table}

\begin{table}[htbp]
\centering
\caption{Mean $\pm$ standard deviation of RMSE, bias, and exact-match accuracy across individual CoT sampling attempts per problem. Thinking-enabled and non-thinking model variants are denoted by \textcolor{green}{T} and \textcolor{red}{T}, respectively. Results are evaluated on the human-labelled \humands dataset without access to a standard markscheme and golden solution.}
\label{tab:human_blind_cot_per_attempt_summary_part2}
\begin{footnotesize}
\begin{tabular}{lrrr}
\toprule
                                    Model &              RMSE &               Bias &             Exact \\
\midrule
   Claude-Sonnet-4.6 (\textcolor{red}{T}) & 2.292 $\pm$ 0.052 & -0.422 $\pm$ 0.086 & 0.214 $\pm$ 0.046 \\
 Claude-Sonnet-4.6 (\textcolor{green}{T}) & 2.006 $\pm$ 0.238 & -0.609 $\pm$ 0.437 & 0.146 $\pm$ 0.055 \\
       DeepSeek-V3.2 (\textcolor{red}{T}) & 3.683 $\pm$ 0.107 & -2.832 $\pm$ 0.103 & 0.133 $\pm$ 0.020 \\
     DeepSeek-V3.2 (\textcolor{green}{T}) & 2.243 $\pm$ 0.128 & -0.871 $\pm$ 0.139 & 0.227 $\pm$ 0.048 \\
             deepseek-v4-flash-nothinking & 3.476 $\pm$ 0.067 & -2.505 $\pm$ 0.075 & 0.156 $\pm$ 0.036 \\
 DeepSeek-V4-Flash (\textcolor{green}{T}) & 2.798 $\pm$ 0.151 & -1.406 $\pm$ 0.027 & 0.229 $\pm$ 0.017 \\
                         Gemini-2.5-Flash & 2.501 $\pm$ 0.254 & -0.028 $\pm$ 0.206 & 0.204 $\pm$ 0.019 \\
Gemini-3-Pro-Prev. (\textcolor{green}{T}) & 2.772 $\pm$ 0.106 &  1.568 $\pm$ 0.180 & 0.240 $\pm$ 0.027 \\
                                   GPT-4o & 2.544 $\pm$ 0.053 &  0.802 $\pm$ 0.065 & 0.098 $\pm$ 0.017 \\
                                    GPT-5 & 2.199 $\pm$ 0.109 & -0.598 $\pm$ 0.161 & 0.211 $\pm$ 0.054 \\
                               GPT-5-Mini & 2.327 $\pm$ 0.134 & -0.136 $\pm$ 0.136 & 0.204 $\pm$ 0.042 \\
                                Kimi-K2.5 & 2.415 $\pm$ 0.107 & -1.001 $\pm$ 0.177 & 0.182 $\pm$ 0.054 \\
                                 Qwen3-8B & 3.143 $\pm$ 0.167 &  1.044 $\pm$ 0.291 & 0.196 $\pm$ 0.046 \\
                              Qwen3.5-27B & 2.284 $\pm$ 0.081 & -0.151 $\pm$ 0.188 & 0.266 $\pm$ 0.046 \\
\bottomrule
\end{tabular}
\end{footnotesize}
\end{table}

\paragraph{Correlation and error patterns on \nospacehumands.}
Across both Full and Blind templates, the strongest graders on \nospacehumands\ are the thinking-enabled variants of frontier models. Under the Full setting, DeepSeek-V4-Flash (Think) leads on most aggregate metrics with mean Pearson $r$ of $0.66$, mean QWK of $0.63$, and mean MAE of $1.38$ across attempts, closely followed by GPT-5 ($0.62$ Pearson, $0.56$ QWK) and DeepSeek-V3.2 (Think) ($0.62$ Pearson, $0.61$ QWK). The non-thinking variants of the same model families lose between $0.05$ and $0.20$ QWK and pick up systematic over-grading bias, often exceeding $+1.0$ points. Smaller open-weight graders such as Qwen3-8B, Prometheus-V2-7B, and Prometheus-V2-8$\times$7B, together with GPT-4o, collapse to near-zero or negative correlation with the human reference, with MAE close to or exceeding $3.0$, which is the level of a constant "award full marks" baseline.

\paragraph{Effect of aggregation.}
Mean-CoT@16 and MajorityVote@16 produce small but consistent gains over a single CoT call. Gemini-2.5-Flash improves from QWK $0.59$ at CoT@1 to $0.68$ at MeanCoT@16 in the Full setting, and DeepSeek-V4-Flash (Think) improves from $0.64$ to $0.69$. MajorityVote@16 produces similar gains except on graders whose all samples agree, in which case voting cannot denoise a deterministic bias (a pattern we revisit in the case-study section). The Blind setting shows the same ordering of graders as Full but with a uniform downward shift in correlation, and a sign flip in bias for several frontier models. DeepSeek-V3.2 (Think), for example, moves from $+0.27$ Full to $-0.88$ Blind, which reflects the absence of the rubric anchor.

\paragraph{Smaller graders behave differently on \nospacehumands.}
A non-trivial ranking shift appears for the small graders. Qwen3-8B, GPT-5 Mini, and Kimi-K2.5 reach better QWK and lower MAE under the Blind template than under the Full template. We discuss this counter-intuitive effect in detail in Section~\ref{app:case_blind_vs_full}. The short version is that the rubric in the Full template acts as a surface-form anchor that biases small graders toward awarding most of the available points on each rubric line.


\begin{table}[htbp]
\caption{Performance of language models across multiple correlation and error metrics. Thinking-enabled and non-thinking model variants are denoted by \textcolor{green}{T} and \textcolor{red}{T}, respectively. Results are evaluated on the human-labelled \humands dataset against a standard markscheme and golden solution.}
\label{tab:human_full_cot}
\begin{tabular}{lrrrrrrr}
\toprule
Model & Pearson $r$ & Kendall $\tau$ & QWK & MAE & RMSE & Bias & WTA$_{\leq 1}$ \\
\midrule
Claude-Sonnet-4.6 (\textcolor{red}{T}) & 0.54 & 0.44 & 0.44 & 1.79 & 2.43 & 1.33 & 0.56 \\
Claude-Sonnet-4.6 (\textcolor{green}{T}) & 0.55 & 0.45 & 0.47 & 1.73 & 2.40 & 1.24 & 0.58 \\
DeepSeek-V3.2 (\textcolor{red}{T}) & 0.43 & 0.35 & 0.29 & 2.27 & 3.04 & 2.02 & 0.48 \\
DeepSeek-V3.2 (\textcolor{green}{T}) & 0.60 & 0.53 & 0.59 & 1.35 & 2.26 & 0.27 & 0.71 \\
DeepSeek-V4-Flash (\textcolor{green}{T}) & 0.66 & 0.54 & 0.64 & 1.33 & 2.08 & -0.50 & 0.69 \\
Gemini-2.5-Flash & 0.62 & 0.50 & 0.59 & 1.54 & 2.23 & 0.63 & 0.58 \\
Gemini-2.5-Pro (\textcolor{red}{T}) & 0.60 & 0.49 & 0.52 & 1.52 & 2.20 & 1.06 & 0.65 \\
Gemini-2.5-Pro (\textcolor{green}{T}) & 0.57 & 0.45 & 0.49 & 1.54 & 2.20 & 1.04 & 0.69 \\
Gemini-3-Pro-Prev. (\textcolor{green}{T}) & 0.54 & 0.44 & 0.45 & 1.65 & 2.50 & 1.35 & 0.65 \\
GPT-4o & 0.20 & 0.17 & 0.06 & 3.10 & 3.83 & 3.06 & 0.31 \\
GPT-5 & 0.57 & 0.47 & 0.51 & 1.56 & 2.22 & 0.98 & 0.65 \\
GPT-5-Mini & 0.63 & 0.52 & 0.53 & 1.67 & 2.25 & 1.25 & 0.54 \\
Kimi-K2.5 & 0.33 & 0.22 & 0.20 & 2.37 & 3.29 & 2.24 & 0.50 \\
Prometheus-V2-7B & -0.09 & -0.12 & -0.07 & 2.92 & 3.66 & 1.71 & 0.35 \\
Prometheus-V2-8$\times$7B & 0.12 & 0.03 & 0.06 & 3.00 & 3.72 & 2.63 & 0.38 \\
Qwen3-8B & 0.15 & 0.16 & 0.07 & 3.04 & 3.83 & 2.83 & 0.34 \\
Qwen3.5-27B & 0.43 & 0.36 & 0.33 & 2.19 & 2.99 & 1.73 & 0.50 \\
Qwen3.5-9B & 0.30 & 0.26 & 0.20 & 2.58 & 3.37 & 2.16 & 0.39 \\
\bottomrule
\end{tabular}
\end{table}

\begin{table}[htbp]
\caption{Performance of language models across multiple correlation and error metrics. Thinking-enabled and non-thinking model variants are denoted by \textcolor{green}{T} and \textcolor{red}{T}, respectively. Results are evaluated on the human-labelled \humands dataset against a standard markscheme and golden solution.}
\label{tab:human_full_mean_cot}
\begin{tabular}{lrrrrrrr}
\toprule
Model & Pearson $r$ & Kendall $\tau$ & QWK & MAE & RMSE & Bias & WTA$_{\leq 1}$ \\
\midrule
Claude-Sonnet-4.6 (\textcolor{red}{T}) & 0.57 & 0.47 & 0.46 & 1.69 & 2.35 & 1.31 & 0.60 \\
Claude-Sonnet-4.6 (\textcolor{green}{T}) & 0.65 & 0.53 & 0.55 & 1.54 & 2.13 & 1.17 & 0.60 \\
DeepSeek-V3.2 (\textcolor{red}{T}) & 0.44 & 0.34 & 0.29 & 2.29 & 3.06 & 2.08 & 0.48 \\
DeepSeek-V3.2 (\textcolor{green}{T}) & 0.65 & 0.53 & 0.65 & 1.35 & 1.93 & 0.23 & 0.67 \\
DeepSeek-V4-Flash (\textcolor{green}{T}) & 0.72 & 0.59 & 0.69 & 1.25 & 1.87 & -0.67 & 0.69 \\
Gemini-2.5-Flash & 0.69 & 0.57 & 0.68 & 1.23 & 1.76 & 0.44 & 0.73 \\
Gemini-2.5-Pro (\textcolor{red}{T}) & 0.59 & 0.48 & 0.48 & 1.52 & 2.24 & 1.23 & 0.69 \\
Gemini-2.5-Pro (\textcolor{green}{T}) & 0.64 & 0.53 & 0.54 & 1.42 & 2.07 & 1.08 & 0.67 \\
Gemini-3-Pro-Prev. (\textcolor{green}{T}) & 0.59 & 0.48 & 0.49 & 1.58 & 2.37 & 1.33 & 0.65 \\
GPT-4o & 0.20 & 0.17 & 0.06 & 3.10 & 3.83 & 3.06 & 0.31 \\
GPT-5 & 0.67 & 0.57 & 0.59 & 1.44 & 2.01 & 0.98 & 0.69 \\
GPT-5-Mini & 0.55 & 0.46 & 0.43 & 1.88 & 2.52 & 1.50 & 0.52 \\
Kimi-K2.5 & 0.48 & 0.36 & 0.31 & 2.13 & 2.83 & 1.92 & 0.46 \\
Prometheus-V2-7B & 0.09 & 0.06 & 0.04 & 2.48 & 3.07 & 1.94 & 0.42 \\
Prometheus-V2-8$\times$7B & 0.13 & 0.17 & 0.05 & 2.73 & 3.42 & 2.48 & 0.33 \\
Qwen3-8B & 0.20 & 0.21 & 0.08 & 3.08 & 3.78 & 2.92 & 0.29 \\
Qwen3.5-27B & 0.44 & 0.37 & 0.32 & 2.17 & 2.93 & 1.83 & 0.48 \\
Qwen3.5-9B & 0.45 & 0.36 & 0.31 & 2.27 & 3.05 & 2.02 & 0.50 \\
\bottomrule
\end{tabular}
\end{table}

\begin{table}[htbp]
\caption{Performance of language models across multiple correlation and error metrics. Thinking-enabled and non-thinking model variants are denoted by \textcolor{green}{T} and \textcolor{red}{T}, respectively. Results are evaluated on the human-labelled \humands dataset against a standard markscheme and golden solution.}
\label{tab:human_full_majority_vote}
\begin{tabular}{lrrrrrrr}
\toprule
Model & Pearson $r$ & Kendall $\tau$ & QWK & MAE & RMSE & Bias & WTA$_{\leq 1}$ \\
\midrule
Claude-Sonnet-4.6 (\textcolor{red}{T}) & 0.58 & 0.47 & 0.46 & 1.75 & 2.37 & 1.38 & 0.58 \\
Claude-Sonnet-4.6 (\textcolor{green}{T}) & 0.61 & 0.53 & 0.52 & 1.65 & 2.25 & 1.23 & 0.56 \\
DeepSeek-V3.2 (\textcolor{red}{T}) & 0.44 & 0.34 & 0.29 & 2.29 & 3.08 & 2.08 & 0.48 \\
DeepSeek-V3.2 (\textcolor{green}{T}) & 0.66 & 0.57 & 0.65 & 1.27 & 2.05 & 0.10 & 0.67 \\
DeepSeek-V4-Flash (\textcolor{green}{T}) & 0.65 & 0.54 & 0.63 & 1.38 & 2.13 & -0.54 & 0.65 \\
Gemini-2.5-Flash & 0.67 & 0.57 & 0.64 & 1.38 & 2.14 & 0.63 & 0.69 \\
Gemini-2.5-Pro (\textcolor{red}{T}) & 0.54 & 0.44 & 0.45 & 1.58 & 2.43 & 1.29 & 0.63 \\
Gemini-2.5-Pro (\textcolor{green}{T}) & 0.63 & 0.51 & 0.55 & 1.42 & 2.16 & 1.13 & 0.71 \\
Gemini-3-Pro-Prev. (\textcolor{green}{T}) & 0.54 & 0.44 & 0.45 & 1.67 & 2.50 & 1.38 & 0.65 \\
GPT-4o & 0.20 & 0.17 & 0.06 & 3.10 & 3.83 & 3.06 & 0.31 \\
GPT-5 & 0.68 & 0.59 & 0.62 & 1.44 & 2.01 & 0.98 & 0.65 \\
GPT-5-Mini & 0.57 & 0.49 & 0.48 & 1.81 & 2.49 & 1.35 & 0.56 \\
Kimi-K2.5 & 0.45 & 0.35 & 0.28 & 2.27 & 3.05 & 2.15 & 0.46 \\
Prometheus-V2-7B & 0.13 & 0.07 & 0.05 & 3.23 & 4.00 & 3.10 & 0.31 \\
Prometheus-V2-8$\times$7B & 0.11 & 0.07 & 0.04 & 3.19 & 3.85 & 2.98 & 0.25 \\
Qwen3-8B & 0.18 & 0.23 & 0.06 & 3.21 & 3.88 & 3.04 & 0.27 \\
Qwen3.5-27B & 0.40 & 0.33 & 0.30 & 2.23 & 3.04 & 1.81 & 0.50 \\
Qwen3.5-9B & 0.37 & 0.30 & 0.26 & 2.46 & 3.29 & 2.04 & 0.44 \\
\bottomrule
\end{tabular}
\end{table}

\begin{table}[htbp]
\caption{Performance of language models across multiple correlation and error metrics. Thinking-enabled and non-thinking model variants are denoted by \textcolor{green}{T} and \textcolor{red}{T}, respectively. Results are evaluated on the human-labelled \humands dataset without access to a standard markscheme and golden solution.}
\label{tab:human_blind_cot}
\begin{tabular}{lrrrrrrr}
\toprule
Model & Pearson $r$ & Kendall $\tau$ & QWK & MAE & RMSE & Bias & WTA$_{\leq 1}$ \\
\midrule
Claude-Sonnet-4.6 (\textcolor{red}{T}) & 0.40 & 0.32 & 0.38 & 1.77 & 2.24 & -0.31 & 0.44 \\
Claude-Sonnet-4.6 (\textcolor{green}{T}) & 0.49 & 0.34 & 0.46 & 1.60 & 1.97 & 0.16 & 0.56 \\
DeepSeek-V3.2 (\textcolor{red}{T}) & 0.15 & 0.04 & 0.06 & 2.96 & 3.64 & -2.71 & 0.29 \\
DeepSeek-V3.2 (\textcolor{green}{T}) & 0.57 & 0.50 & 0.51 & 1.54 & 2.12 & -0.88 & 0.56 \\
DeepSeek-V4-Flash (\textcolor{green}{T}) & 0.40 & 0.31 & 0.34 & 2.25 & 2.94 & -1.38 & 0.46 \\
Gemini-2.5-Flash & 0.42 & 0.32 & 0.43 & 1.86 & 2.39 & -0.01 & 0.46 \\
Gemini-3-Pro-Prev. (\textcolor{green}{T}) & 0.48 & 0.39 & 0.40 & 2.00 & 2.73 & 1.46 & 0.52 \\
GPT-4o & 0.06 & 0.00 & 0.04 & 2.17 & 2.55 & 0.79 & 0.40 \\
GPT-5 & 0.45 & 0.32 & 0.39 & 1.71 & 2.32 & -0.87 & 0.56 \\
GPT-5-Mini & 0.37 & 0.23 & 0.37 & 1.92 & 2.49 & -0.29 & 0.44 \\
Kimi-K2.5 & 0.41 & 0.31 & 0.33 & 1.92 & 2.35 & -0.92 & 0.48 \\
Qwen3-8B & 0.28 & 0.21 & 0.26 & 2.35 & 3.13 & 0.91 & 0.44 \\
Qwen3.5-27B & 0.44 & 0.34 & 0.44 & 1.69 & 2.31 & -0.31 & 0.52 \\
\bottomrule
\end{tabular}
\end{table}

\begin{table}[htbp]
\caption{Performance of language models across multiple correlation and error metrics. Thinking-enabled and non-thinking model variants are denoted by \textcolor{green}{T} and \textcolor{red}{T}, respectively. Results are evaluated on the human-labelled \humands dataset without access to a standard markscheme and golden solution.}
\label{tab:human_blind_mean_cot}
\begin{tabular}{lrrrrrrr}
\toprule
Model & Pearson $r$ & Kendall $\tau$ & QWK & MAE & RMSE & Bias & WTA$_{\leq 1}$ \\
\midrule
Claude-Sonnet-4.6 (\textcolor{red}{T}) & 0.39 & 0.31 & 0.35 & 1.77 & 2.22 & -0.40 & 0.44 \\
Claude-Sonnet-4.6 (\textcolor{green}{T}) & 0.48 & 0.36 & 0.40 & 1.63 & 2.03 & -0.21 & 0.50 \\
DeepSeek-V3.2 (\textcolor{red}{T}) & 0.20 & 0.12 & 0.06 & 3.00 & 3.63 & -2.83 & 0.31 \\
DeepSeek-V3.2 (\textcolor{green}{T}) & 0.61 & 0.48 & 0.52 & 1.52 & 2.03 & -0.85 & 0.60 \\
DeepSeek-V4-Flash (\textcolor{green}{T}) & 0.51 & 0.39 & 0.42 & 2.04 & 2.67 & -1.46 & 0.50 \\
Gemini-2.5-Flash & 0.44 & 0.33 & 0.43 & 1.75 & 2.29 & -0.25 & 0.54 \\
Gemini-3-Pro-Prev. (\textcolor{green}{T}) & 0.50 & 0.42 & 0.39 & 2.06 & 2.63 & 1.56 & 0.46 \\
GPT-4o & 0.05 & 0.01 & 0.03 & 2.10 & 2.54 & 0.81 & 0.44 \\
GPT-5 & 0.53 & 0.40 & 0.48 & 1.58 & 2.08 & -0.63 & 0.56 \\
GPT-5-Mini & 0.53 & 0.39 & 0.53 & 1.63 & 2.07 & -0.08 & 0.54 \\
Kimi-K2.5 & 0.54 & 0.44 & 0.43 & 1.71 & 2.14 & -0.92 & 0.48 \\
Qwen3-8B & 0.33 & 0.25 & 0.28 & 2.08 & 2.58 & 1.04 & 0.46 \\
Qwen3.5-27B & 0.49 & 0.38 & 0.48 & 1.56 & 2.12 & -0.10 & 0.56 \\
\bottomrule
\end{tabular}
\end{table}

\begin{table}[htbp]
\caption{Performance of language models across multiple correlation and error metrics. Thinking-enabled and non-thinking model variants are denoted by \textcolor{green}{T} and \textcolor{red}{T}, respectively. Results are evaluated on the human-labelled \humands dataset without access to a standard markscheme and golden solution.}
\label{tab:human_blind_majority_vote}
\begin{tabular}{lrrrrrrr}
\toprule
Model & Pearson $r$ & Kendall $\tau$ & QWK & MAE & RMSE & Bias & WTA$_{\leq 1}$ \\
\midrule
Claude-Sonnet-4.6 (\textcolor{red}{T}) & 0.46 & 0.38 & 0.42 & 1.67 & 2.15 & -0.46 & 0.48 \\
Claude-Sonnet-4.6 (\textcolor{green}{T}) & 0.47 & 0.35 & 0.40 & 1.63 & 2.03 & -0.13 & 0.54 \\
DeepSeek-V3.2 (\textcolor{red}{T}) & 0.09 & 0.06 & 0.02 & 3.15 & 3.83 & -3.02 & 0.29 \\
DeepSeek-V3.2 (\textcolor{green}{T}) & 0.59 & 0.48 & 0.49 & 1.58 & 2.18 & -1.08 & 0.58 \\
DeepSeek-V4-Flash (\textcolor{green}{T}) & 0.49 & 0.39 & 0.42 & 2.10 & 2.74 & -1.35 & 0.48 \\
Gemini-2.5-Flash & 0.38 & 0.29 & 0.38 & 1.85 & 2.56 & -0.06 & 0.52 \\
Gemini-3-Pro-Prev. (\textcolor{green}{T}) & 0.42 & 0.35 & 0.33 & 2.17 & 2.86 & 1.58 & 0.46 \\
GPT-4o & 0.05 & 0.01 & 0.03 & 2.08 & 2.52 & 0.79 & 0.46 \\
GPT-5 & 0.50 & 0.38 & 0.47 & 1.60 & 2.17 & -0.56 & 0.54 \\
GPT-5-Mini & 0.51 & 0.38 & 0.51 & 1.63 & 2.24 & -0.25 & 0.56 \\
Kimi-K2.5 & 0.50 & 0.41 & 0.42 & 1.79 & 2.29 & -1.00 & 0.50 \\
Qwen3-8B & 0.29 & 0.25 & 0.25 & 2.40 & 3.16 & 1.27 & 0.44 \\
Qwen3.5-27B & 0.46 & 0.36 & 0.46 & 1.65 & 2.26 & -0.31 & 0.56 \\
\bottomrule
\end{tabular}
\end{table}

\clearpage
\subsection{\synthds}

\paragraph{Correlation and error patterns on \nospacesynthds.}
The synthetic benchmark is constructed so that every candidate solution has a target rubric-aligned grade, which makes the Full setting easier than on \nospacehumands. Under the Full template, Claude-Sonnet-4.6 (Think) achieves Pearson $r$ of $0.81$ and QWK of $0.75$ at the aggregated level; DeepSeek-V3.2 (Think) reaches $0.81$ Pearson and $0.77$ QWK with MeanCoT@16; and GPT-5 reaches $0.77$ Pearson and $0.69$ QWK. Both Mean-CoT and MajorityVote lift aggregate quality by a few points relative to a single CoT call. MAE values cluster around or below $1.0$ for the strongest graders, which means the synthetic distribution is concentrated where rubric-following alone is enough to recover the target grade.

\paragraph{Failure modes carry over from \nospacehumands.}
Two of the failure modes we observe on \nospacehumands\ also appear on the synthetic data. First, GPT-4o has near-zero QWK ($0.12$) despite a non-trivial Pearson correlation ($0.53$) and a bias of $+3.37$, which is the signature of a grader that ranks proofs in roughly the right order but inflates every grade to the maximum. Second, the lower-tier graders (Kimi-K2.5, Qwen3.5-27B, GPT-5 Mini) carry a positive bias of about $+1.6$ to $+2.2$, again driven by over-crediting partially correct rubric items. Under the Blind template, the systematic over-grading shrinks because the rubric anchor is removed, and most graders move to a small positive or near-zero bias under MajorityVote@16.

\paragraph{Cross-set consistency.}
The relative ordering of graders is largely preserved between \nospacehumands\ and \nospacesynthds. The top three graders on the Full template are the same on both datasets, and the bottom three are also the same. The absolute values are higher on the synthetic set because the candidate distribution has lower per-rubric-item ambiguity by construction, but the directional conclusions match. This cross-set consistency is the basis for our claim in the main text that fine-tuning on the synthetic set transfers usefully to the human set.

\clearpage

\begin{table}[htbp]
\centering
\caption{Mean $\pm$ standard deviation of Pearson $r$, Kendall $\tau$, QWK, and MAE across individual CoT sampling attempts per problem. Thinking-enabled and non-thinking model variants are denoted by \textcolor{green}{T} and \textcolor{red}{T}, respectively. Results are evaluated on the synthetically-generated dataset, \nospacesynthds, against a standard markscheme and golden solution.}
\label{tab:synthetic_200_full_cot_per_attempt_summary_part1}
\begin{footnotesize}
\begin{tabular}{lrrrr}
\toprule
                                   Model &       Pearson $r$ &    Kendall $\tau$ &               QWK &               MAE \\
\midrule
Claude-Sonnet-4.6 (\textcolor{green}{T}) & 0.772 $\pm$ 0.067 & 0.677 $\pm$ 0.079 & 0.713 $\pm$ 0.075 & 0.843 $\pm$ 0.157 \\
      DeepSeek-V3.2 (\textcolor{red}{T}) & 0.729 $\pm$ 0.007 & 0.630 $\pm$ 0.007 & 0.423 $\pm$ 0.007 & 2.008 $\pm$ 0.018 \\
    DeepSeek-V3.2 (\textcolor{green}{T}) & 0.768 $\pm$ 0.019 & 0.652 $\pm$ 0.021 & 0.724 $\pm$ 0.019 & 1.020 $\pm$ 0.039 \\
DeepSeek-V4-Flash (\textcolor{green}{T}) & 0.587 $\pm$ 0.042 & 0.488 $\pm$ 0.037 & 0.533 $\pm$ 0.036 & 1.380 $\pm$ 0.051 \\
                                  GPT-4o & 0.527 $\pm$ 0.007 & 0.454 $\pm$ 0.004 & 0.120 $\pm$ 0.003 & 3.371 $\pm$ 0.009 \\
                                   GPT-5 & 0.745 $\pm$ 0.018 & 0.626 $\pm$ 0.017 & 0.658 $\pm$ 0.012 & 1.121 $\pm$ 0.039 \\
                              GPT-5-Mini & 0.657 $\pm$ 0.008 & 0.519 $\pm$ 0.007 & 0.445 $\pm$ 0.004 & 1.779 $\pm$ 0.018 \\
                               Kimi-K2.5 & 0.692 $\pm$ 0.026 & 0.572 $\pm$ 0.024 & 0.363 $\pm$ 0.018 & 2.246 $\pm$ 0.033 \\
                             Qwen3.5-27B & 0.680 $\pm$ 0.011 & 0.572 $\pm$ 0.016 & 0.395 $\pm$ 0.006 & 2.183 $\pm$ 0.025 \\
                              Qwen3.5-9B & 0.732 $\pm$ 0.017 & 0.604 $\pm$ 0.019 & 0.528 $\pm$ 0.014 & 1.731 $\pm$ 0.055 \\
\bottomrule
\end{tabular}
\end{footnotesize}
\end{table}

\begin{table}[htbp]
\centering
\caption{Mean $\pm$ standard deviation of RMSE, bias, and exact-match accuracy across individual CoT sampling attempts per problem. Thinking-enabled and non-thinking model variants are denoted by \textcolor{green}{T} and \textcolor{red}{T}, respectively. Results are evaluated on the synthetically-generated dataset, \nospacesynthds, against a standard markscheme and golden solution.}
\label{tab:synthetic_200_full_cot_per_attempt_summary_part2}
\begin{footnotesize}
\begin{tabular}{lrrr}
\toprule
                                   Model &              RMSE &               Bias &             Exact \\
\midrule
Claude-Sonnet-4.6 (\textcolor{green}{T}) & 1.256 $\pm$ 0.189 &  0.609 $\pm$ 0.093 & 0.436 $\pm$ 0.080 \\
      DeepSeek-V3.2 (\textcolor{red}{T}) & 2.337 $\pm$ 0.020 &  1.978 $\pm$ 0.018 & 0.080 $\pm$ 0.005 \\
    DeepSeek-V3.2 (\textcolor{green}{T}) & 1.398 $\pm$ 0.052 &  0.093 $\pm$ 0.037 & 0.312 $\pm$ 0.016 \\
DeepSeek-V4-Flash (\textcolor{green}{T}) & 1.848 $\pm$ 0.068 & -0.530 $\pm$ 0.041 & 0.237 $\pm$ 0.006 \\
                                  GPT-4o & 3.628 $\pm$ 0.010 &  3.371 $\pm$ 0.009 & 0.005 $\pm$ 0.000 \\
                                   GPT-5 & 1.549 $\pm$ 0.022 &  0.769 $\pm$ 0.038 & 0.312 $\pm$ 0.019 \\
                              GPT-5-Mini & 2.278 $\pm$ 0.022 &  1.661 $\pm$ 0.042 & 0.195 $\pm$ 0.003 \\
                               Kimi-K2.5 & 2.592 $\pm$ 0.048 &  2.223 $\pm$ 0.036 & 0.074 $\pm$ 0.005 \\
                             Qwen3.5-27B & 2.611 $\pm$ 0.013 &  2.100 $\pm$ 0.014 & 0.126 $\pm$ 0.013 \\
                              Qwen3.5-9B & 2.242 $\pm$ 0.018 &  1.498 $\pm$ 0.030 & 0.210 $\pm$ 0.024 \\
\bottomrule
\end{tabular}
\end{footnotesize}
\end{table}

\begin{table}[htbp]
\centering
\caption{Mean $\pm$ standard deviation of Pearson $r$, Kendall $\tau$, QWK, and MAE across individual CoT sampling attempts per problem. Thinking-enabled and non-thinking model variants are denoted by \textcolor{green}{T} and \textcolor{red}{T}, respectively. Results are evaluated on the synthetically-generated dataset, \nospacesynthds, without access to a standard markscheme and golden solution.}
\label{tab:synthetic_200_blind_cot_per_attempt_summary_part1}
\begin{footnotesize}
\begin{tabular}{lrrrr}
\toprule
                                   Model &       Pearson $r$ &    Kendall $\tau$ &               QWK &               MAE \\
\midrule
Claude-Sonnet-4.6 (\textcolor{green}{T}) & 0.507 $\pm$ 0.067 & 0.411 $\pm$ 0.058 & 0.493 $\pm$ 0.058 & 1.188 $\pm$ 0.066 \\
DeepSeek-V4-Flash (\textcolor{green}{T}) & 0.464 $\pm$ 0.044 & 0.367 $\pm$ 0.044 & 0.432 $\pm$ 0.041 & 1.561 $\pm$ 0.054 \\
                                  GPT-4o & 0.427 $\pm$ 0.005 & 0.368 $\pm$ 0.015 & 0.120 $\pm$ 0.003 & 3.034 $\pm$ 0.010 \\
                                   GPT-5 & 0.490 $\pm$ 0.024 & 0.397 $\pm$ 0.031 & 0.459 $\pm$ 0.025 & 1.341 $\pm$ 0.079 \\
                              GPT-5-Mini & 0.503 $\pm$ 0.013 & 0.394 $\pm$ 0.014 & 0.423 $\pm$ 0.012 & 1.538 $\pm$ 0.029 \\
                               Kimi-K2.5 & 0.452 $\pm$ 0.035 & 0.352 $\pm$ 0.018 & 0.391 $\pm$ 0.031 & 1.590 $\pm$ 0.052 \\
                             Qwen3.5-27B & 0.379 $\pm$ 0.021 & 0.245 $\pm$ 0.021 & 0.346 $\pm$ 0.020 & 1.787 $\pm$ 0.037 \\
                              Qwen3.5-9B & 0.409 $\pm$ 0.090 & 0.336 $\pm$ 0.071 & 0.314 $\pm$ 0.070 & 1.900 $\pm$ 0.133 \\
\bottomrule
\end{tabular}
\end{footnotesize}
\end{table}

\begin{table}[htbp]
\centering
\caption{Mean $\pm$ standard deviation of RMSE, bias, and exact-match accuracy across individual CoT sampling attempts per problem. Thinking-enabled and non-thinking model variants are denoted by \textcolor{green}{T} and \textcolor{red}{T}, respectively. Results are evaluated on the synthetically-generated dataset, \nospacesynthds, without access to a standard markscheme and golden solution.}
\label{tab:synthetic_200_blind_cot_per_attempt_summary_part2}
\begin{footnotesize}
\begin{tabular}{lrrr}
\toprule
                                   Model &              RMSE &               Bias &             Exact \\
\midrule
Claude-Sonnet-4.6 (\textcolor{green}{T}) & 1.563 $\pm$ 0.065 &  0.267 $\pm$ 0.235 & 0.264 $\pm$ 0.029 \\
DeepSeek-V4-Flash (\textcolor{green}{T}) & 1.994 $\pm$ 0.082 & -0.374 $\pm$ 0.058 & 0.203 $\pm$ 0.037 \\
                                  GPT-4o & 3.332 $\pm$ 0.011 &  2.986 $\pm$ 0.013 & 0.031 $\pm$ 0.003 \\
                                   GPT-5 & 1.709 $\pm$ 0.067 &  0.576 $\pm$ 0.043 & 0.212 $\pm$ 0.033 \\
                              GPT-5-Mini & 1.943 $\pm$ 0.031 &  0.990 $\pm$ 0.045 & 0.193 $\pm$ 0.030 \\
                               Kimi-K2.5 & 1.978 $\pm$ 0.076 &  0.892 $\pm$ 0.016 & 0.175 $\pm$ 0.018 \\
                             Qwen3.5-27B & 2.288 $\pm$ 0.047 & -0.040 $\pm$ 0.024 & 0.196 $\pm$ 0.007 \\
                              Qwen3.5-9B & 2.336 $\pm$ 0.148 &  1.290 $\pm$ 0.083 & 0.139 $\pm$ 0.025 \\
\bottomrule
\end{tabular}
\end{footnotesize}
\end{table}

\begin{table}[htbp]
\caption{Performance of language models across multiple correlation and error metrics. Thinking-enabled and non-thinking model variants are denoted by \textcolor{green}{T} and \textcolor{red}{T}, respectively. Results are evaluated on the synthetically-generated dataset, \nospacesynthds, against a standard markscheme and golden solution.}
\label{tab:synthetic_200_full_cot}
\begin{tabular}{lrrrrrrr}
\toprule
Model & Pearson $r$ & Kendall $\tau$ & QWK & MAE & RMSE & Bias & WTA$_{\leq 1}$ \\
\midrule
Claude-Sonnet-4.6 (\textcolor{green}{T}) & 0.81 & 0.73 & 0.75 & 0.78 & 1.12 & 0.61 & 0.80 \\
DeepSeek-V3.2 (\textcolor{red}{T}) & 0.72 & 0.62 & 0.42 & 2.01 & 2.35 & 1.98 & 0.36 \\
DeepSeek-V3.2 (\textcolor{green}{T}) & 0.77 & 0.65 & 0.73 & 1.00 & 1.37 & 0.05 & 0.79 \\
DeepSeek-V4-Flash (\textcolor{green}{T}) & 0.57 & 0.47 & 0.52 & 1.43 & 1.89 & -0.48 & 0.63 \\
GPT-4o & 0.53 & 0.45 & 0.12 & 3.37 & 3.63 & 3.37 & 0.10 \\
GPT-5 & 0.73 & 0.61 & 0.64 & 1.18 & 1.58 & 0.78 & 0.67 \\
GPT-5-Mini & 0.65 & 0.51 & 0.44 & 1.79 & 2.30 & 1.67 & 0.52 \\
Kimi-K2.5 & 0.67 & 0.55 & 0.35 & 2.24 & 2.59 & 2.20 & 0.31 \\
Qwen3.5-27B & 0.69 & 0.59 & 0.40 & 2.19 & 2.60 & 2.10 & 0.35 \\
Qwen3.5-9B & 0.75 & 0.63 & 0.54 & 1.75 & 2.24 & 1.51 & 0.52 \\
\bottomrule
\end{tabular}
\end{table}

\begin{table}[htbp]
\caption{Performance of language models across multiple correlation and error metrics. Thinking-enabled and non-thinking model variants are denoted by \textcolor{green}{T} and \textcolor{red}{T}, respectively. Results are evaluated on the synthetically-generated dataset, \nospacesynthds, against a standard markscheme and golden solution.}
\label{tab:synthetic_200_full_mean_cot}
\begin{tabular}{lrrrrrrr}
\toprule
Model & Pearson $r$ & Kendall $\tau$ & QWK & MAE & RMSE & Bias & WTA$_{\leq 1}$ \\
\midrule
Claude-Sonnet-4.6 (\textcolor{green}{T}) & 0.74 & 0.65 & 0.66 & 0.86 & 1.36 & 0.70 & 0.78 \\
DeepSeek-V3.2 (\textcolor{red}{T}) & 0.73 & 0.63 & 0.42 & 2.01 & 2.32 & 1.98 & 0.35 \\
DeepSeek-V3.2 (\textcolor{green}{T}) & 0.81 & 0.69 & 0.77 & 0.92 & 1.23 & 0.05 & 0.83 \\
DeepSeek-V4-Flash (\textcolor{green}{T}) & 0.63 & 0.53 & 0.59 & 1.23 & 1.65 & -0.51 & 0.69 \\
GPT-4o & 0.53 & 0.46 & 0.12 & 3.37 & 3.63 & 3.37 & 0.10 \\
GPT-5 & 0.77 & 0.66 & 0.69 & 1.02 & 1.45 & 0.73 & 0.75 \\
GPT-5-Mini & 0.67 & 0.53 & 0.46 & 1.72 & 2.20 & 1.64 & 0.52 \\
Kimi-K2.5 & 0.70 & 0.58 & 0.36 & 2.25 & 2.58 & 2.23 & 0.31 \\
Qwen3.5-27B & 0.69 & 0.58 & 0.40 & 2.17 & 2.58 & 2.10 & 0.37 \\
Qwen3.5-9B & 0.73 & 0.60 & 0.52 & 1.70 & 2.18 & 1.49 & 0.52 \\
\bottomrule
\end{tabular}
\end{table}

\begin{table}[htbp]
\caption{Performance of language models across multiple correlation and error metrics. Thinking-enabled and non-thinking model variants are denoted by \textcolor{green}{T} and \textcolor{red}{T}, respectively. Results are evaluated on the synthetically-generated dataset, \nospacesynthds, against a standard markscheme and golden solution.}
\label{tab:synthetic_200_full_majority_vote}
\begin{tabular}{lrrrrrrr}
\toprule
Model & Pearson $r$ & Kendall $\tau$ & QWK & MAE & RMSE & Bias & WTA$_{\leq 1}$ \\
\midrule
Claude-Sonnet-4.6 (\textcolor{green}{T}) & 0.72 & 0.62 & 0.64 & 0.90 & 1.44 & 0.74 & 0.76 \\
DeepSeek-V3.2 (\textcolor{red}{T}) & 0.73 & 0.63 & 0.42 & 2.01 & 2.33 & 1.98 & 0.35 \\
DeepSeek-V3.2 (\textcolor{green}{T}) & 0.77 & 0.65 & 0.72 & 1.03 & 1.41 & 0.08 & 0.77 \\
DeepSeek-V4-Flash (\textcolor{green}{T}) & 0.59 & 0.49 & 0.54 & 1.36 & 1.83 & -0.50 & 0.66 \\
GPT-4o & 0.52 & 0.45 & 0.12 & 3.38 & 3.64 & 3.38 & 0.10 \\
GPT-5 & 0.77 & 0.65 & 0.67 & 1.09 & 1.52 & 0.79 & 0.73 \\
GPT-5-Mini & 0.66 & 0.52 & 0.45 & 1.78 & 2.29 & 1.67 & 0.52 \\
Kimi-K2.5 & 0.70 & 0.57 & 0.36 & 2.26 & 2.61 & 2.25 & 0.30 \\
Qwen3.5-27B & 0.69 & 0.58 & 0.40 & 2.14 & 2.56 & 2.07 & 0.37 \\
Qwen3.5-9B & 0.73 & 0.60 & 0.52 & 1.76 & 2.25 & 1.53 & 0.52 \\
\bottomrule
\end{tabular}
\end{table}

\begin{table}[htbp]
\caption{Performance of language models across multiple correlation and error metrics. Thinking-enabled and non-thinking model variants are denoted by \textcolor{green}{T} and \textcolor{red}{T}, respectively. Results are evaluated on the synthetically-generated dataset, \nospacesynthds, without access to a standard markscheme and golden solution.}
\label{tab:synthetic_200_blind_cot}
\begin{tabular}{lrrrrrrr}
\toprule
Model & Pearson $r$ & Kendall $\tau$ & QWK & MAE & RMSE & Bias & WTA$_{\leq 1}$ \\
\midrule
Claude-Sonnet-4.6 (\textcolor{green}{T}) & 0.49 & 0.37 & 0.46 & 1.16 & 1.55 & 0.48 & 0.71 \\
DeepSeek-V4-Flash (\textcolor{green}{T}) & 0.50 & 0.41 & 0.47 & 1.50 & 1.90 & -0.36 & 0.60 \\
GPT-4o & 0.43 & 0.36 & 0.12 & 3.03 & 3.32 & 2.98 & 0.14 \\
GPT-5 & 0.50 & 0.40 & 0.46 & 1.36 & 1.72 & 0.62 & 0.63 \\
GPT-5-Mini & 0.51 & 0.40 & 0.44 & 1.52 & 1.90 & 0.93 & 0.56 \\
Kimi-K2.5 & 0.41 & 0.33 & 0.36 & 1.64 & 2.05 & 0.91 & 0.52 \\
Qwen3.5-27B & 0.35 & 0.22 & 0.32 & 1.84 & 2.35 & -0.06 & 0.51 \\
Qwen3.5-9B & 0.29 & 0.24 & 0.22 & 2.06 & 2.50 & 1.26 & 0.40 \\
\bottomrule
\end{tabular}
\end{table}

\begin{table}[htbp]
\caption{Performance of language models across multiple correlation and error metrics. Thinking-enabled and non-thinking model variants are denoted by \textcolor{green}{T} and \textcolor{red}{T}, respectively. Results are evaluated on the synthetically-generated dataset, \nospacesynthds, without access to a standard markscheme and golden solution.}
\label{tab:synthetic_200_blind_mean_cot}
\begin{tabular}{lrrrrrrr}
\toprule
Model & Pearson $r$ & Kendall $\tau$ & QWK & MAE & RMSE & Bias & WTA$_{\leq 1}$ \\
\midrule
Claude-Sonnet-4.6 (\textcolor{green}{T}) & 0.50 & 0.41 & 0.50 & 1.18 & 1.55 & 0.09 & 0.70 \\
DeepSeek-V4-Flash (\textcolor{green}{T}) & 0.48 & 0.38 & 0.46 & 1.47 & 1.86 & -0.35 & 0.61 \\
GPT-4o & 0.43 & 0.38 & 0.12 & 3.04 & 3.33 & 2.99 & 0.14 \\
GPT-5 & 0.49 & 0.41 & 0.46 & 1.30 & 1.68 & 0.53 & 0.66 \\
GPT-5-Mini & 0.53 & 0.43 & 0.45 & 1.45 & 1.85 & 0.99 & 0.59 \\
Kimi-K2.5 & 0.52 & 0.41 & 0.44 & 1.45 & 1.78 & 0.91 & 0.58 \\
Qwen3.5-27B & 0.39 & 0.26 & 0.36 & 1.67 & 2.19 & -0.10 & 0.55 \\
Qwen3.5-9B & 0.46 & 0.38 & 0.36 & 1.76 & 2.15 & 1.25 & 0.46 \\
\bottomrule
\end{tabular}
\end{table}

\begin{table}[htbp]
\caption{Performance of language models across multiple correlation and error metrics. Thinking-enabled and non-thinking model variants are denoted by \textcolor{green}{T} and \textcolor{red}{T}, respectively. Results are evaluated on the synthetically-generated dataset, \nospacesynthds, without access to a standard markscheme and golden solution.}
\label{tab:synthetic_200_blind_majority_vote}
\begin{tabular}{lrrrrrrr}
\toprule
Model & Pearson $r$ & Kendall $\tau$ & QWK & MAE & RMSE & Bias & WTA$_{\leq 1}$ \\
\midrule
Claude-Sonnet-4.6 (\textcolor{green}{T}) & 0.47 & 0.37 & 0.47 & 1.25 & 1.63 & 0.07 & 0.67 \\
DeepSeek-V4-Flash (\textcolor{green}{T}) & 0.51 & 0.41 & 0.47 & 1.50 & 1.91 & -0.34 & 0.58 \\
GPT-4o & 0.42 & 0.36 & 0.12 & 3.06 & 3.35 & 3.01 & 0.14 \\
GPT-5 & 0.49 & 0.40 & 0.46 & 1.35 & 1.72 & 0.57 & 0.64 \\
GPT-5-Mini & 0.53 & 0.41 & 0.45 & 1.51 & 1.89 & 0.96 & 0.55 \\
Kimi-K2.5 & 0.49 & 0.39 & 0.44 & 1.54 & 1.89 & 0.80 & 0.55 \\
Qwen3.5-27B & 0.36 & 0.22 & 0.33 & 1.83 & 2.33 & -0.16 & 0.51 \\
Qwen3.5-9B & 0.40 & 0.33 & 0.31 & 1.92 & 2.36 & 1.25 & 0.44 \\
\bottomrule
\end{tabular}
\end{table}

\clearpage

\subsection{Additional evaluation of BoN methods}

The table below extends the main-text \method comparison to the synthetic benchmark. Table~\ref{tab:mab-synthetic-full} reports the same baselines (CoT@1, MajorityVote@16 with three templates, BoN@16 with four reward models) on \nospacesynthds\ Full, and confirms that \method continues to lead on QWK and RMSE for the three graders run in this setting. The same pattern we observe on \nospacehumands\ holds here: token-level sampling baselines (MajorityVote, BoN with off-the-shelf reward models) yield essentially flat improvements over a single CoT call, while the prompt-side hypothesis enumeration in \method is what produces the consistent gain across graders.

\begin{table*}[h]
    \centering
    \caption{Comparative analysis of model inference performance on synthetically generated data across QWK, RMSE, and Bias metrics for DeepSeek-V4-Flash, Kimi-K2.5, Gemini-3-Pro-Preview, and Qwen3.5-27B. Evaluation is performed \textit{with} a reference solution and a corresponding marking scheme.}
    \label{tab:mab-synthetic-full}
\begin{footnotesize}
\begin{tabular}{lc*{11}{c}}
  \toprule
  & \multicolumn{4}{c}{\textbf{QWK ($\uparrow$)}} & \multicolumn{4}{c}{\textbf{RMSE ($\downarrow$)}} & \multicolumn{4}{c}{\textbf{Bias ($\approx 0$)}} \\
  \cmidrule(lr){2-5}\cmidrule(lr){6-9}\cmidrule(lr){10-13}
  & \makecell{DS\\V4F} & \makecell{Kimi \\ K2.5} & \makecell{Gem.\\3P} & \makecell{Q3.5\\27B} & \makecell{DS-\\V4F} & \makecell{Kimi \\ K2.5} & \makecell{Gem.\\3P} & \makecell{Q3.5\\27B} & \makecell{DS-\\V4F} & \makecell{Kimi \\ K2.5} & \makecell{Gem.\\3P} & \makecell{Q3.5\\27B} \\
  \midrule 
  \multicolumn{12}{l}{\textit{Chain-of-Thought (CoT@1)}} \\
  \midrule 
  MathArena~\cite{DBLP:journals/corr/abs-2505-23281}         & 0.52 & 0.35 & - & 0.40 & 1.89 & 2.59 & - & 2.60 & -0.48 & 2.20 & - & 2.10 \\
  \midrule 
  \multicolumn{12}{l}{\textit{Chain-of-Thought with Majority Vote Selection (MajorityVote@16)}} \\
  \midrule 
  MathArena~\cite{DBLP:journals/corr/abs-2505-23281}         & 0.54 & 0.36 & - & 0.41 & 1.83 & 2.56 & - & 2.55 & -0.50 & 2.25 & - & 2.20 \\
  \midrule 
  \multicolumn{12}{l}{\textit{Chain-of-Thought with Best-of-N Selection (BoN@16)}} \\
  \midrule 
  Skywork~\cite{DBLP:journals/corr/abs-2410-18451}           & 0.51 & 0.34 & - & 0.41 & 1.81 & 2.61 & - & 2.67 & -0.52 & 2.35 & - & 2.15 \\
  ArmoRM~\cite{DBLP:conf/emnlp/00030X0024}                   & 0.55 & 0.35 & - & 0.41 & 1.79 & 2.67 & - & 2.64 & -0.43 & 2.32 & - & 2.11 \\
  RLHFlow~\cite{dong2024rlhf}                                & 0.53 & 0.35 & - & 0.39 & 1.85 & 2.62 & - & 2.61 & -0.41 & 2.28 & - & 2.18 \\
  Shepherd~\cite{DBLP:conf/acl/WangLSXDLCWS24}               & 0.52 & 0.36 & - & 0.37 & 1.89 & 2.66 & - & 2.70 & -0.40 & 2.30 & - & 2.20 \\
  \midrule
  \multicolumn{12}{l}{\textit{Proposed Method (\nospacemethod @16)}} \\
  \midrule
  \method & 0.58 & 0.49 & - & 0.53 & 1.67 & 2.24 & - & 1.90 & -0.50 & 0.71 & - & -0.96 \\
  \bottomrule
\end{tabular}
\end{footnotesize}
\end{table*}


\clearpage
\section{Extended Bar plots}
\label{app:barplots}

\begin{figure}[htbp]
    \centering
    \includegraphics[width=\textwidth]{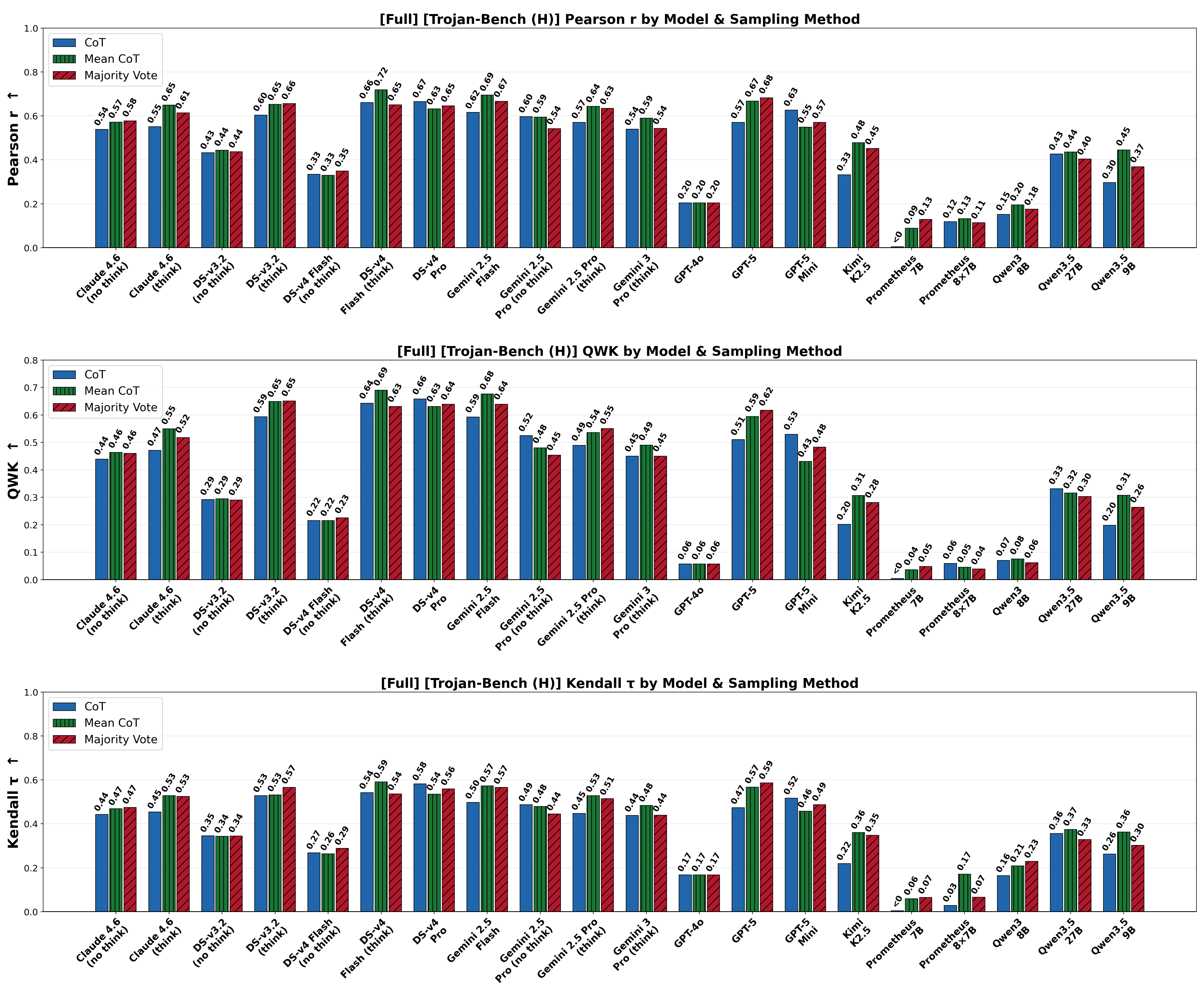}
    \caption{\textbf\humands (Full): Correlation metrics (Pearson $r$, QWK, Kendall $\tau$) by model and sampling method.}
    \label{fig:bars_usamo_2025_full_correlation}
\end{figure}

\begin{figure}[htbp]
    \centering
    \includegraphics[width=\textwidth]{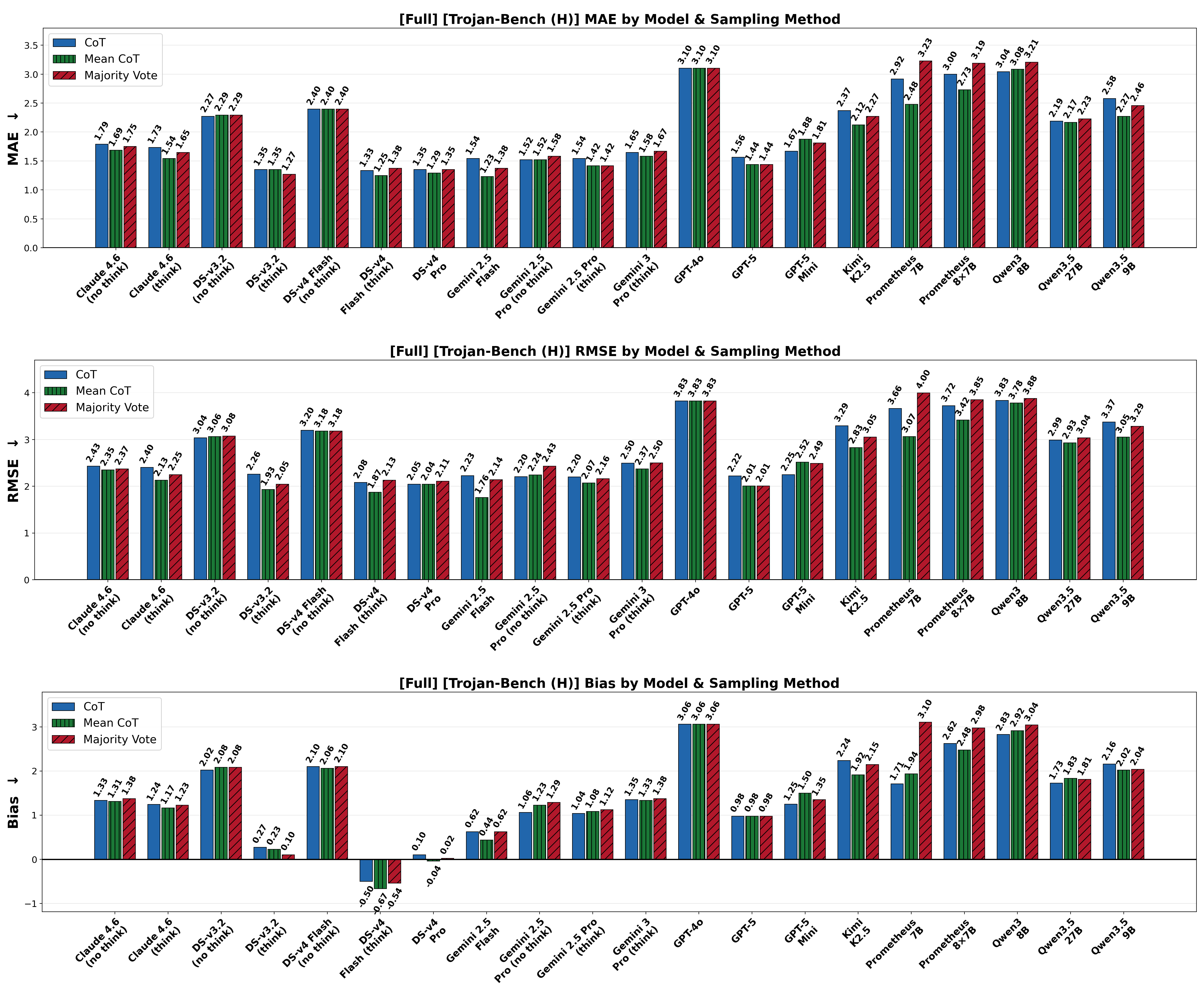}
    \caption{\textbf\humands (Full): Error metrics (MAE, RMSE, Bias) by model and sampling method.}
    \label{fig:bars_usamo_2025_full_error}
\end{figure}

\begin{figure}[htbp]
    \centering
    \includegraphics[width=\textwidth]{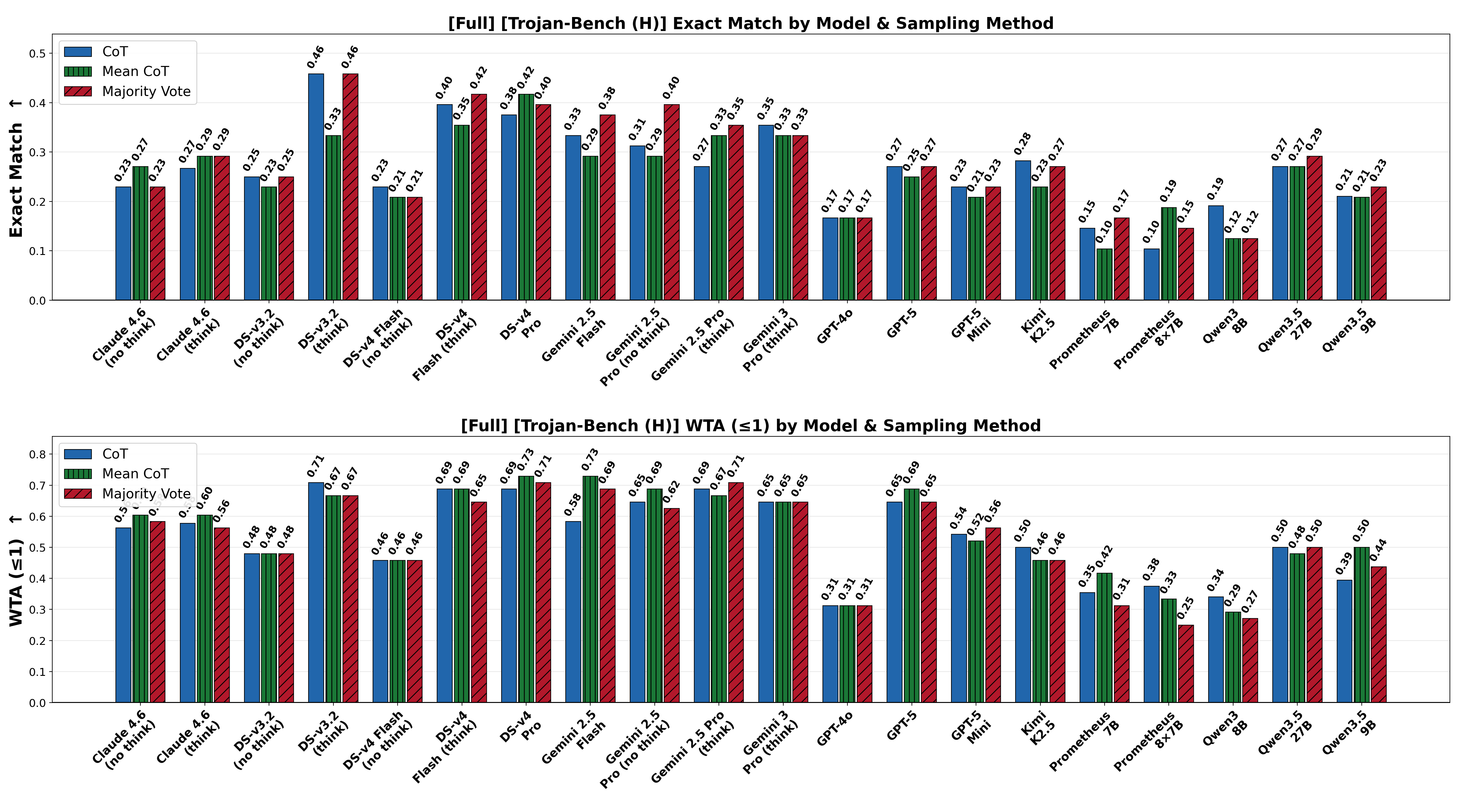}
    \caption{\textbf\humands (Full): Agreement metrics (Exact Match, WTA $\leq 1$) by model and sampling method.}
    \label{fig:bars_usamo_2025_full_agreement}
\end{figure}

\begin{figure}[htbp]
    \centering
    \includegraphics[width=\textwidth]{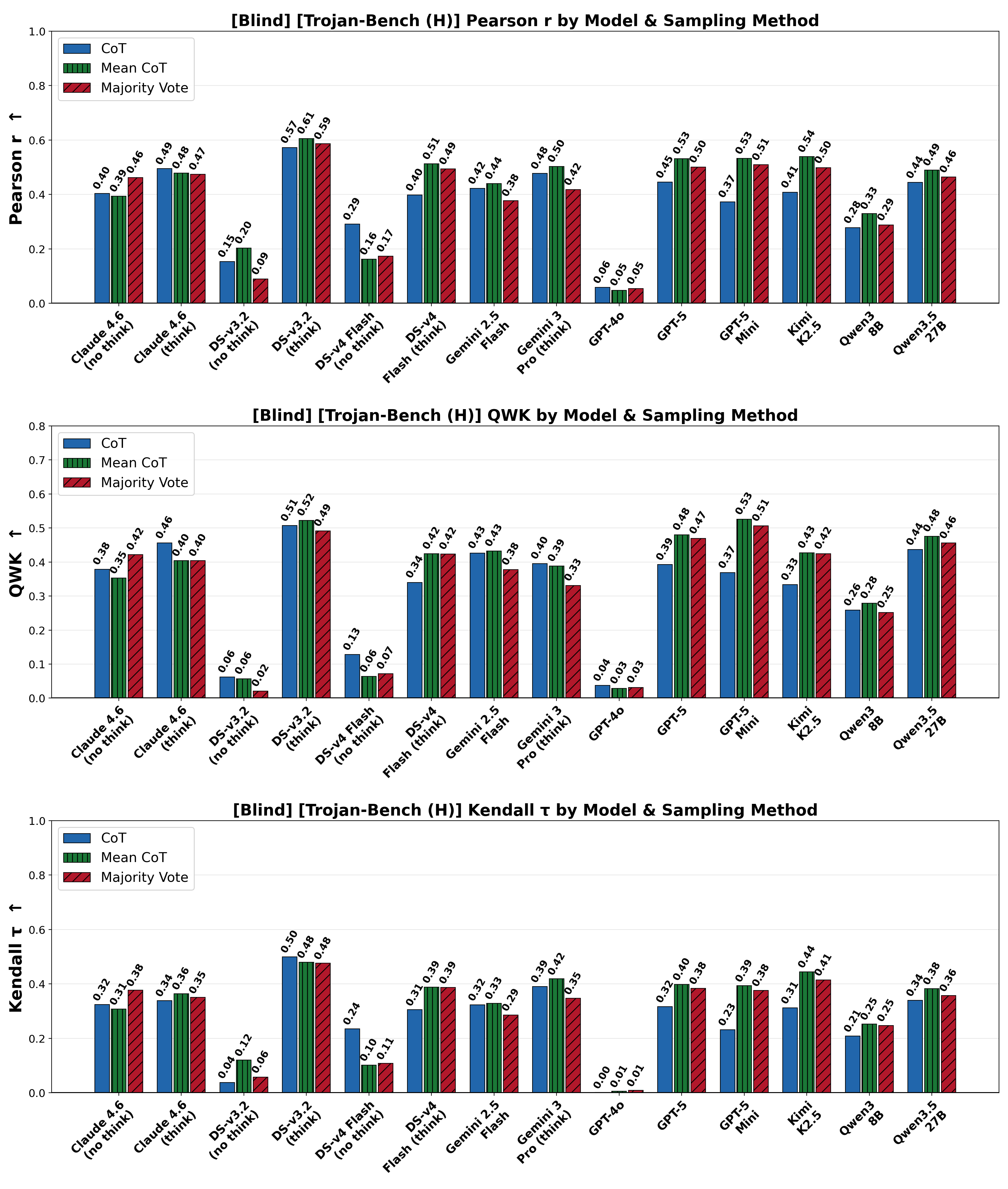}
    \caption{\textbf\humands (Blind): Correlation metrics (Pearson $r$, QWK, Kendall $\tau$) by model and sampling method.}
    \label{fig:bars_usamo_2025_blind_correlation}
\end{figure}

\begin{figure}[htbp]
    \centering
    \includegraphics[width=\textwidth]{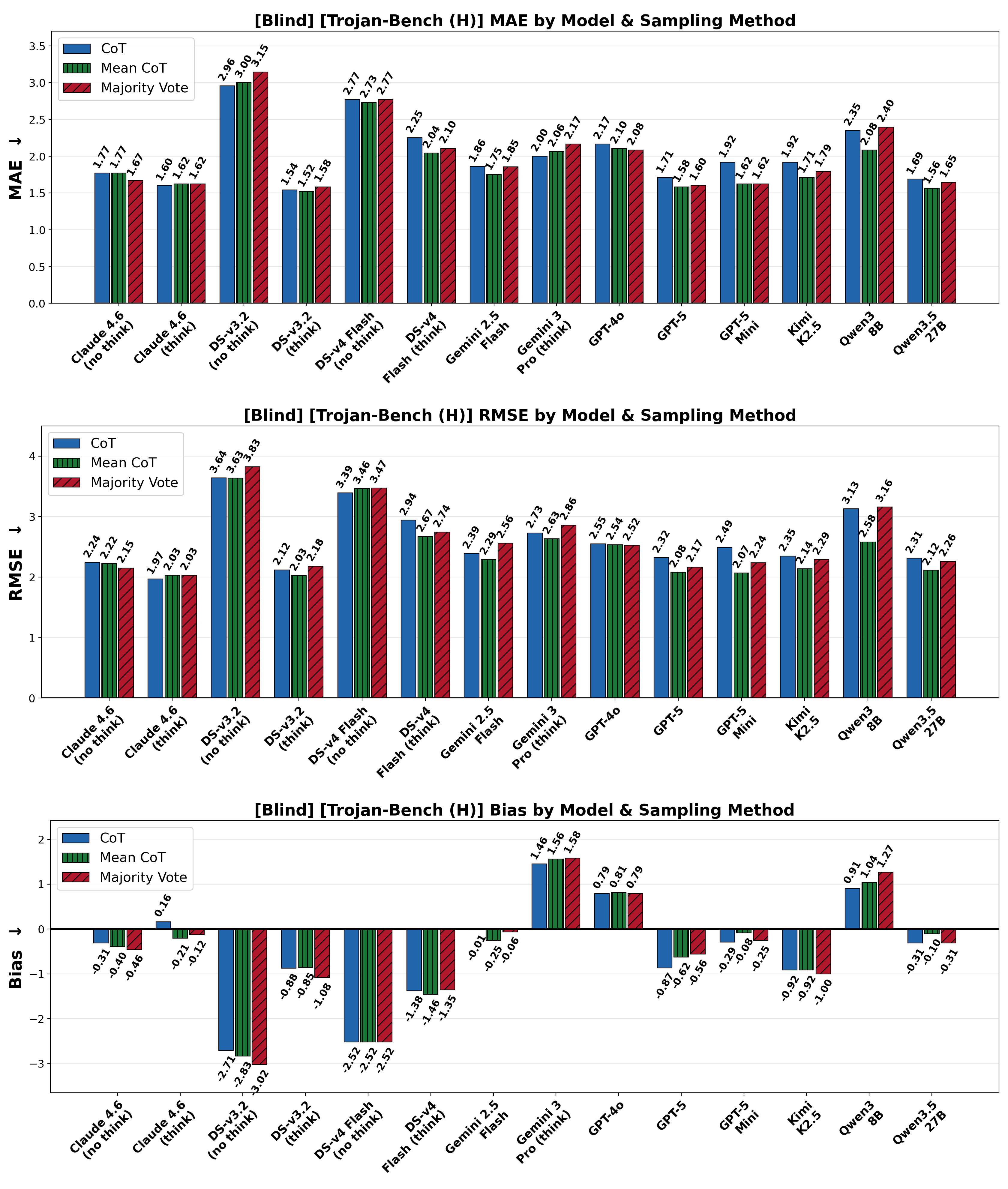}
    \caption{\textbf\humands (Blind): Error metrics (MAE, RMSE, Bias) by model and sampling method.}
    \label{fig:bars_usamo_2025_blind_error}
\end{figure}

\begin{figure}[htbp]
    \centering
    \includegraphics[width=\textwidth]{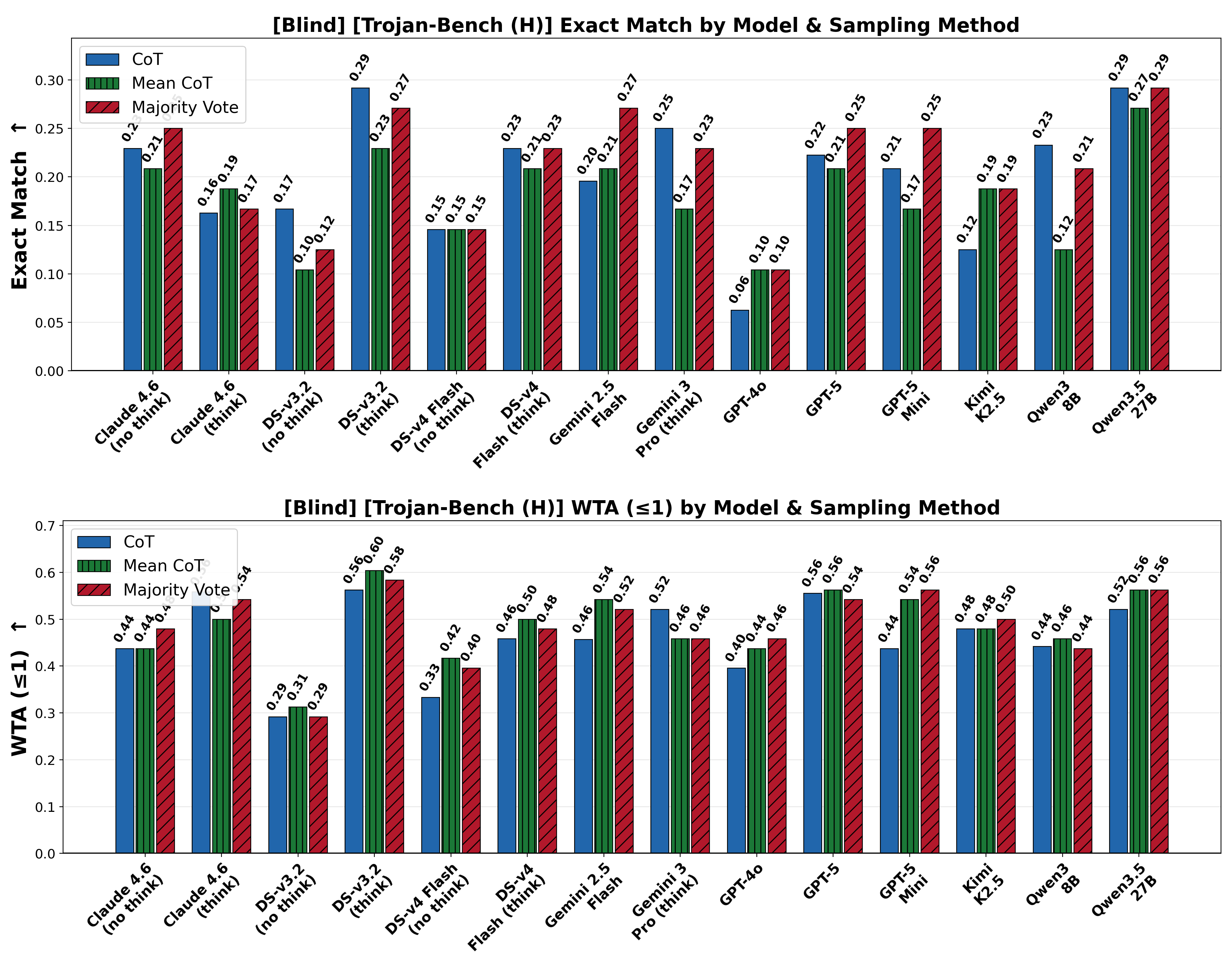}
    \caption{\textbf\humands (Blind): Agreement metrics (Exact Match, WTA $\leq 1$) by model and sampling method.}
    \label{fig:bars_usamo_2025_blind_agreement}
\end{figure}

\begin{figure}[htbp]
    \centering
    \includegraphics[width=\textwidth]{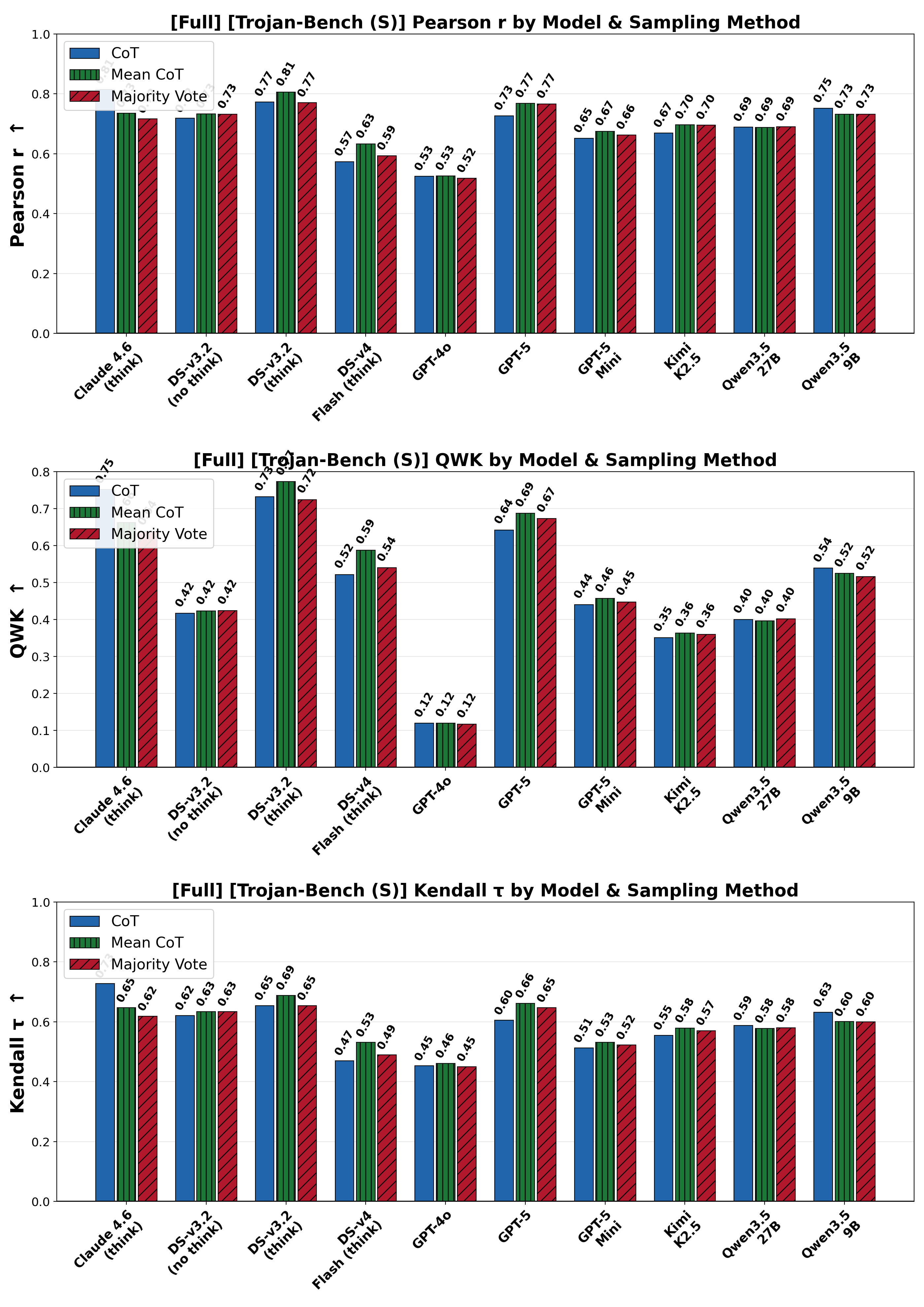}
    \caption{\textbf{\synthds} (Full): Correlation metrics (Pearson $r$, QWK, Kendall $\tau$) by model and sampling method.}
    \label{fig:bars_synthetic_200_full_correlation}
\end{figure}

\begin{figure}[htbp]
    \centering
    \includegraphics[width=\textwidth]{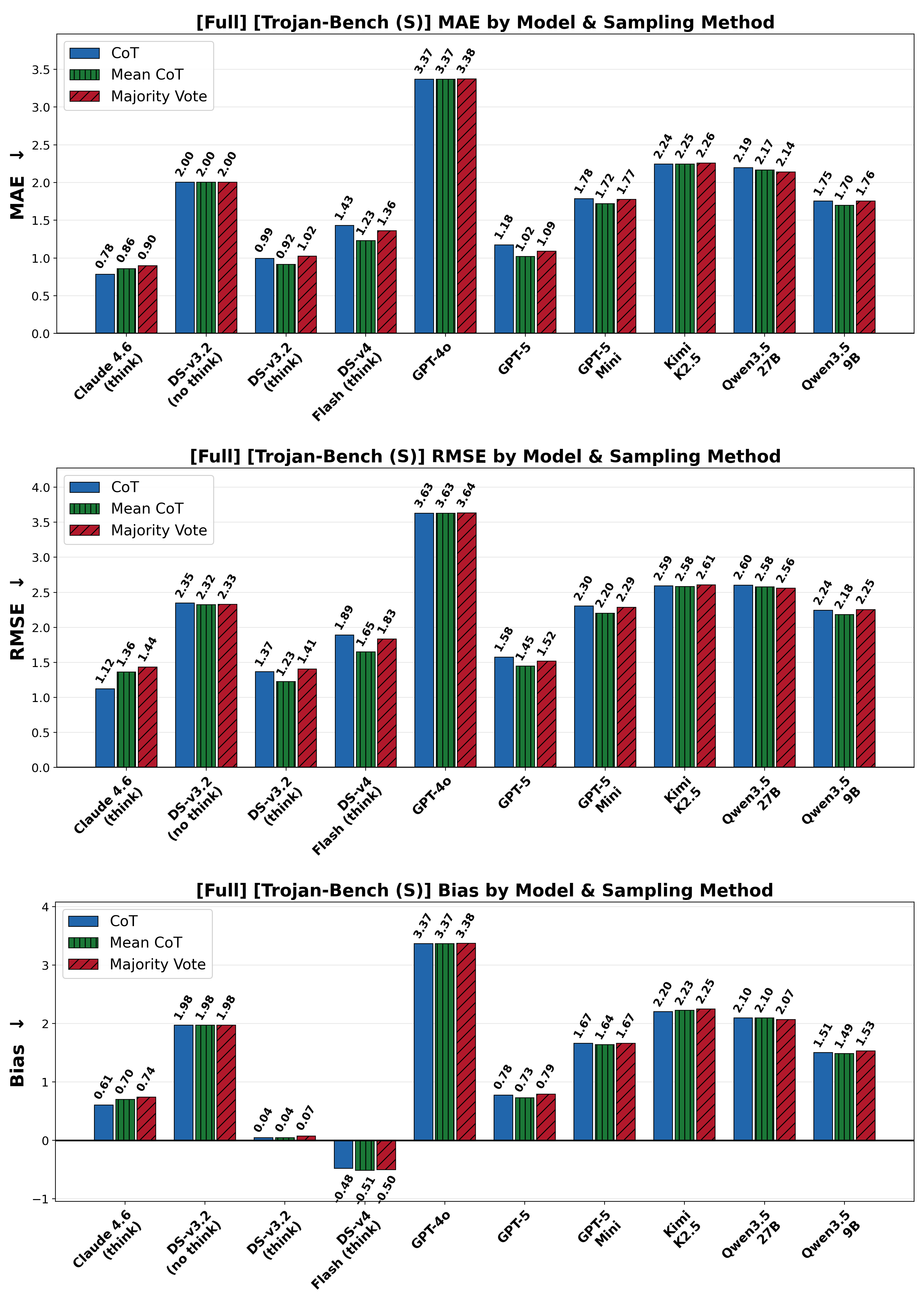}
    \caption{\textbf{\synthds} (Full): Error metrics (MAE, RMSE, Bias) by model and sampling method.}
    \label{fig:bars_synthetic_200_full_error}
\end{figure}

\begin{figure}[htbp]
    \centering
    \includegraphics[width=\textwidth]{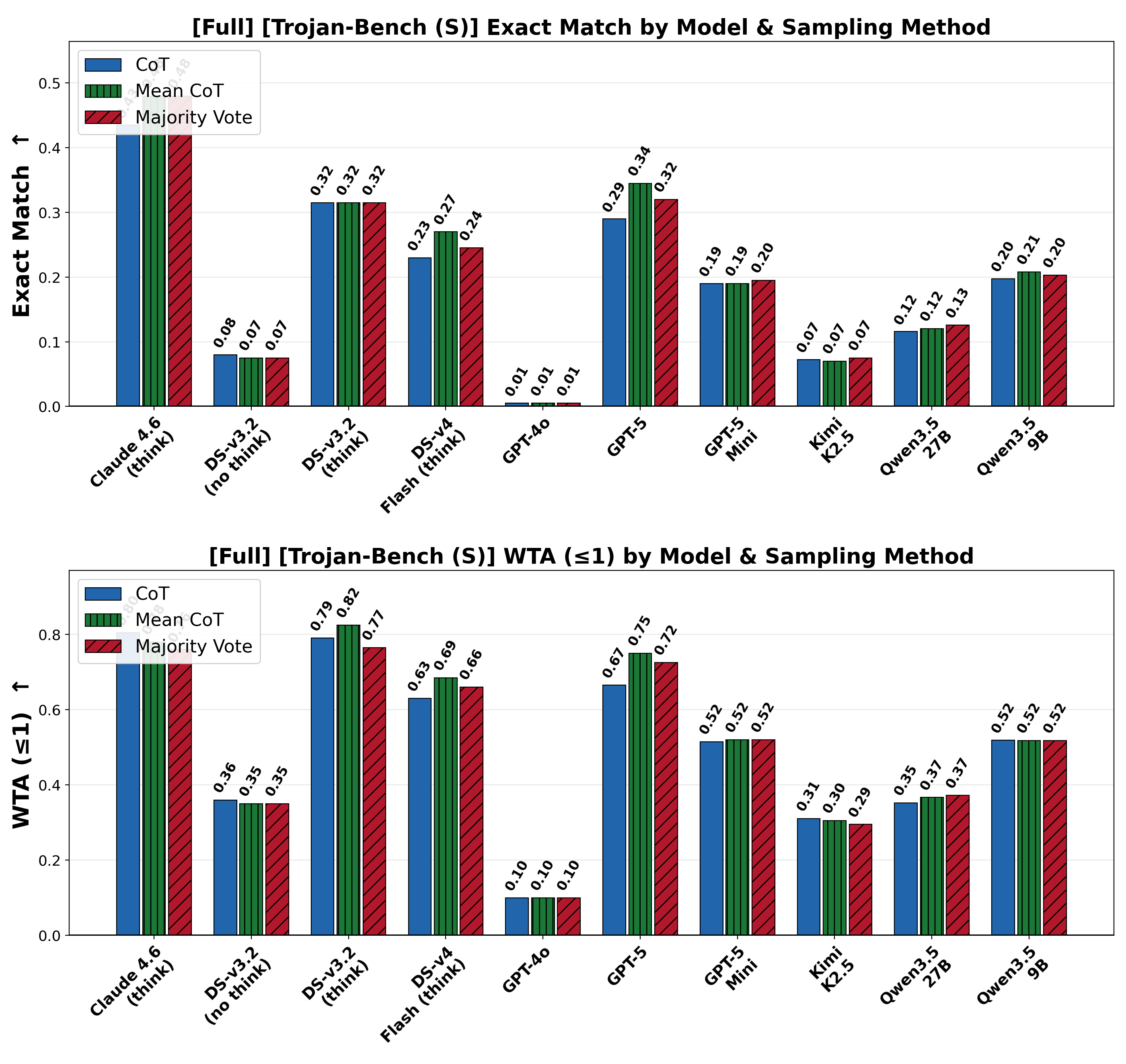}
    \caption{\textbf{\synthds} (Full): Agreement metrics (Exact Match, WTA $\leq 1$) by model and sampling method.}
    \label{fig:bars_synthetic_200_full_agreement}
\end{figure}

\begin{figure}[htbp]
    \centering
    \includegraphics[width=\textwidth]{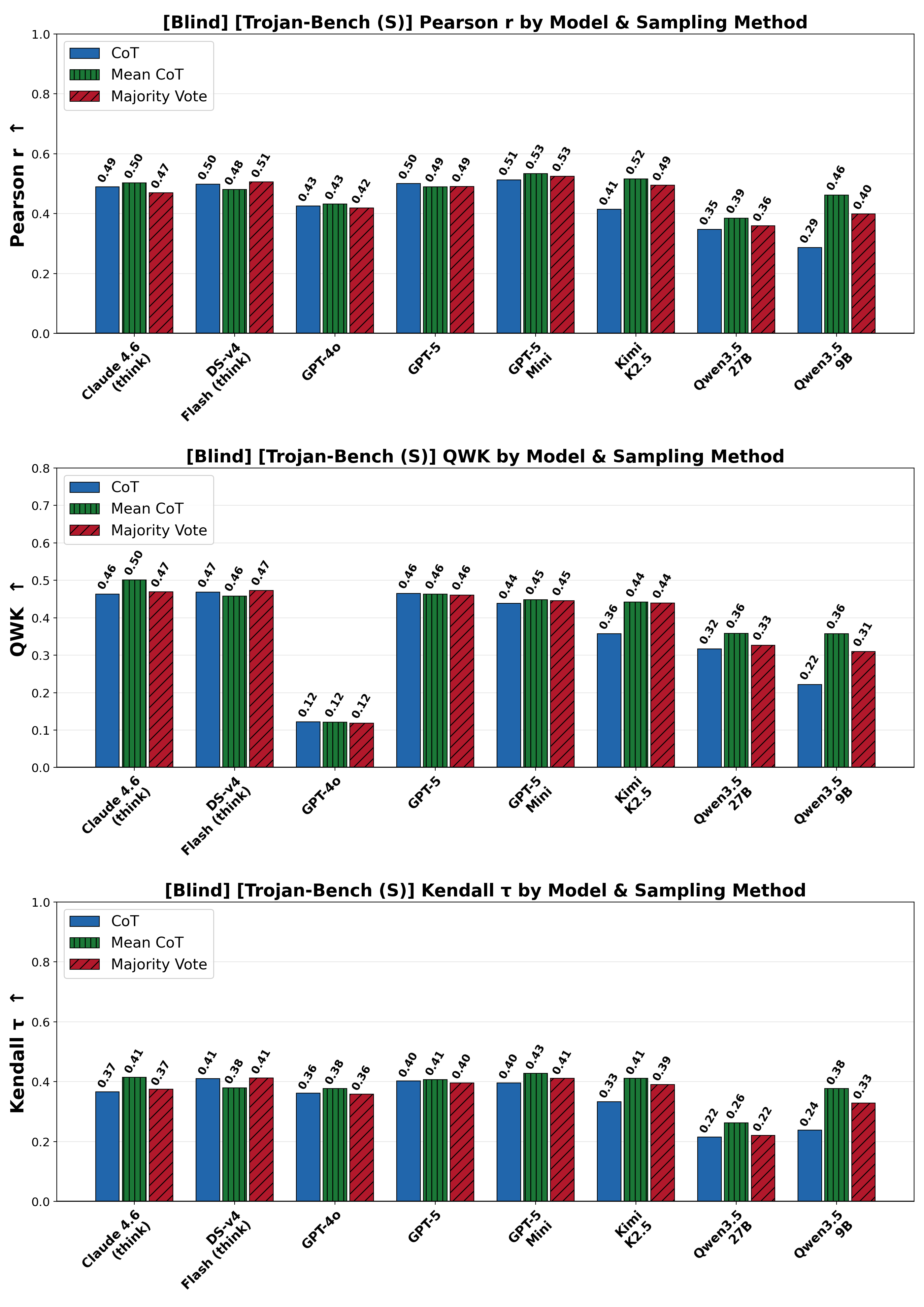}
    \caption{\textbf{\synthds} (Blind): Correlation metrics (Pearson $r$, QWK, Kendall $\tau$) by model and sampling method.}
    \label{fig:bars_synthetic_200_blind_correlation}
\end{figure}

\begin{figure}[htbp]
    \centering
    \includegraphics[width=\textwidth]{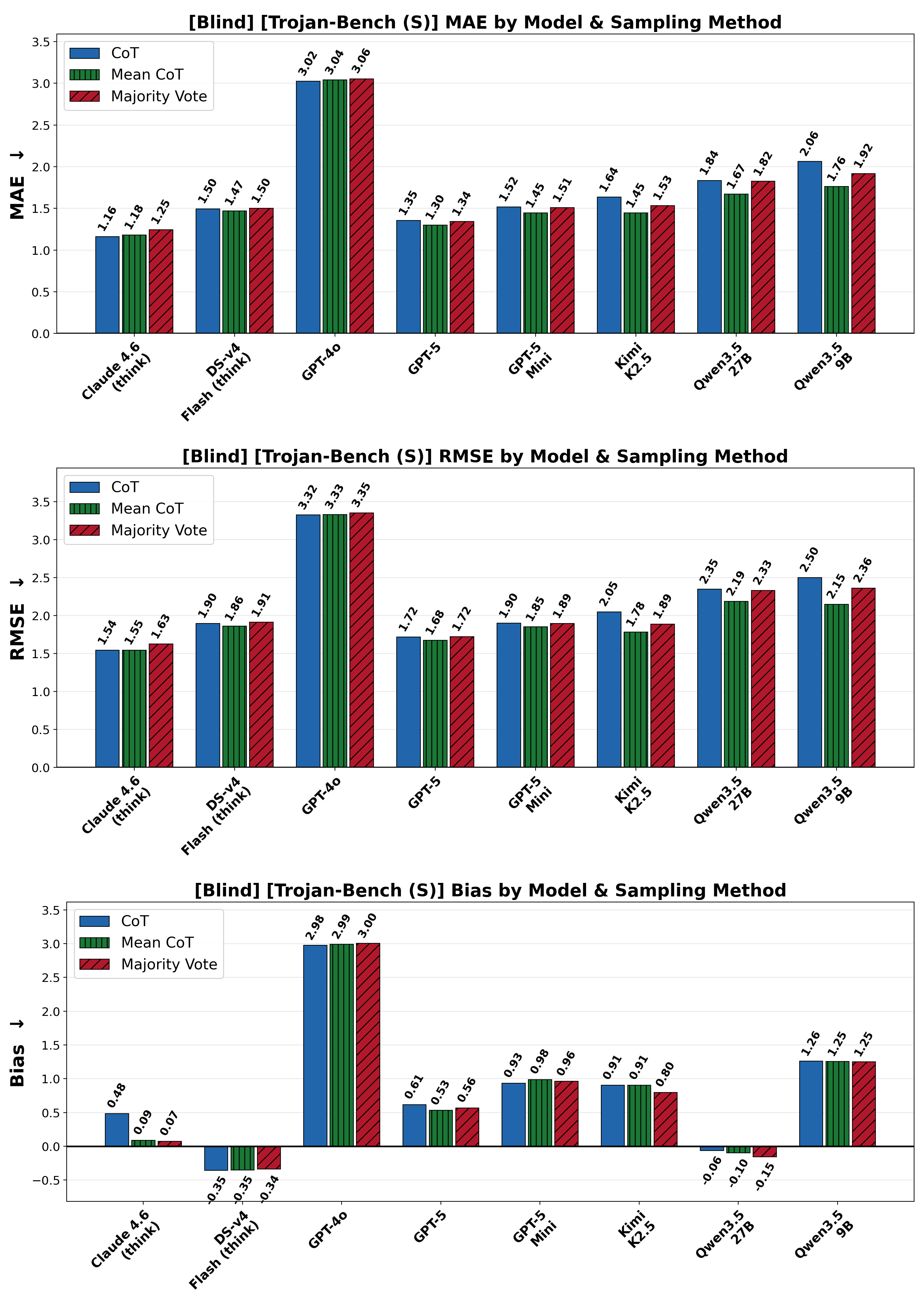}
    \caption{\textbf{\synthds} (Blind): Error metrics (MAE, RMSE, Bias) by model and sampling method.}
    \label{fig:bars_synthetic_200_blind_error}
\end{figure}

\begin{figure}[htbp]
    \centering
    \includegraphics[width=\textwidth]{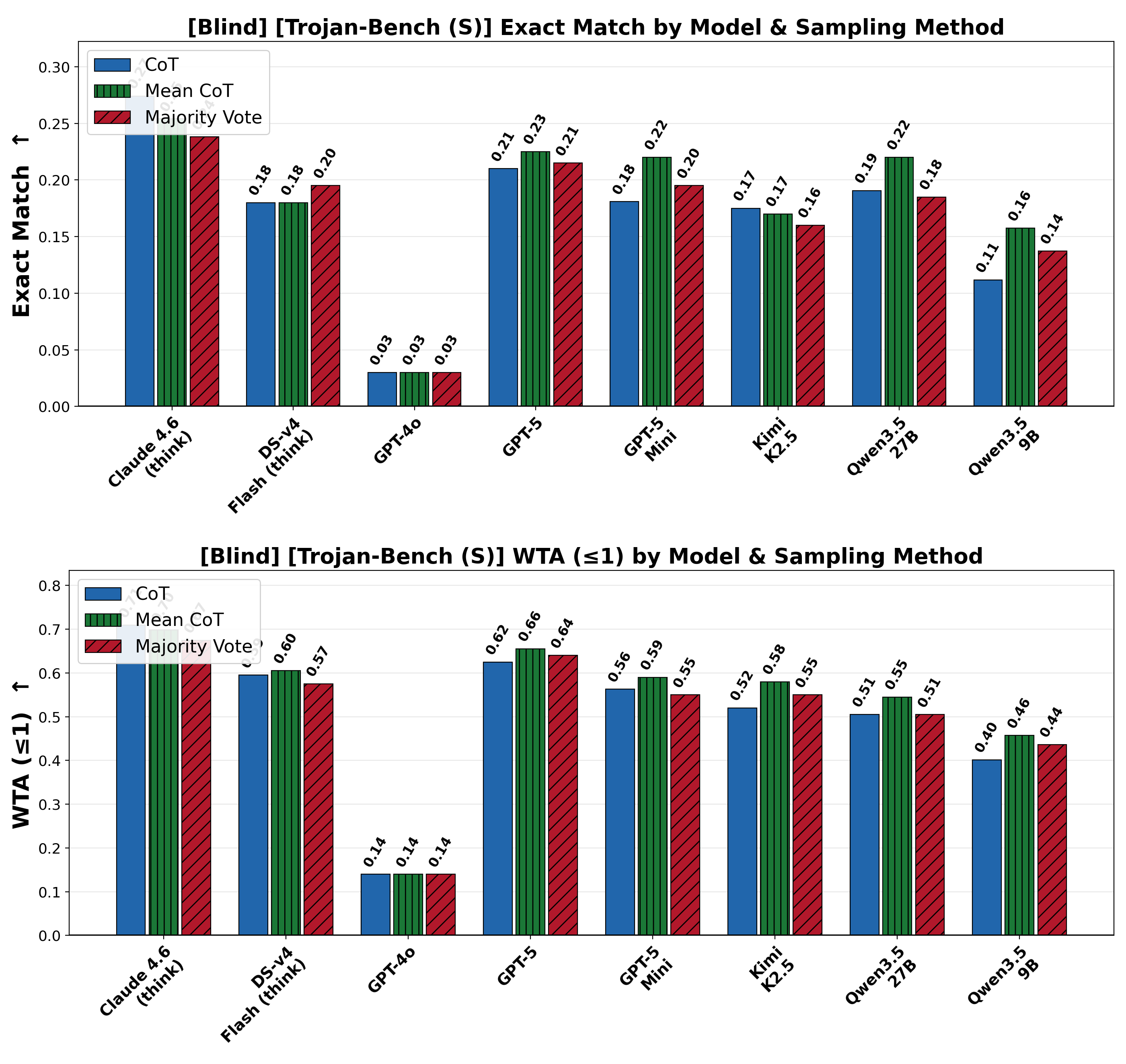}
    \caption{\textbf{\synthds} (Blind): Agreement metrics (Exact Match, WTA $\leq 1$) by model and sampling method.}
    \label{fig:bars_synthetic_200_blind_agreement}
\end{figure}

The bar plots in Figures~\ref{fig:bars_usamo_2025_full_correlation} - \ref{fig:bars_synthetic_200_blind_agreement} are a visual companion to the numerical tables of the previous section. Each plot reports a single metric family (correlation, error, or agreement) for one (dataset, style) pair, with three bars per model corresponding to CoT@1, Mean-CoT@16, and MajorityVote@16. We include them so that a reader can compare every grader and every aggregation method at a glance for any single metric, without having to cross-reference the numerical tables.

The plots make the patterns from the previous section visually obvious. Frontier thinking-enabled models cluster in the upper band of the correlation plots and in the lower band of the error plots, whereas smaller graders separate cleanly into a lower-correlation, higher-MAE cluster. The agreement plots (Exact Match and WTA $\leq 1$) show that close-but-not-exact agreement with the human reference is achievable for most frontier models, while exact agreement remains low even for the strongest graders. The synthetic-set plots show the same ordering as the human-set plots but with a uniform upward shift in correlation and a downward shift in error, consistent with the cross-set discussion above.

\clearpage
\section{Evaluation of LLM correlation with human and other LLM annotators}
\label{app:heatmaps}

\begin{figure}[htbp]
    \centering
    \includegraphics[width=0.5\textwidth]{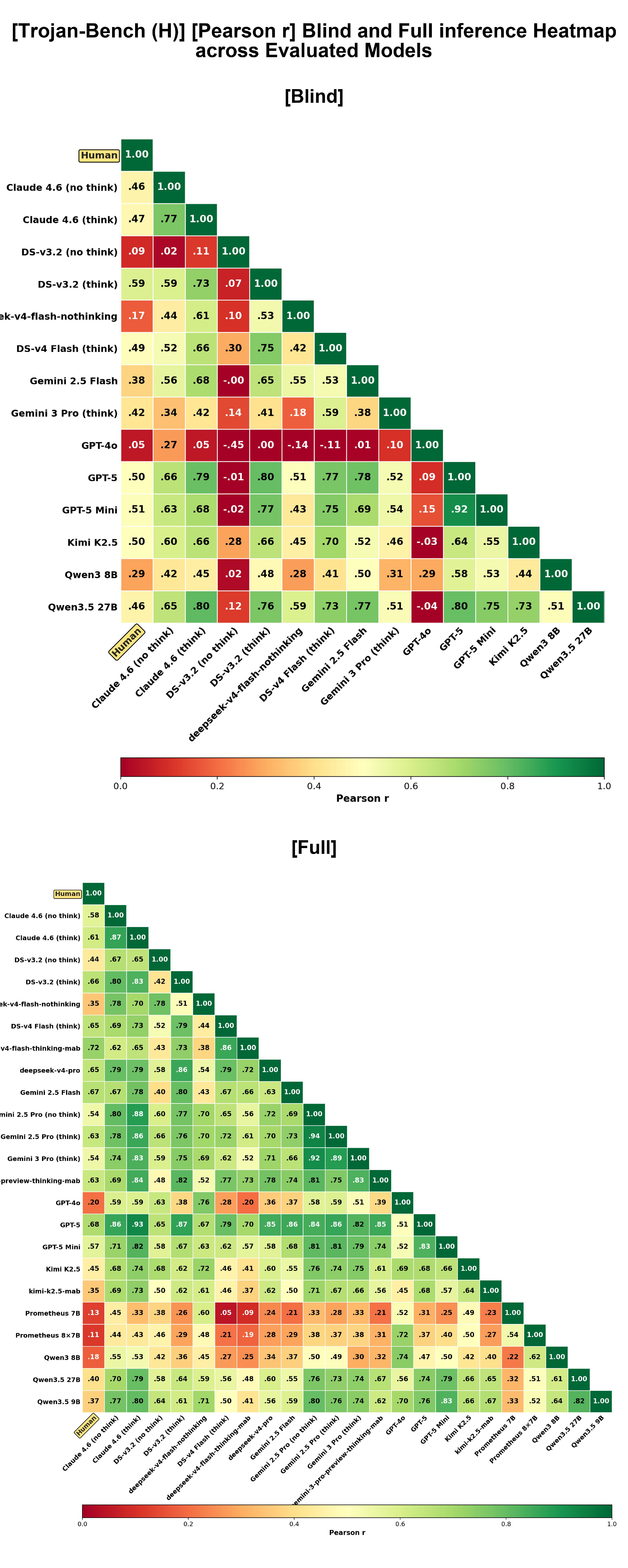}
    \caption{\textbf{\humands}: Pearson $r$ blind and full inference heatmap across evaluated models (Majority Vote; top: Blind, bottom: Full).}
    \label{fig:heatmap_usamo_2025_pearson_r}
\end{figure}

\begin{figure}[htbp]
    \centering
    \includegraphics[width=0.5\textwidth]{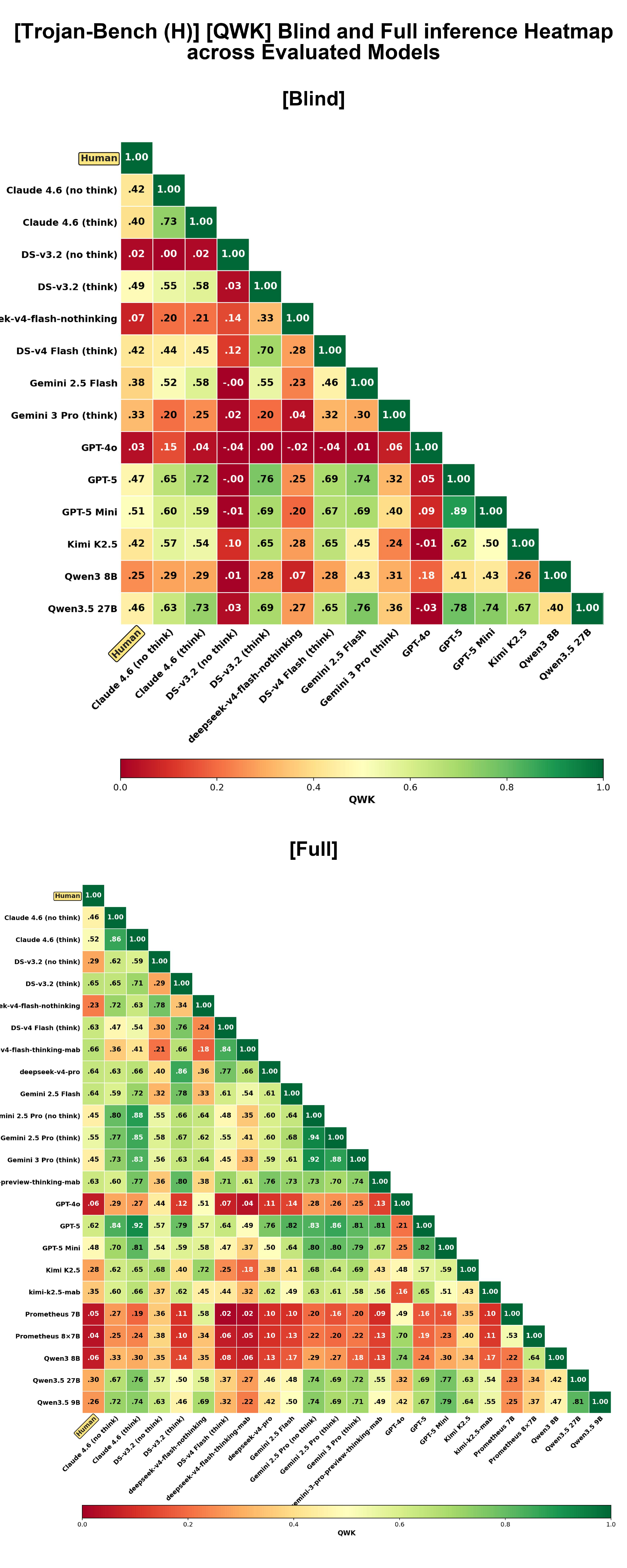}
    \caption{\textbf{\humands}: QWK blind and full inference heatmap across evaluated models (Majority Vote; top: Blind, bottom: Full).}
    \label{fig:heatmap_usamo_2025_qwk}
\end{figure}

\begin{figure}[htbp]
    \centering
    \includegraphics[width=0.5\textwidth]{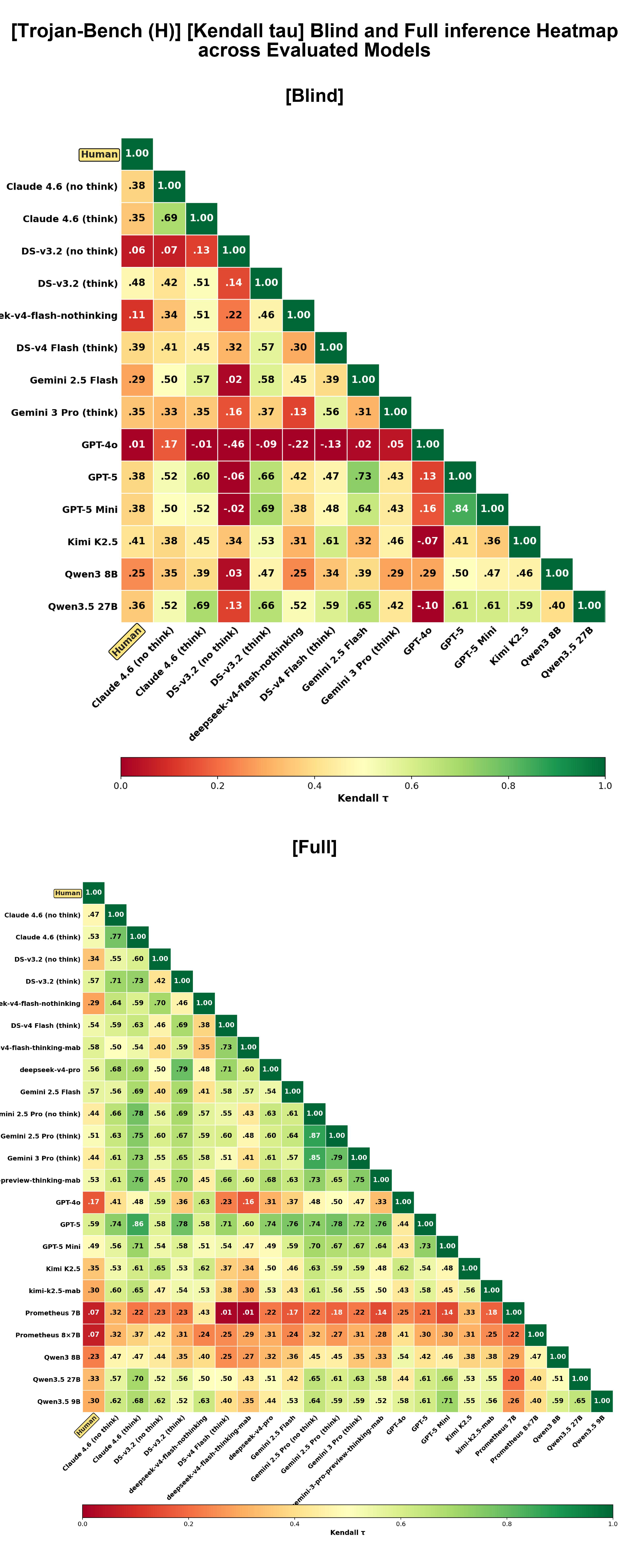}
    \caption{\textbf{\humands}: Kendall $\tau$ blind and full inference heatmap across evaluated models (Majority Vote; top: Blind, bottom: Full).}
    \label{fig:heatmap_usamo_2025_kendall_tau}
\end{figure}

\begin{figure}[htbp]
    \centering
    \includegraphics[width=0.5\textwidth]{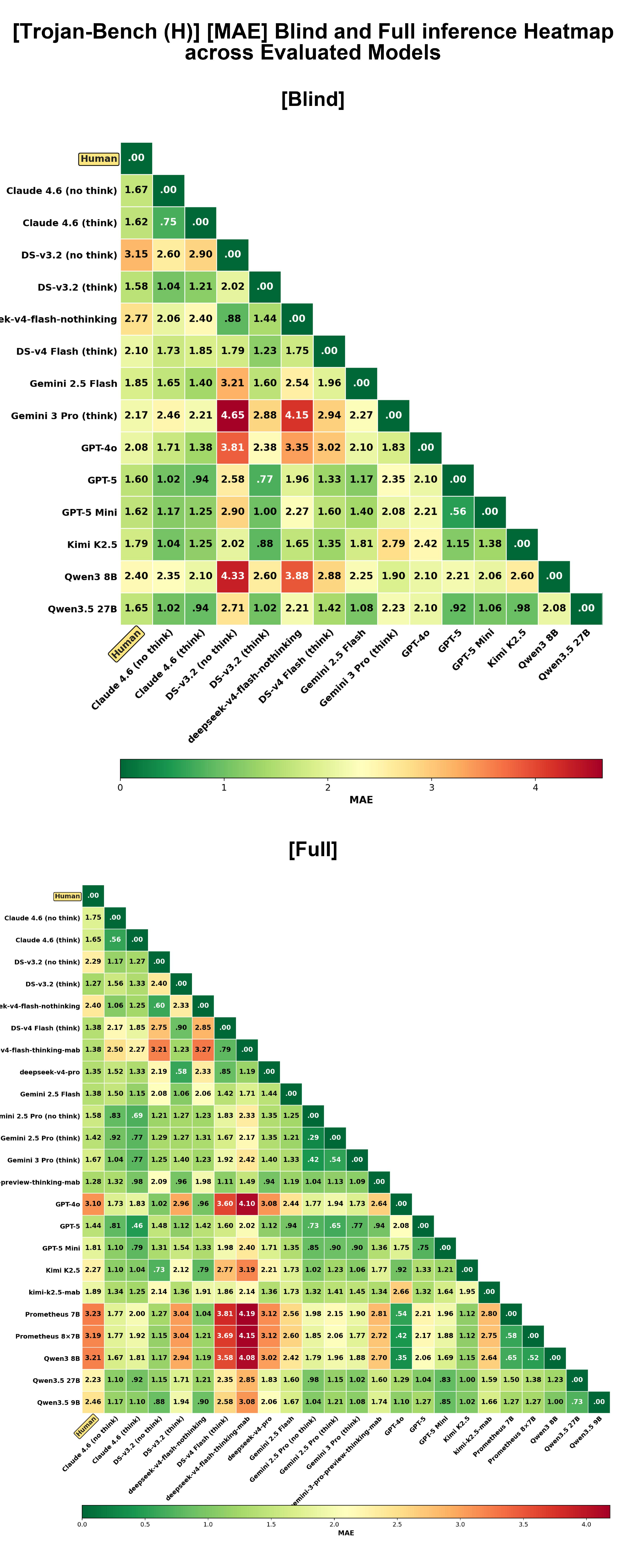}
    \caption{\textbf{\humands}: MAE blind and full inference heatmap across evaluated models (Majority Vote; top: Blind, bottom: Full).}
    \label{fig:heatmap_usamo_2025_mae}
\end{figure}

\begin{figure}[htbp]
    \centering
    \includegraphics[width=0.5\textwidth]{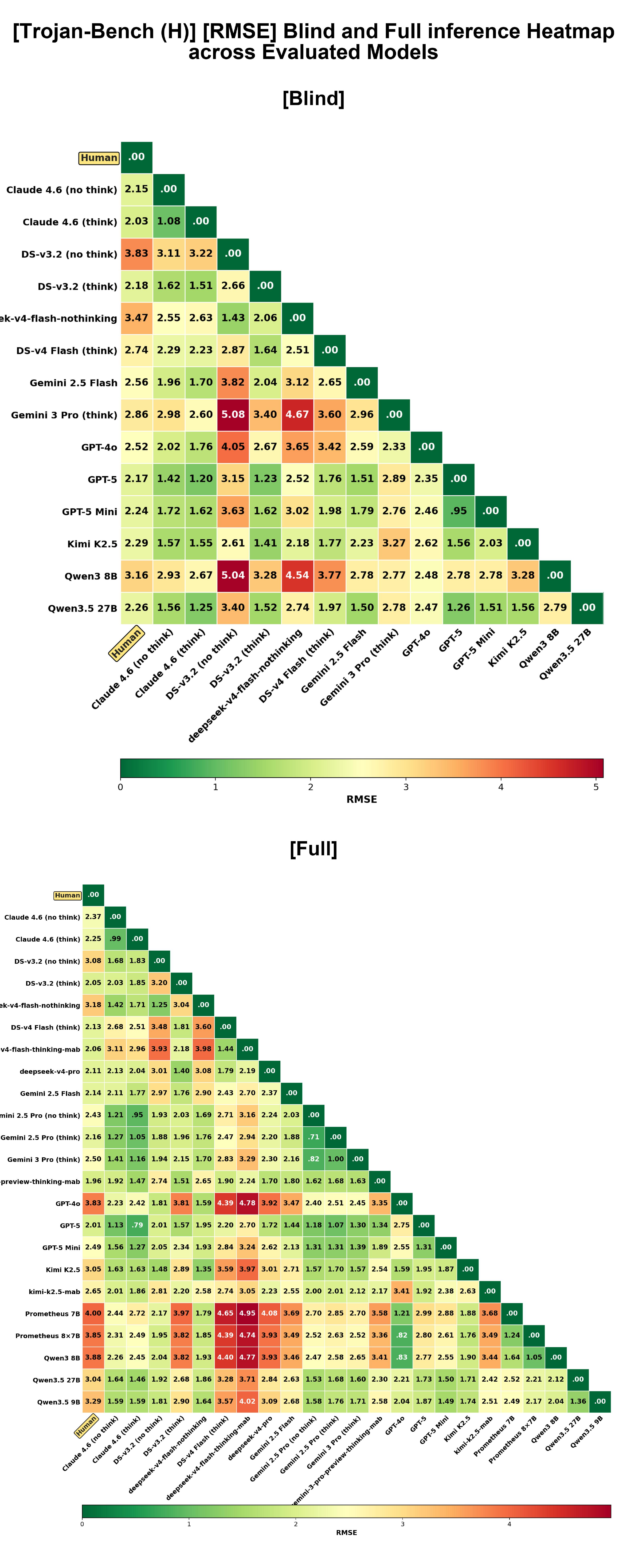}
    \caption{\textbf{\humands}: RMSE blind and full inference heatmap across evaluated models (Majority Vote; top: Blind, bottom: Full).}
    \label{fig:heatmap_usamo_2025_rmse}
\end{figure}

\begin{figure}[htbp]
    \centering
    \includegraphics[width=0.5\textwidth]{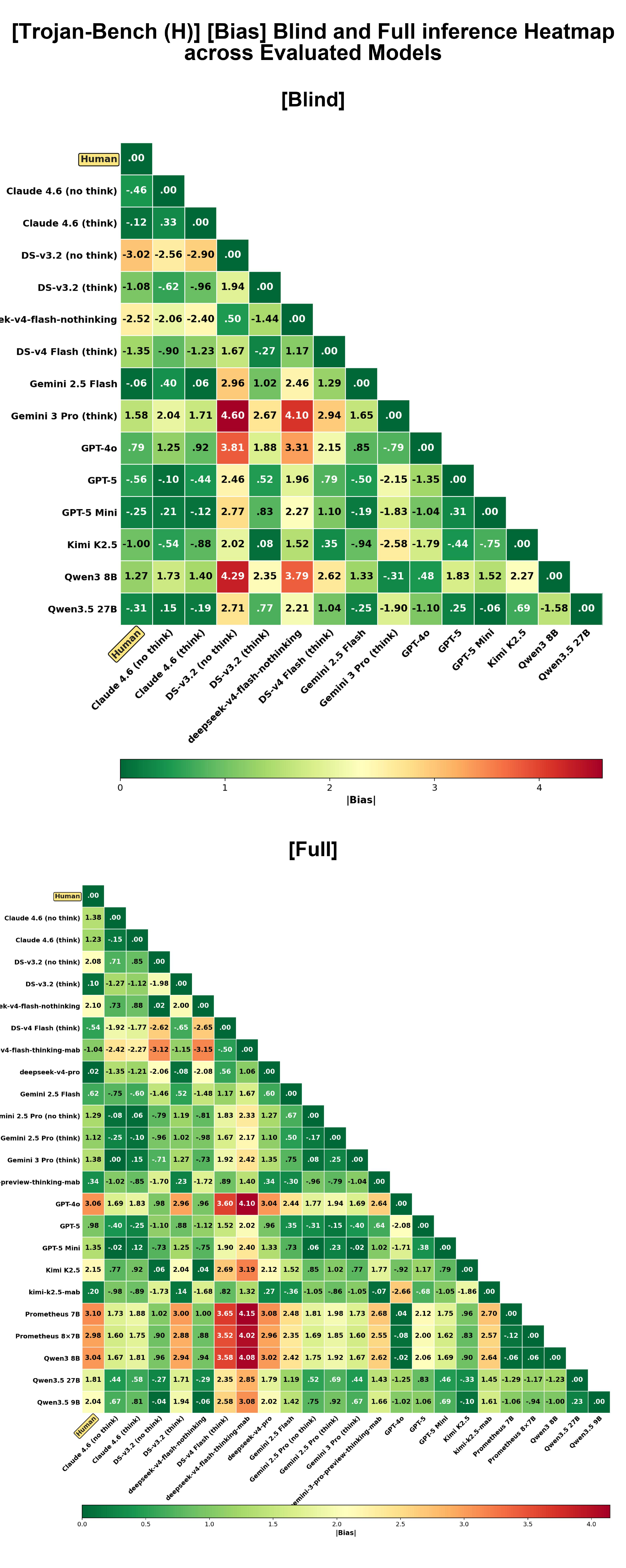}
    \caption{\textbf{\humands}: Bias blind and full inference heatmap across evaluated models (Majority Vote; top: Blind, bottom: Full).}
    \label{fig:heatmap_usamo_2025_bias}
\end{figure}

\begin{figure}[htbp]
    \centering
    \includegraphics[width=0.5\textwidth]{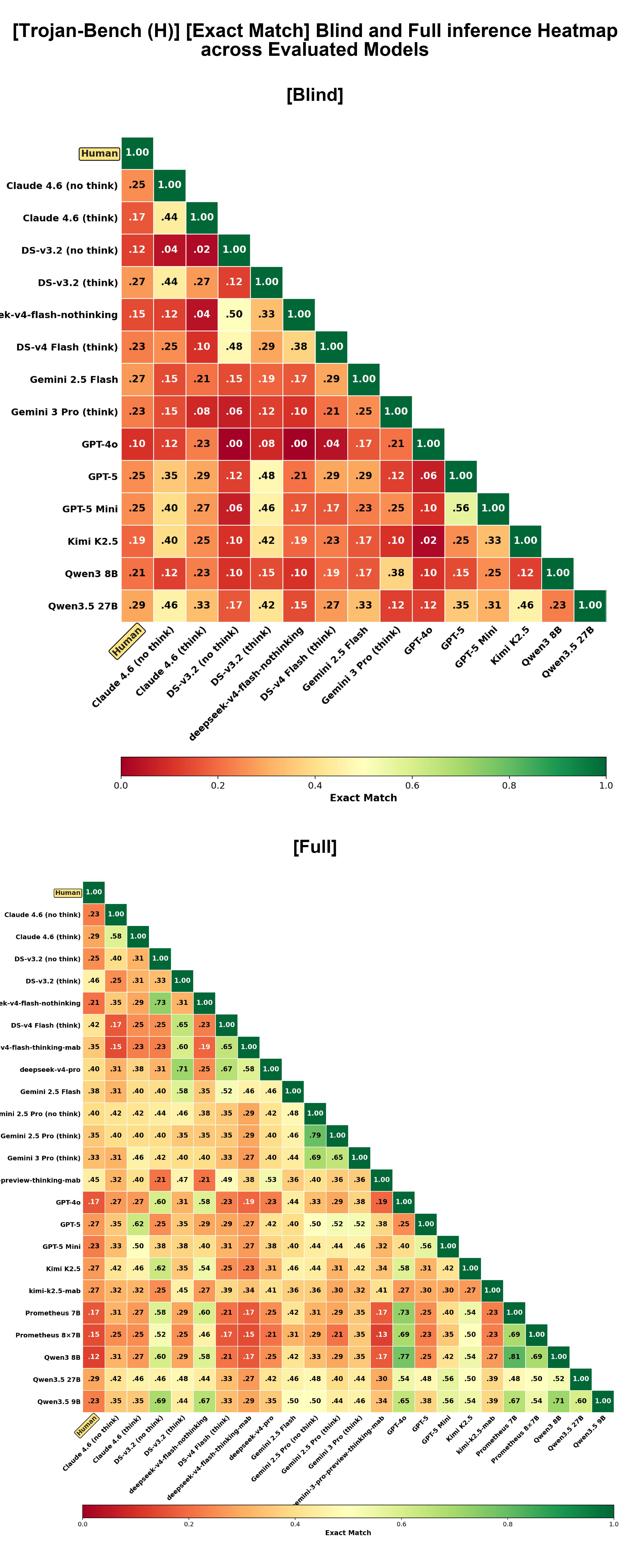}
    \caption{\textbf{\humands}: Exact Match blind and full inference heatmap across evaluated models (Majority Vote; top: Blind, bottom: Full).}
    \label{fig:heatmap_usamo_2025_exact_match}
\end{figure}

\begin{figure}[htbp]
    \centering
    \includegraphics[width=0.5\textwidth]{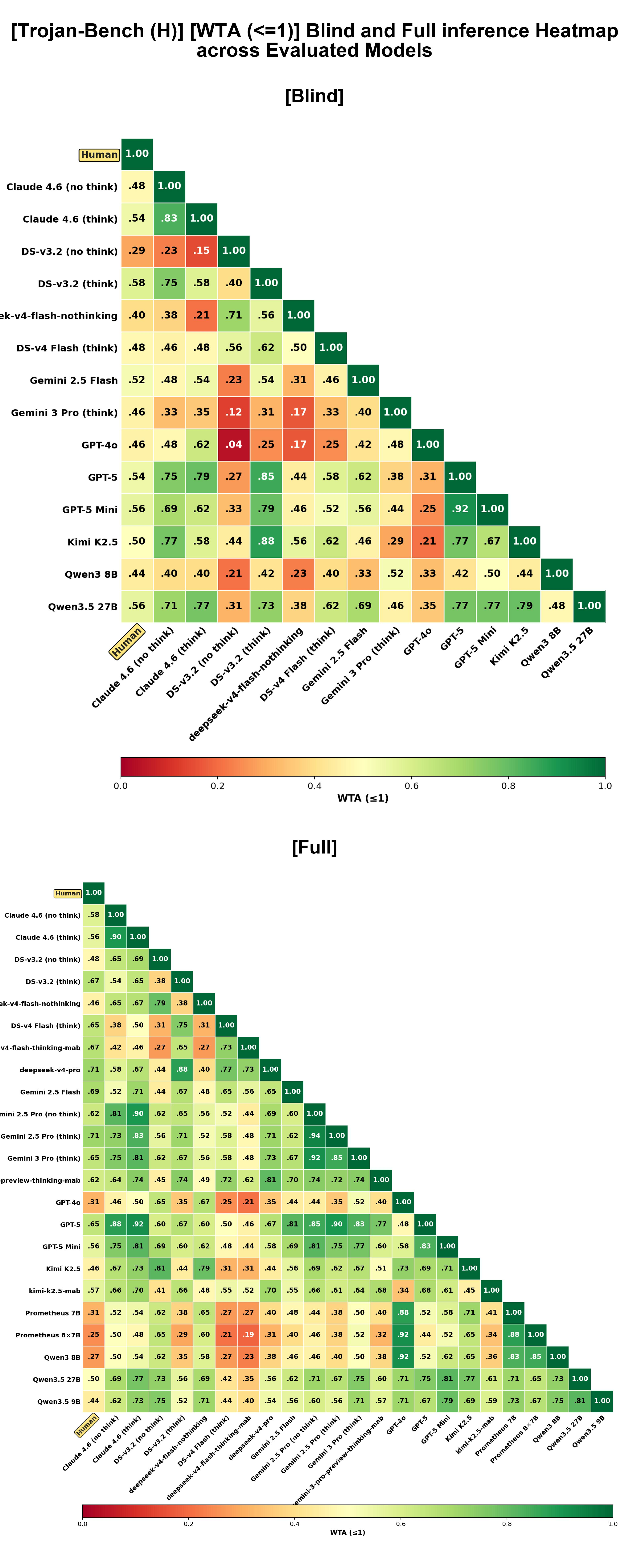}
    \caption{\textbf{\humands}: WTA ($\leq 1$) blind and full inference heatmap across evaluated models (Majority Vote; top: Blind, bottom: Full).}
    \label{fig:heatmap_usamo_2025_wta_leq1}
\end{figure}

\begin{figure}[htbp]
    \centering
    \includegraphics[width=0.5\textwidth]{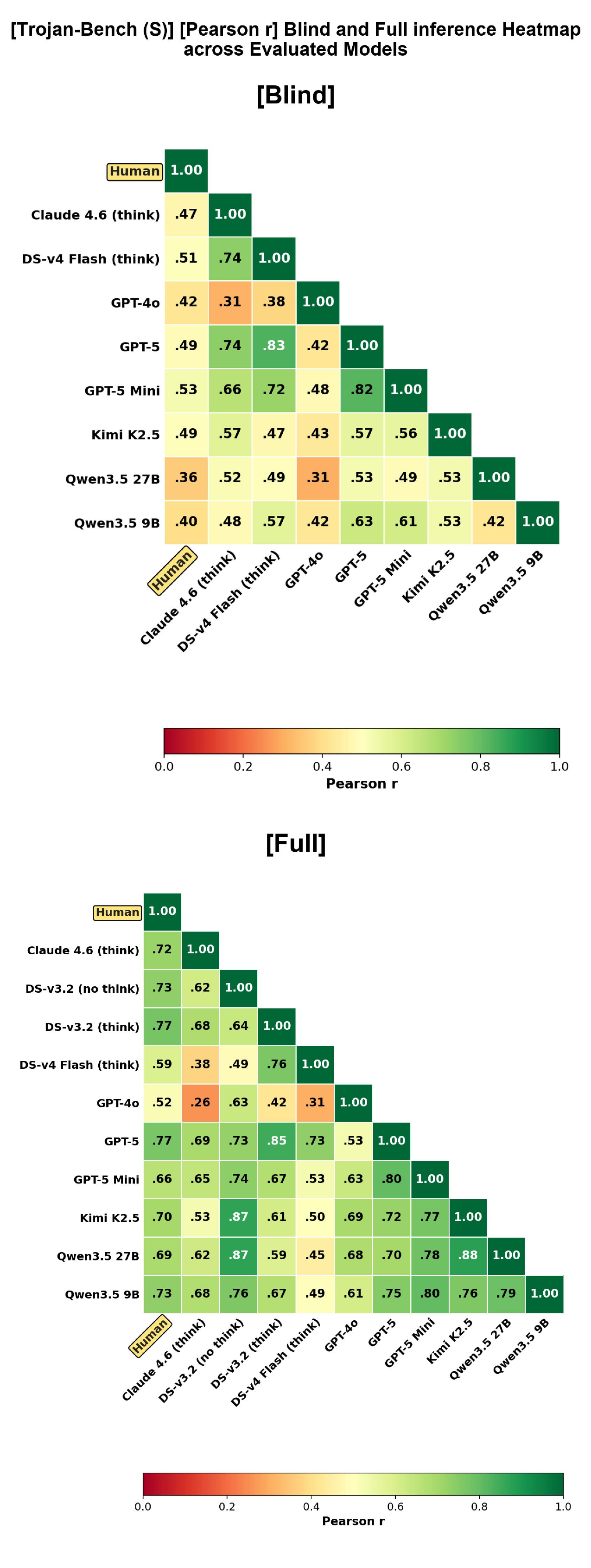}
    \caption{\textbf{\synthds}: Pearson $r$ blind and full inference heatmap across evaluated models (Majority Vote; top: Blind, bottom: Full). Note that "human" in this heatmap indicates an LLM assigned grade during data generation with \synthmethod and serves as a reference point when evaluating the LLMs.}
    \label{fig:heatmap_synthetic_200_pearson_r}
\end{figure}

\begin{figure}[htbp]
    \centering
    \includegraphics[width=0.5\textwidth]{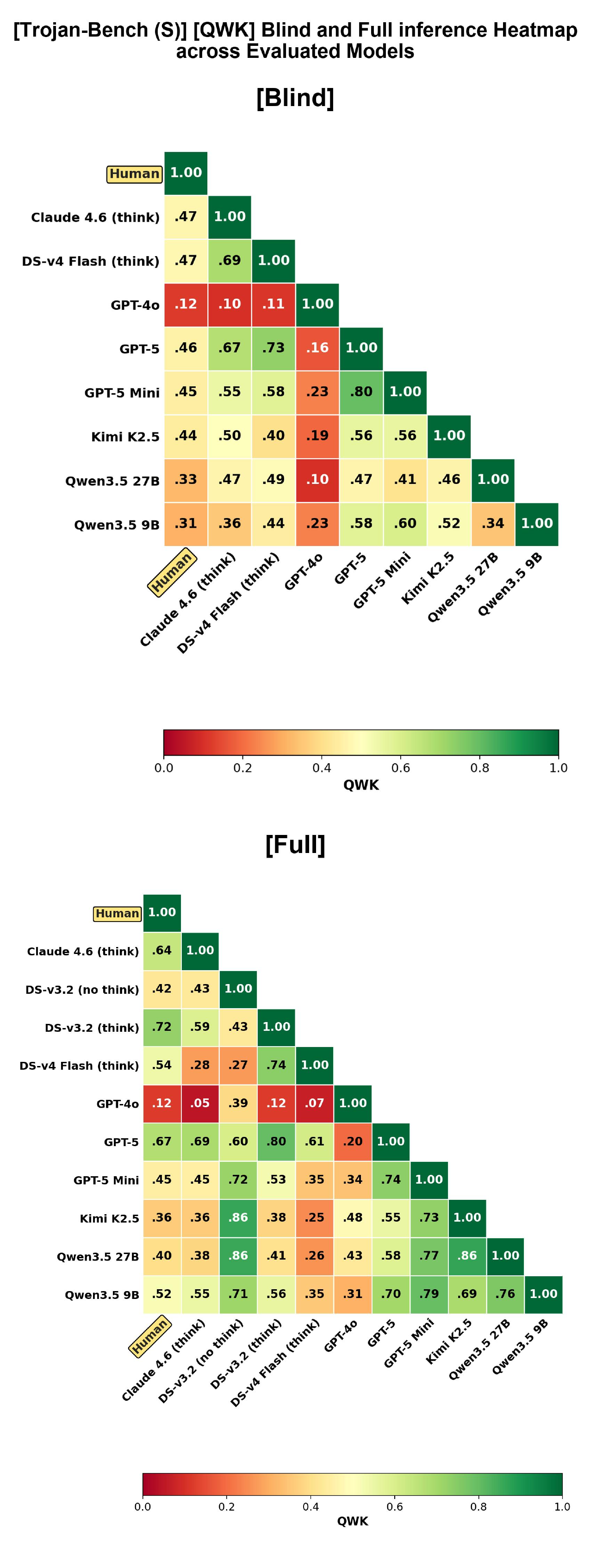}
    \caption{\textbf{\synthds}: QWK blind and full inference heatmap across evaluated models (Majority Vote; top: Blind, bottom: Full). Note that "human" in this heatmap indicates an LLM assigned grade during data generation with \synthmethod and serves as a reference point when evaluating the LLMs.}
    \label{fig:heatmap_synthetic_200_qwk}
\end{figure}

\begin{figure}[htbp]
    \centering
    \includegraphics[width=0.5\textwidth]{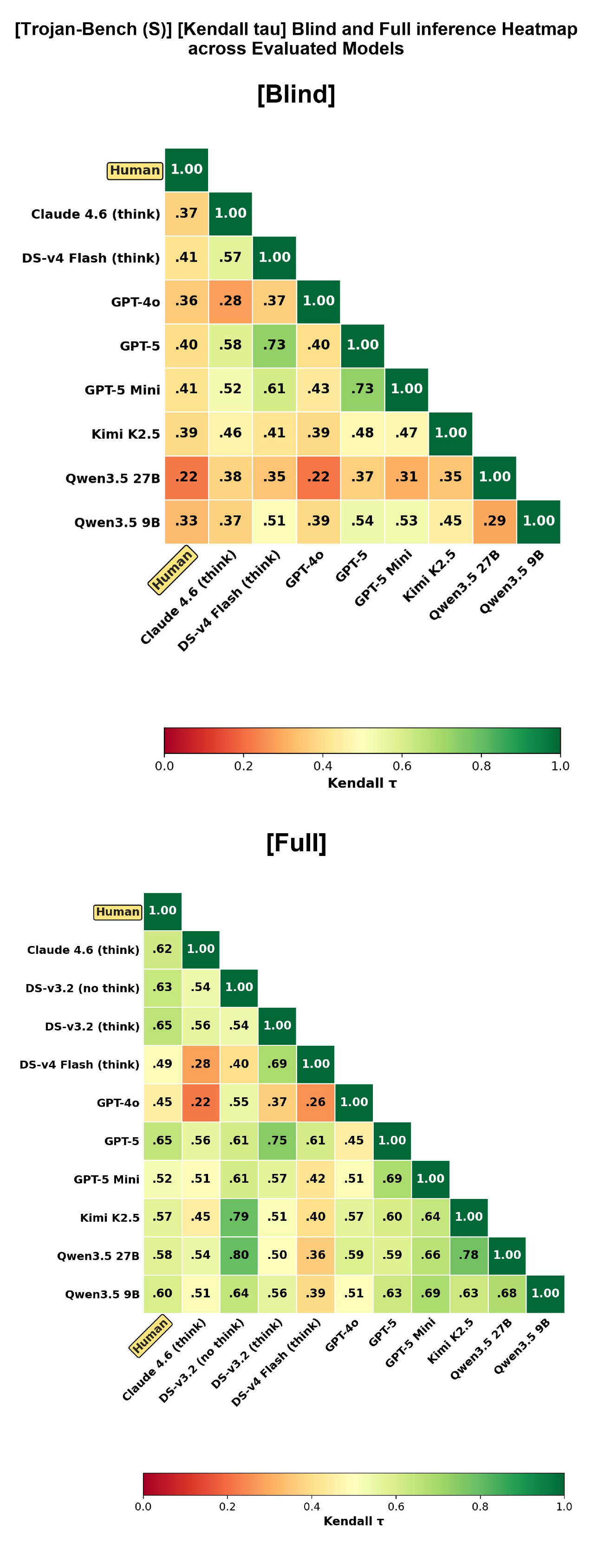}
    \caption{\textbf{\synthds}: Kendall $\tau$ blind and full inference heatmap across evaluated models (Majority Vote; top: Blind, bottom: Full). Note that "human" in this heatmap indicates an LLM assigned grade during data generation with \synthmethod and serves as a reference point when evaluating the LLMs.}
    \label{fig:heatmap_synthetic_200_kendall_tau}
\end{figure}

\begin{figure}[htbp]
    \centering
    \includegraphics[width=0.5\textwidth]{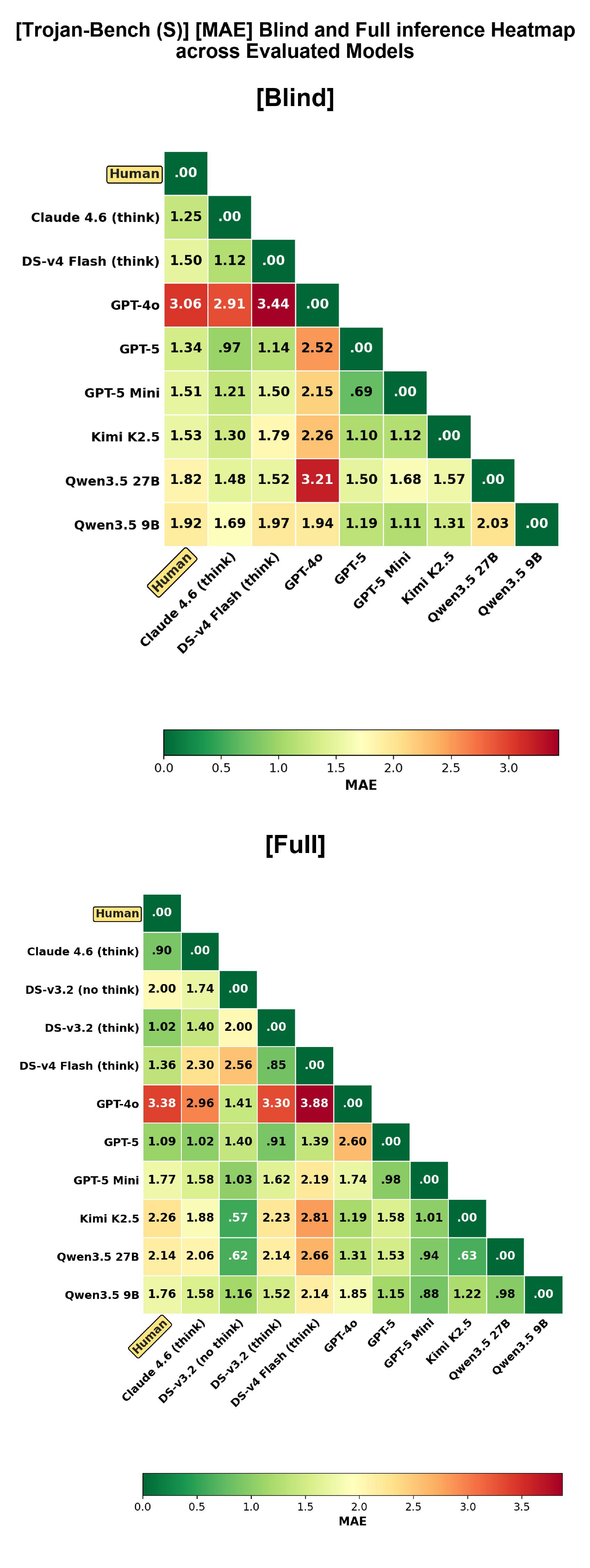}
    \caption{\textbf{\synthds}: MAE blind and full inference heatmap across evaluated models (Majority Vote; top: Blind, bottom: Full). Note that "human" in this heatmap indicates an LLM assigned grade during data generation with \synthmethod and serves as a reference point when evaluating the LLMs.}
    \label{fig:heatmap_synthetic_200_mae}
\end{figure}

\begin{figure}[htbp]
    \centering
    \includegraphics[width=0.5\textwidth]{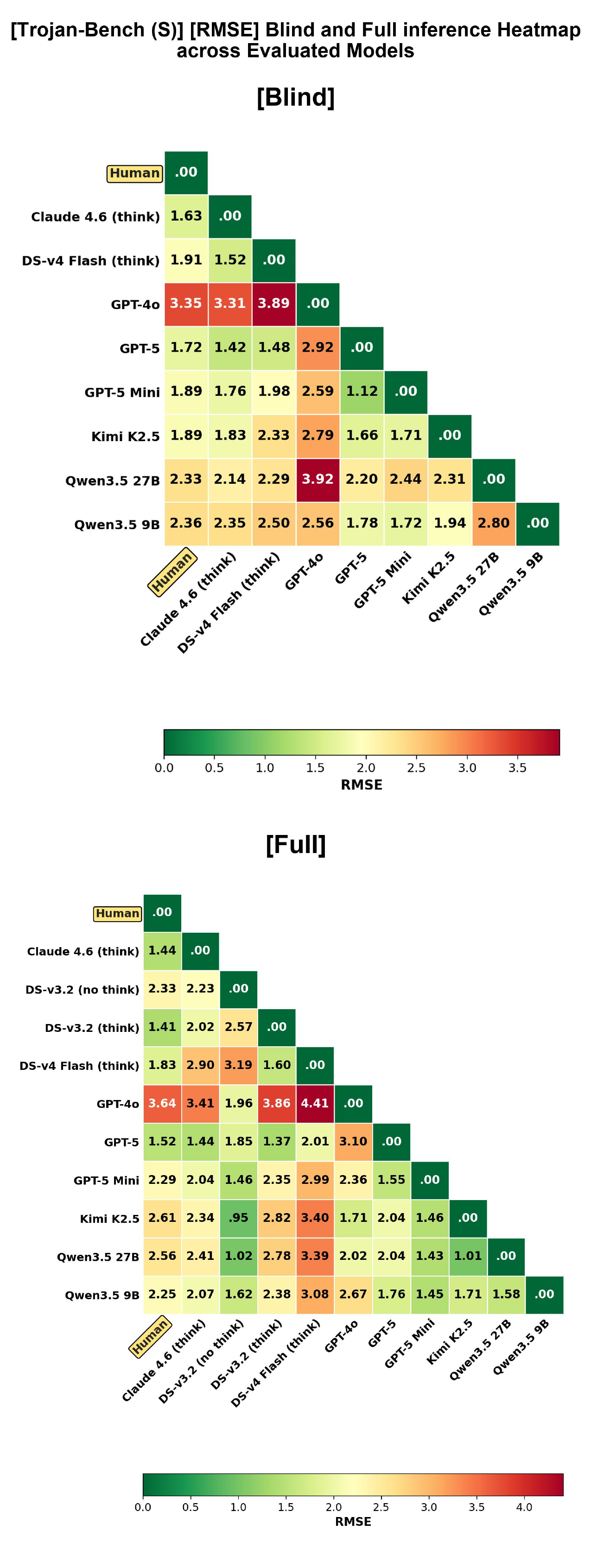}
    \caption{\textbf{\synthds}: RMSE blind and full inference heatmap across evaluated models (Majority Vote; top: Blind, bottom: Full). Note that "human" in this heatmap indicates an LLM assigned grade during data generation with \synthmethod and serves as a reference point when evaluating the LLMs.}
    \label{fig:heatmap_synthetic_200_rmse}
\end{figure}

\begin{figure}[htbp]
    \centering
    \includegraphics[width=0.5\textwidth]{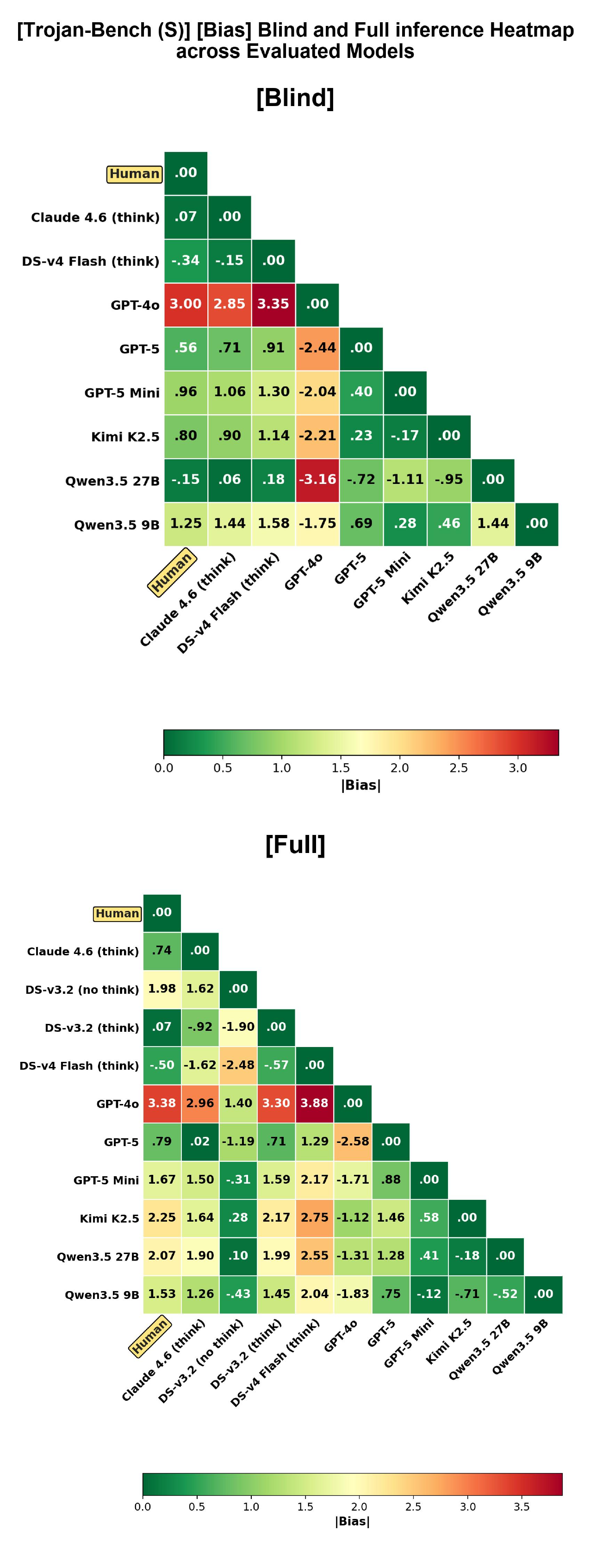}
    \caption{\textbf{\synthds}: Bias blind and full inference heatmap across evaluated models (Majority Vote; top: Blind, bottom: Full). Note that "human" in this heatmap indicates an LLM assigned grade during data generation with \synthmethod and serves as a reference point when evaluating the LLMs.}
    \label{fig:heatmap_synthetic_200_bias}
\end{figure}

\begin{figure}[htbp]
    \centering
    \includegraphics[width=0.5\textwidth]{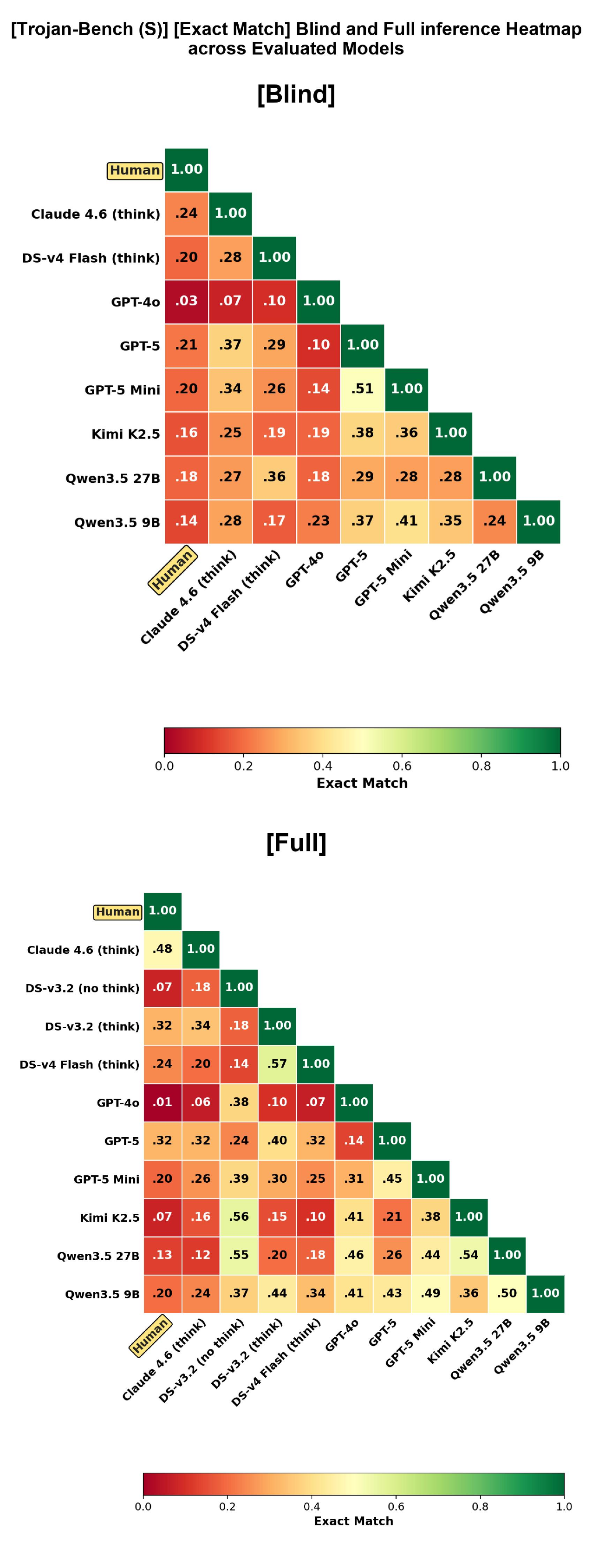}
    \caption{\textbf{\synthds}: Exact Match blind and full inference heatmap across evaluated models (Majority Vote; top: Blind, bottom: Full). Note that "human" in this heatmap indicates an LLM assigned grade during data generation with \synthmethod and serves as a reference point when evaluating the LLMs.}
    \label{fig:heatmap_synthetic_200_exact_match}
\end{figure}

\begin{figure}[htbp]
    \centering
    \includegraphics[width=0.5\textwidth]{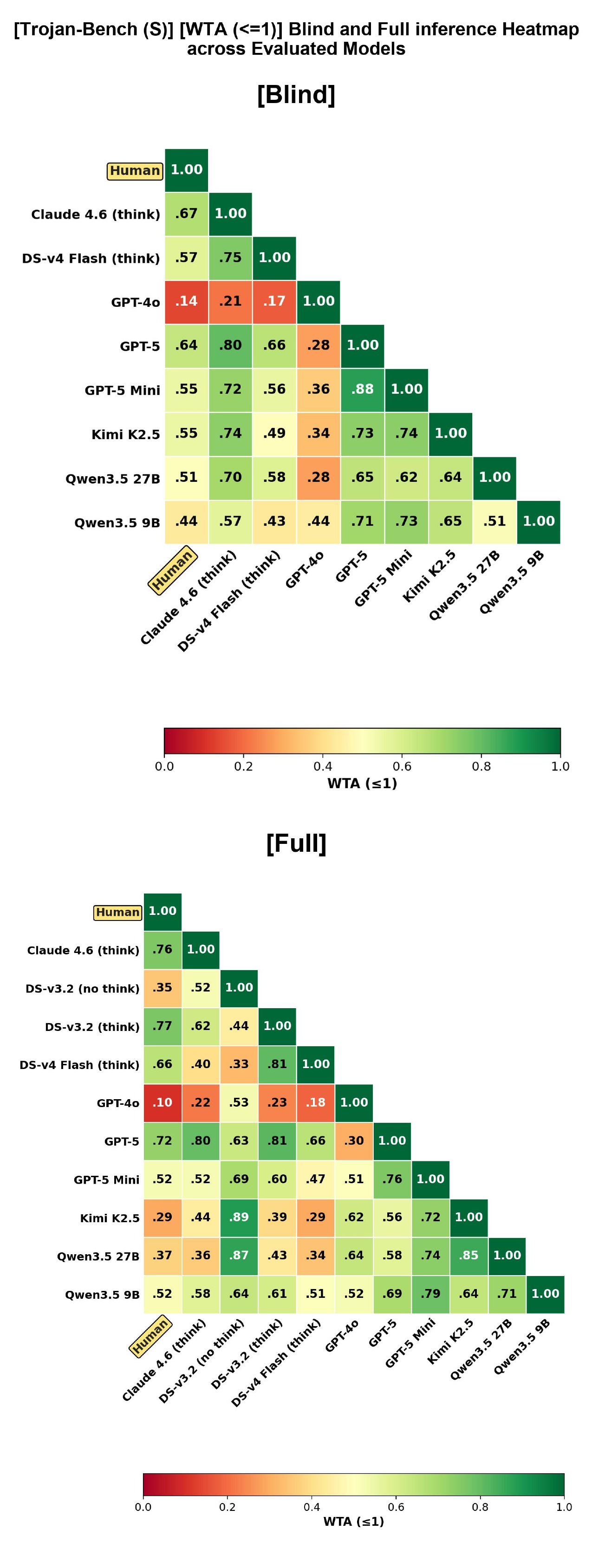}
    \caption{\textbf{\synthds}: WTA ($\leq 1$) blind and full inference heatmap across evaluated models (Majority Vote; top: Blind, bottom: Full). Note that "human" in this heatmap indicates an LLM assigned grade during data generation with \synthmethod and serves as a reference point when evaluating the LLMs.}
    \label{fig:heatmap_synthetic_200_wta_leq1}
\end{figure}

The heatmaps in Figures~\ref{fig:heatmap_usamo_2025_pearson_r}--\ref{fig:heatmap_synthetic_200_wta_leq1} report pairwise agreement between every pair of graders in our suite (LLM-vs-LLM cells) together with their agreement with the human reference (LLM-vs-human row and column). Each figure covers one metric and one dataset, with the upper half reporting the Blind template and the lower half reporting the Full template. We include them so that the inter-grader correlation structure is visible at a glance, since a single number per (grader, metric) cell does not capture how strongly graders agree with each other rather than with the reference. All statistics cited in this section are recomputed from the MajorityVote@16 sampling that produced the heatmaps, and the discussion is restricted to the 14 graders run on \nospacehumands\ in both templates, and the 8 graders run on \nospacesynthds.

The most striking pattern across both datasets is that hiding the markscheme degrades ordinal agreement with the human reference far more than it degrades the magnitude of grading error. On \nospacehumands\ the mean Human-vs-LLM Pearson $r$ drops from $0.50$ to $0.39$ ($\Delta = -0.11$), Kendall $\tau$ from $0.42$ to $0.30$, and QWK from $0.41$ to $0.33$, yet mean absolute error is essentially unchanged ($1.99$ vs $2.00$) and RMSE actually edges down slightly ($2.71 \to 2.60$). The \nospacesynthds\ set shows the same dissociation more sharply: Pearson $r$ collapses by $0.21$ ($0.67 \to 0.46$) and Kendall $\tau$ by $0.20$, while MAE improves under Blind ($1.83 \to 1.74$). Removing the markscheme does not cause judges to pick wildly different absolute numbers, since they still place most attempts within roughly one point of the reference, but it strips them of the ability to order attempts correctly.

A second pattern is a near-universal sign flip in bias on \nospacehumands, and a much subtler one on \nospacesynthds. Under the Full setting the average judge is stricter than the human (mean bias $= -1.48$); under Blind that average flips to $+0.51$, a swing of almost two grade points, and 10 of 14 \nospacehumands\ judges individually change sign, including GPT-5 ($-0.98 \to +0.56$), Claude 4.6 thinking ($-1.23 \to +0.13$), DeepSeek v3.2 thinking ($-0.10 \to +1.08$), and DeepSeek v3.2 no-thinking ($-2.08 \to +3.02$). Only Gemini 3 Pro doubles down on harshness without the rubric ($-1.38 \to -1.58$), and GPT-4o together with Qwen3-8B remain stricter than the human in both regimes. On \nospacesynthds, where the reference is itself an LLM-rubric consensus rather than a human grade, only 1 of 8 judges flips bias sign, which confirms that the flip on \nospacehumands\ is a real markscheme-versus-human-style effect rather than an artifact of model calibration. Read together with the previous observation, the markscheme behaves like a penalty list: when it is taken away, judges default to a more generous approach.

The LLM-vs-LLM blocks of the heatmaps reveal that consensus between judges itself depends heavily on the rubric and depends on it more strongly than human alignment does. Under Full, the frontier reasoning graders form a tight high-agreement block, with pairwise Pearson typically above $0.7$ and pairwise QWK above $0.6$, well above their mean agreement with the human reference. Hiding the rubric pulls this block apart: on \nospacehumands\ the mean inter-LLM Pearson $r$ drops from $0.64$ to $0.44$ and its pairwise standard deviation almost doubles ($0.15 \to 0.28$), with DeepSeek v3.2 no-thinking becoming essentially decorrelated from every other judge ($|r| \le 0.30$, including $r = -0.45$ against GPT-4o) and GPT-4o losing all coherence with the strong-reasoning cluster. The \nospacesynthds\ inter-LLM block weakens less under rubric removal (mean $r$ from $0.63$ to $0.54$, $\Delta = 0.09$), and there the gap between LLM-consensus and Human-agreement actually flips sign: with the rubric, LLMs match the synthetic ground truth slightly better than they match each other (gap $-0.04$), but without it they revert to agreeing with each other more than with the reference (gap $+0.09$). Per-model rubric gains are also heterogeneous: frontier reasoning models gain large boosts (DeepSeek v3.2 no-thinking $+0.35$ in $r$, Gemini 2.5 Flash $+0.29$, GPT-5 $+0.18$), whereas Qwen3-8B, Qwen3.5-27B, and Kimi-K2.5 actually lose a small amount of human alignment when given the rubric. Taken together, the markscheme is not a uniform injection of correct information but a structural scaffold whose benefit scales with the judge's reasoning capacity, and whose absence pushes weaker judges toward random consensus and stronger judges toward systematic leniency. The negative-rubric-gain pattern for small graders is examined further in Section~\ref{app:case_blind_vs_full}.

\clearpage

\section{Annotation procedure}
\label{app:annotation}

\begin{figure}[htbp]
    \centering
    \includegraphics[width=\linewidth]{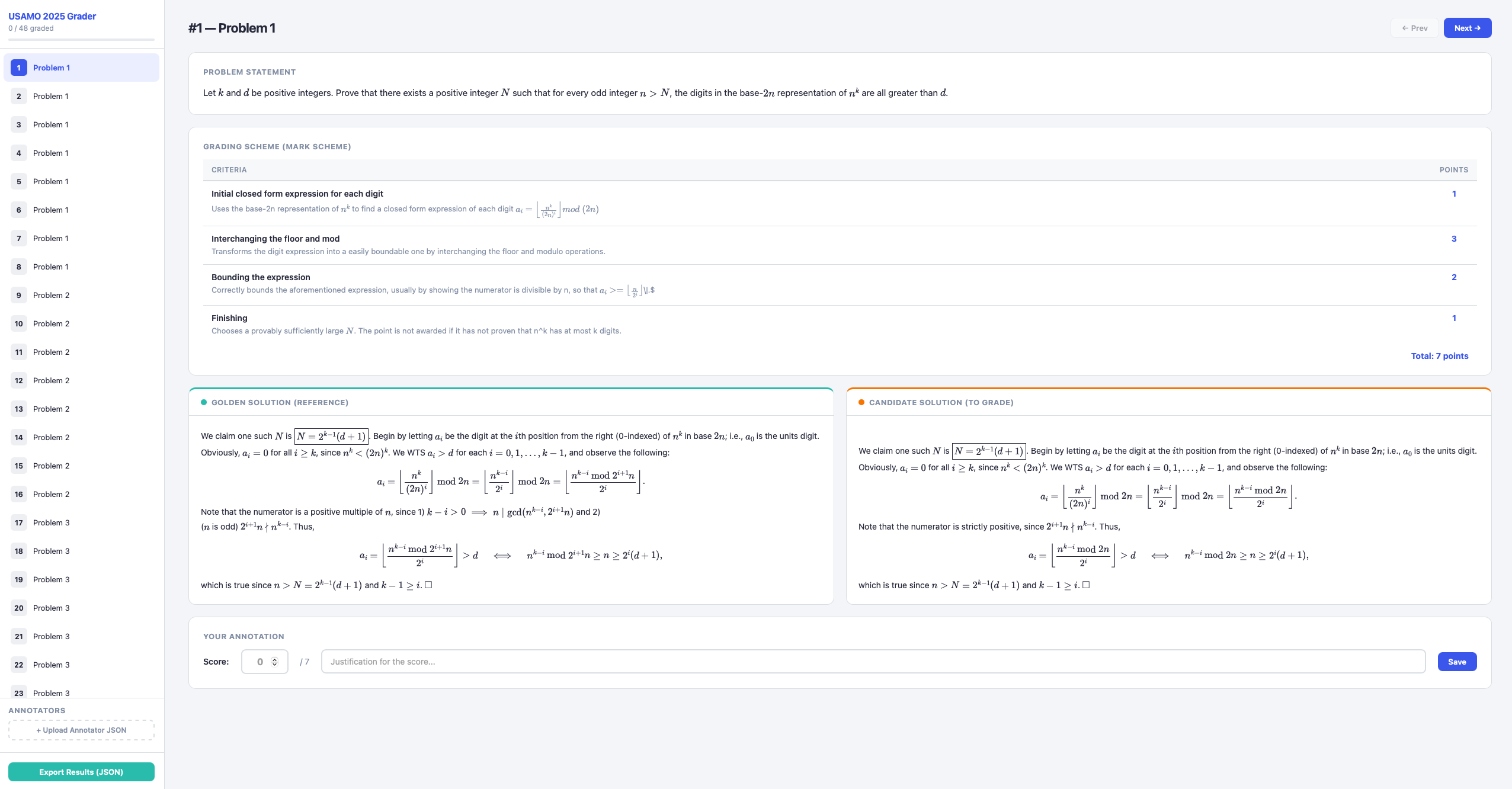}
    \caption{Annotation website used by two domain experts to grade USAMO 2025 candidate solutions. The interface presents the problem statement, marking scheme, golden solution, and candidate solution side by side, and stores per-problem scores together with free-form justifications as JSON.}
    \label{fig:screenshot}
\end{figure}

Figure~\ref{fig:screenshot} shows the custom web-based annotation tool that two domain experts used to grade the 48 candidate solutions of the \nospacehumands\ benchmark. The left panel lists the 48 problems with a counter at the top tracking annotation progress. The main panel shows, in order, the problem statement, the official marking scheme broken into criteria and their point values, the golden reference solution, and the candidate solution to grade. The annotator selects an integer score in $[0,7]$ from a drop-down, writes a free-form justification, and advances to the next problem. The tool stores annotations as JSON, which is the format we used to compute the inter-annotator agreement reported in the main text.

We followed the two-pass procedure of \cite{rein2024gpqa}. Each annotator graded all 48 problems independently. The disagreements were inspected jointly, and a single consensus grade was committed for each problem. The expert-vs-expert agreement statistics are reported in the main text and indicate that the human reference is reliable enough to serve as the gold standard for the LLM evaluation that follows.

\section{Additional details on SFT training}
\label{app:sft}

Tables~\ref{tab:sft-qwen3.5-9b} and~\ref{tab:sft-qwen3.5-27b} document the supervised fine-tuning configurations used to produce the synthetic-data-trained variants of Qwen3.5-9B and Qwen3.5-27B reported in the main text. The 9B model is trained with full fine-tuning, the 27B model with LoRA at rank 8, and both share the data mixture (Full and Blind \nospacesynthds\ samples from four competitions), the optimisation schedule, and the framework (LLaMA-Factory). We include the configurations for reproducibility.


\begin{table}[t]
\centering
\small
\setlength{\tabcolsep}{6pt}
\renewcommand{\arraystretch}{1.05}
\caption{Supervised fine-tuning configuration for Qwen3.5-9B.}
\begin{tabular}{l l}
\toprule
\textbf{Setting} & \textbf{Value} \\
\midrule
\multicolumn{2}{l}{\emph{Model}} \\
\midrule
Base model                       & Qwen3.5-9B \\
Tuning regime                    & Full fine-tuning \\
Precision                        & bfloat16 \\
\midrule
\multicolumn{2}{l}{\emph{Data}} \\
\midrule
Training corpus                  & Synthetic data from four competitions \\
Verification settings            & Full and Blind \\
Train/validation split           & 95/5 \\
Maximum context length           & 16K tokens \\
\midrule
\multicolumn{2}{l}{\emph{Optimization}} \\
\midrule
Peak learning rate               & $1\mathrm{e}{-5}$ \\
LR schedule                      & Cosine \\
Warmup ratio                     & 10\% \\
\midrule
\multicolumn{2}{l}{\emph{Batching \& Schedule}} \\
\midrule
Effective global batch size      & 16 \\
Epochs                           & 2 \\
\midrule
\multicolumn{2}{l}{\emph{System}} \\
\midrule
Training framework               & LLaMA-Factory \\
\bottomrule
\end{tabular}
\label{tab:sft-qwen3.5-9b}
\end{table}


\begin{table}[t]
\centering
\small
\setlength{\tabcolsep}{6pt}
\renewcommand{\arraystretch}{1.05}
\caption{Supervised fine-tuning configuration for Qwen3.5-27B.}
\begin{tabular}{l l}
\toprule
\textbf{Setting} & \textbf{Value} \\
\midrule
\multicolumn{2}{l}{\emph{Model}} \\
\midrule
Base model                       & Qwen3.5-27B \\
Tuning regime                    & LoRA \\
LoRA rank ($r$)                  & 8 \\
Precision                        & bfloat16 \\
\midrule
\multicolumn{2}{l}{\emph{Data}} \\
\midrule
Training corpus                  & Synthetic data from four competitions \\
Verification settings            & Full and Blind \\
Train/validation split           & 95/5 \\
Maximum context length           & 16K tokens \\
\midrule
\multicolumn{2}{l}{\emph{Optimization}} \\
\midrule
Peak learning rate               & $1\mathrm{e}{-4}$ \\
LR schedule                      & Cosine \\
Warmup ratio                     & 10\% \\
\midrule
\multicolumn{2}{l}{\emph{Batching \& Schedule}} \\
\midrule
Effective global batch size      & 16 \\
Epochs                           & 2 \\
\midrule
\multicolumn{2}{l}{\emph{System}} \\
\midrule
Training framework               & LLaMA-Factory \\
\bottomrule
\end{tabular}
\label{tab:sft-qwen3.5-27b}
\end{table}

\clearpage
\section{Hyperparameter $c$ ablation in \method}
\label{app:mab-ablation}

Table~\ref{tab:mab-ablation} reports the effect of the UCB exploration coefficient $c$ on \method when the base grader is DeepSeek-V4-Flash-Thinking. We sweep $c \in \{0.10, 0.25, 0.50, 0.75, 1.00\}$ and report QWK, RMSE, and bias on the human-annotated USAMO 2025 benchmark with $16$ oracle pulls per problem. The metric profile is non-monotone in $c$. Setting $c=0.25$ achieves the best QWK ($0.657$), the lowest RMSE ($2.06$), and a near-best bias ($-1.04$). Larger values ($c \geq 0.5$) over-explore weak arms and degrade all three metrics, while $c=0.10$ under-explores and converges too eagerly to the first arm whose oracle reward looks plausible. We therefore use $c=0.25$ for all main-text \method results, paired with $16$ iterations as documented in Section~\ref{sec:approach}.

\begin{table}[t]
    \centering
    \caption{Parameter $c$ ablation with DeepSeek-V4-Flash-Thinking base model.}
    \begin{tabular}{lrrr}
    \toprule
    UCB exploration $c$ & QWK & RMSE & Bias \\
    \midrule
    $c=0.10$ & 0.641 & 2.38 & -1.17 \\
    $c=0.25$ & 0.657 & 2.06 & -1.04 \\
    $c=0.50$ & 0.634 & 2.59 & -1.34 \\
    $c=0.75$ & 0.613 & 2.69 & -1.40 \\
    $c=1.00$ & 0.616 & 2.46 & -1.47 \\
    \bottomrule
    \end{tabular}
    \label{tab:mab-ablation}
\end{table}

\clearpage
\section{Case studies}

\subsection{Behavior of smaller models on Full and Blind settings}
\label{app:case_blind_vs_full}

A counter-intuitive effect emerges from our \nospacehumands\ evaluation, where the three smaller LLMs we study, i.e., \texttt{Qwen3-8B}, \texttt{GPT-5 Mini}, and \texttt{Kimi-K2.5}, correlate \emph{better} with the human reference under the \textbf{Blind} template (no rubric provided) than under the \textbf{Full} template (problem $+$ golden solution $+$ markscheme). The Blind template strictly removes information from the prompt, so intuitively, the Full template should grade more reliably. Yet on \humands every one of the three small graders has lower MAE and higher QWK under Blind than under Full:

\begin{center}
\footnotesize
\renewcommand{\arraystretch}{1.1}
\begin{tabular}{l|cc|cc|cc|cc}
\toprule
\textbf{Grader} & \multicolumn{2}{c|}{Pearson $r$} & \multicolumn{2}{c|}{Kendall $\tau$} & \multicolumn{2}{c|}{QWK} & \multicolumn{2}{c}{MAE} \\
& Blind & Full & Blind & Full & Blind & Full & Blind & Full \\
\midrule
\texttt{Qwen3-8B}    & \textbf{0.196} & 0.180 & 0.154 & \textbf{0.186} & \textbf{0.175} & 0.076 & \textbf{2.44} & 3.07 \\
\texttt{GPT-5 Mini}  & 0.458 & \textbf{0.519} & 0.337 & \textbf{0.436} & \textbf{0.455} & 0.421 & \textbf{1.77} & 1.92 \\
\texttt{Kimi-K2.5}   & \textbf{0.434} & 0.412 & 0.319 & \textbf{0.328} & \textbf{0.372} & 0.282 & \textbf{1.88} & 2.23 \\
\bottomrule
\end{tabular}
\end{center}

\noindent
The QWK and MAE gaps are large (for example, $+0.13$ QWK and $-0.36$ MAE for Kimi-K2.5). Pearson $r$ and Kendall $\tau$ are also higher with an exception of GPT-5 Mini. QWK and MAE penalise the much larger errors the Full template commits on weak (0 to 2 points) answers.

\paragraph{Six concrete cases (two per model).}
The six text boxes below summarise, for a single \emph{(problem, candidate-solution)} pair at a time, the verdicts the \emph{same} grader returns under the Blind and Full templates. Each box reports the mean grade and the per-instance MAE under each template, gives an aggregated per-instance gap, and excerpts representative free-form text from the grader on each side. The cases were selected as the largest Blind-vs-Full instance gaps within each model on the \humands dataset.

\begin{tcolorbox}[
    colback=gray!2!white,
    colframe=gray!75!black,
    fonttitle=\bfseries,
    coltitle=white,
    title={Case A.\ \texttt{Qwen3-8B} on \humands P3.\ \textbf{Human grade $= 0$.}},
    label={box:case_A}]
\footnotesize

\textbf{Aggregate grading behaviour on this single (problem, candidate-solution) pair.}\\[2pt]
\textcolor{BrickRed}{\textbf{Blind}} (16 attempts): mean grade $\mathbf{0.85}$, instance MAE against the human reference $0.80$.\\
\textcolor{ForestGreen}{\textbf{Full}} (16 attempts): mean grade $\mathbf{5.75}$, instance MAE against the human reference $5.75$.\\
\textbf{Per-instance gap: \textcolor{red!70!black}{$+4.90$~pts} of inflation under the full template.}

\vspace{0.4em}\hrule\vspace{0.4em}

\begin{multicols}{2}

\textcolor{BrickRed}{\textbf{Blind grader (representative response).}}
\textit{"The solution is \textbf{fundamentally flawed}. The key error lies in the definition of the Bob-graph, which is unrelated to the problem's road construction condition. \ldots\ The disk with diameter $AB$ (radius $0.5$) cannot contain any other cities due to the 1-unit distance constraint, making the Bob-graph empty and disconnected. The claim that the graph is connected and planar is \textbf{invalid}, and the conclusion that Bob wins is \textbf{incorrect}."}

\columnbreak

\textcolor{ForestGreen}{\textbf{Full grader (representative response).}}
Awards the rubric items for \texttt{Connectivity}, \texttt{Finiteness of shortest paths}, and \texttt{Planarity}, with reasons such as "\textit{the construction is correct and induces a graph that is connected}" and "\textit{the construction is correct and induces a planar graph}". Only the \texttt{Finishing} item is denied, on the grounds that the solution "\textit{incorrectly concludes that Bob wins.}"
\end{multicols}

\vspace{-0.3em}
\textbf{What this case shows.} The same model awards positive credit to three rubric items (connectivity, finiteness, planarity) whose underlying claims it has just declared invalid under the blind template. The model mathces general approach to the markscheme without deeper analysis on the correctness of the response.
\end{tcolorbox}

\begin{tcolorbox}[
    colback=gray!2!white,
    colframe=gray!75!black,
    fonttitle=\bfseries,
    coltitle=white,
    title={Case B.\ \texttt{Qwen3-8B} on \humands P5.\ \textbf{Human grade $= 0$.}},
    label={box:case_B}]
\footnotesize

\textbf{Aggregate grading behaviour on this single (problem, candidate-solution) pair.}\\[2pt]
\textcolor{BrickRed}{\textbf{Blind}} (16 attempts): mean grade $\mathbf{2.23}$, instance MAE against the human reference $2.23$.\\
\textcolor{ForestGreen}{\textbf{Full}} (16 attempts): mean grade $\mathbf{6.94}$, instance MAE against the human reference $6.94$. The full template returned a perfect grade in nearly every attempt.\\
\textbf{Per-instance gap: \textcolor{red!70!black}{$+4.71$~pts} of inflation under the full template.}

\vspace{0.4em}\hrule\vspace{0.4em}

\begin{multicols}{2}

\textcolor{BrickRed}{\textbf{Blind grader (representative response).}}\\[2pt]
\textit{"The solution correctly identifies that odd $k$ fail by testing $n=2$, but the proof for even $k$ is \textbf{fundamentally flawed}. The key error lies in the induction step where the solution incorrectly shifts the problem to a different sum and fails to address the original requirement that $\frac{1}{n+1}\sum_{i=0}^n \binom{n}{i}^k$ must be an integer. The inductive argument \textbf{does not establish divisibility by $n+1$}, and the reindexing step is \textbf{not rigorously justified}, rendering the entire solution incorrect."}

\columnbreak

\textcolor{ForestGreen}{\textbf{Full grader (representative response).}}\\[2pt]
Awards every rubric item its maximum credit. The reduction line accepts that the solution "\textit{Uses Chinese Remainder Theorem to reduce the problem to a system of equations modulo $p_l^{e_l}$}"; the two remainder lines (for the cases when $p_l$ does not divide $j$ and when $p_l$ divides $j$) are similarly granted on the basis that the solution "\textit{represents the term $\binom{n}{i}$ as a product of fractions $\frac{n-j}{j}$ and finds the remainder of the terms}".
\end{multicols}

\vspace{-0.3em}
\textbf{What this case shows.} A free-form rejection ("the inductive argument does not establish divisibility by $n+1$") is overridden by surface-form pattern matching. 
\end{tcolorbox}

\begin{tcolorbox}[
    colback=gray!2!white,
    colframe=gray!75!black,
    fonttitle=\bfseries,
    coltitle=white,
    title={Case C.\ \texttt{GPT-5 Mini} on \humands P6.\ \textbf{Human grade $= 0$.}},
    label={box:case_C}]
\footnotesize

\textbf{Aggregate grading behaviour on this single (problem, candidate-solution) pair.}\\[2pt]
\textcolor{BrickRed}{\textbf{Blind}} (16 attempts): mean grade $\mathbf{0.75}$, instance MAE against the human reference $0.75$.\\
\textcolor{ForestGreen}{\textbf{Full}} (16 attempts): mean grade $\mathbf{5.62}$, instance MAE against the human reference $5.62$.\\
\textbf{Per-instance gap: \textcolor{red!70!black}{$+4.87$~pts} of inflation under the full template.}

\vspace{0.4em}\hrule\vspace{0.4em}

\begin{multicols}{2}

\textcolor{BrickRed}{\textbf{Blind grader (representative response).}}\\[2pt]
\textit{"The submitted solution is incorrect and fails to rigorously prove the claim. The main \textbf{fatal flaw is the very first modeling step}: an edge is defined between person $p$ and cupcake $b$ iff $p$ values $b$ at least $1$. \ldots\ Subsequent arguments (\textbf{application of Hall's theorem}, the entire \textbf{removal/merging procedure}) \textbf{rely on that incorrect model} and contain additional imprecise or false statements."}

\columnbreak

\textcolor{ForestGreen}{\textbf{Full grader (representative response).}}\\[2pt]
Denies the \texttt{Defining a graph} item, since "\textit{the reformulation is incorrect}". Yet partial credit is still awarded on the \texttt{Hall's theorem} item ("\textit{correctly invokes \ldots\ Hall's theorem \ldots\ \textbf{relies on the erroneous graph formulation from the first part}, so full credit is not warranted}") and on the \texttt{Reducing the problem} item ("\textit{essentially correct in idea and form but \textbf{relies on the earlier mistaken definition}}").
\end{multicols}

\vspace{-0.3em}
\textbf{What this case shows.} The full grader \emph{explicitly states} that the Hall's-theorem step and the reduction step rely on a definition it has just declared wrong, yet still grants them partial credit. Under Blind setting, model correctly identifies that the proof structure is incorrect from the beginning and abstains from awarding more points to a fundamentally flawed solution.
\end{tcolorbox}

\begin{tcolorbox}[
    colback=gray!2!white,
    colframe=gray!75!black,
    fonttitle=\bfseries,
    coltitle=white,
    title={Case D.\ \texttt{GPT-5 Mini} on \humands P3.\ \textbf{Human grade $= 1$.}},
    label={box:case_D}]
\footnotesize

\textbf{Aggregate grading behaviour on this single (problem, candidate-solution) pair.}\\[2pt]
\textcolor{BrickRed}{\textbf{Blind}} (16 attempts): mean grade $\mathbf{2.94}$, instance MAE against the human reference $1.94$.\\
\textcolor{ForestGreen}{\textbf{Full}} (16 attempts): mean grade $\mathbf{7.00}$, instance MAE against the human reference $6.00$. The full template returned the maximum grade in every attempt.\\
\textbf{Per-instance gap: \textcolor{red!70!black}{$+4.06$~pts} of inflation under the full template.}

\vspace{0.4em}\hrule\vspace{0.4em}

\begin{multicols}{2}

\textcolor{BrickRed}{\textbf{Blind grader (representative response).}}\\[2pt]
\textit{"\ldots\ The crucial equivalence between the road-construction rule and the empty-diameter-disk condition for edges is \textbf{never justified}. \ldots\ The connectivity proof is not presented cleanly or rigorously: the inequality chain $\delta_i^2 \leq \delta_{i-1}^2 - 1$ is \textbf{asserted without clear derivation}. \ldots\ the missing justification of the crucial equivalence \ldots\ are serious."}

\columnbreak

\textcolor{ForestGreen}{\textbf{Full grader (representative response).}}\\[2pt]
\textit{"The contradiction argument building an infinite sequence of blocking points and the inequality $\delta_i^2 \leq \delta_{i-1}^2 - 1$ is \textbf{clear and valid}, yielding the required contradiction. \ldots\ a \textbf{correct and complete proof}."}\ Every rubric line is granted in full, and this same verdict is returned in every attempt.
\end{multicols}

\vspace{-0.3em}
\textbf{What this case shows.} \emph{Polar-opposite verdicts on the very same sentence.} The blind grader flags the inequality chain $\delta_i^2 \leq \delta_{i-1}^2 - 1$ as "asserted without clear derivation", whereas the full grader, looking at the same response, calls that exact step "clear and valid". It appears that model prefers surface level matching rooted in markscheme, without making an attempt to analyze the underlying statements.
\end{tcolorbox}

\begin{tcolorbox}[
    colback=gray!2!white,
    colframe=gray!75!black,
    fonttitle=\bfseries,
    coltitle=white,
    title={Case E.\ \texttt{Kimi-K2.5} on \humands P6.\ \textbf{Human grade $= 0$.}},
    label={box:case_E}]
\footnotesize

\textbf{Aggregate grading behaviour on this single (problem, candidate-solution) pair.}\\[2pt]
\textcolor{BrickRed}{\textbf{Blind}} (16 attempts): mean grade $\mathbf{0.69}$, instance MAE against the human reference $0.69$.\\
\textcolor{ForestGreen}{\textbf{Full}} (16 attempts): mean grade $\mathbf{7.00}$, instance MAE against the human reference $7.00$. The full template returned the maximum grade in every attempt.\\
\textbf{Per-instance gap: \textcolor{red!70!black}{$+6.31$~pts} of inflation under the full template (the largest gap observed).}

\vspace{0.4em}\hrule\vspace{0.4em}

\begin{multicols}{2}

\textcolor{BrickRed}{\textbf{Blind grader (representative response).}}\\[2pt]
The blind grader enumerates \textbf{six} "Major Issues":
\begin{enumerate}
\setlength\itemsep{0pt}\setlength\parsep{0pt}
\item \textit{"Fundamental Misunderstanding of the Graph Construction"}
\item \textit{"The `First Person Connects to All Bubbles' Claim is asserted without justification and is false."}
\item \textit{"The Inductive Reduction Argument is hand-wavy and potentially incorrect."}
\item \textit{"Case 2 Analysis is Broken."}
\item \textit{"Circular Arrangement Issues."}
\item \textit{"The Core Mathematical Structure is Missing."}
\end{enumerate}
Conclusion: \textit{"\ldots\ deserves at most 1 point for the attempt at structure and induction framework."}

\columnbreak

\textcolor{ForestGreen}{\textbf{Full grader (representative response).}}\\[2pt]
Awards every rubric item in full. The \texttt{Defining a graph} item is justified as "\textit{a viable reformulation that captures the essential structure of the problem}"; the \texttt{Hall's theorem} item is granted because "\textit{the iterative removal process is described correctly}"; and the \texttt{Reducing the problem} item is given on the basis that the solution "\textit{rigorously handles both topological cases \ldots\ merging operations preserve the invariant}". The same verdict is returned in every attempt.
\end{multicols}

\vspace{-0.3em}
\textbf{What this case shows.} The free-form judgement and the rubric judgement \emph{agree on no single statement}. Blind: "the graph construction is deeply flawed". Full: "a viable reformulation". Blind: "the inductive reduction is hand-wavy". Full: "described correctly". Blind: "Case 2 analysis is broken". Full: "rigorously handles both topological cases". The same model produces a near-zero blind mean and a perfect full mean, an instance-level inflation of $\underline{6.31}$ points and the largest gap in our study.
\end{tcolorbox}

\begin{tcolorbox}[
    colback=gray!2!white,
    colframe=gray!75!black,
    fonttitle=\bfseries,
    coltitle=white,
    title={Case F.\ \texttt{Kimi-K2.5} on \humands P4.\ \textbf{Human grade $= 2$.}},
    label={box:case_F}]
\footnotesize

\textbf{Aggregate grading behaviour on this single (problem, candidate-solution) pair.}\\[2pt]
\textcolor{BrickRed}{\textbf{Blind}} (16 attempts): mean grade $\mathbf{1.94}$, instance MAE against the human reference $0.06$.\\
\textcolor{ForestGreen}{\textbf{Full}} (16 attempts): mean grade $\mathbf{7.00}$, instance MAE against the human reference $5.00$. The full template returned the maximum grade in every attempt.\\
\textbf{Per-instance gap: \textcolor{red!70!black}{$+4.94$~pts} of inflation under the full template.}

\vspace{0.4em}\hrule\vspace{0.4em}

\begin{multicols}{2}

\textcolor{BrickRed}{\textbf{Blind grader (representative response).}}\\[2pt]
The blind grader walks through the proof claim by claim and identifies four structural errors:
\begin{itemize}
\setlength\itemsep{1pt}
\item "$AC'CP$ is a trapezoid \ldots\ \textbf{is asserted without clear proof}";
\item the "perpendicular bisector of $AP$ \ldots\ at $M$" argument is \textbf{"unclear"};
\item "$AC' \perp CF$" is wrong; in fact $AC' \parallel CF$ (both $\perp AB$);
\item "$AC'CH'$ is not a rectangle in general, the claim is \textbf{false}".
\end{itemize}
Conclusion: \textit{"the core insight \ldots\ might be salvageable but the proof is broken".}

\columnbreak

\textcolor{ForestGreen}{\textbf{Full grader (representative response).}}\\[2pt]
Awards each rubric item in full. The \texttt{Properties of the construction} item is granted on the explicit claim that the solution "\textit{Proves that $M$ is the circumcenter of $\triangle AFP$}"; the \texttt{Additional construction} and \texttt{Finishing} items are likewise accepted. The same verdict is returned in every attempt.
\end{multicols}

\vspace{-0.3em}
\textbf{What this case shows.} The full grader's rubric line "\texttt{Properties of the construction}: \emph{Proves that $M$ is the circumcenter of $\triangle AFP$}" is \emph{exactly the claim} that the blind grader has just identified as \emph{not proved} (the trapezoid step, the perpendicular-bisector step, and the rectangle step are all flagged as missing or false).
\end{tcolorbox}

\paragraph{Global vs.\ local scoring.} The Blind template asks the model to "read the proof; output one number $0$--$7$", which forces an \emph{integrating} operation. A fatal flaw in step~2 must reduce the credit step~5 contributes. The Full template asks "for each rubric item, decide if the student earned the points", which factors the task into \emph{independent} per-line judgements. Small graders are good at local pattern matching, for instance "does the student use the words \emph{Hall's theorem}? yes $\to$ award the Hall's theorem point" (Cases C and E). They are weak at compositional reasoning, namely "step~5 uses the bipartite graph from step~1; step~1 was wrong; therefore step~5 is also wrong". The Full template omits the compositional task by design, exposing the small grader's weakness. Larger LLMs such as Claude Sonnet 4.6, GPT-5, and DeepSeek-v4-Flash propagate the error across rubric items since they consider local and global grading patterns together.

\paragraph{The rubric is a high-precision surface-form matcher.} Every rubric line is, by construction, a description of the \emph{correct} proof's structure: "Defining a graph", "Hall's theorem", "Reducing the problem", "Connectivity of construction", "Showing the graph is planar", and so on. A student who reaches for the right tools but applies them incorrectly produces text that contains exactly those keywords, so the surface-form match is near-perfect even when the substance is wrong. In Cases A, C and E the wrong proofs all \emph{look like} the right proof. They reach for the right theorem (Hall, induction, planarity) and follow the right outline. Since the smaller models are less capable in uncovering the smaller failures in each claim, they inflate the final score. In Blind setting, there is no markscheme to base the reasoning on, so it forces to inspect proofs closer, which inadvertently improves the performance compared to the Full setting.

\paragraph{Reasoning revolves too much around the rubric and not the substance.} When the prompt says "this rubric line is worth $3$ points; explain whether the student earned them", the small instruction-tuned model is biased toward awarding most of the $3$. The default behaviour of small instruct-tuned models is to follow the prompt's expectation. The prompt asks the model to "explain how many of the $3$ points the student got", so the model produces a small downward adjustment from the maximum (gives $2$ instead of $3$) rather than zeroing out. The Blind prompt has no such anchor. The model has to commit to a number from $\{0,\ldots,7\}$ after reading the proof, and the natural anchor becomes the proof's actual quality. This is why the Blind grader can output a low value confidently when it sees a wrong proof, whereas the Full grader output for the same proof rarely falls into the lower grades territory, such as in Cases A, C, and E.

In summary, for a frontier grader, the rubric is additional context that informs but does not override compositional judgement. For a small grader, the rubric forces the model's attention to it, and the LLM behaves more like pattern matching and less like a proof verifier. This effect is not present in larger models, since they verify the proof and match the reasoning together with a markscheme that provides additional context to how to properly distribute the grade. Strong vs. weak compositional ability is exactly what separates the two evaluation scenarios.

\subsection{Effect of \method on grading procedures}
\label{app:case_mab_vs_cot}

We now study how \method\ corrects the failure modes of vanilla chain-of-thought (CoT) grading on the same proofs. Both methods use the \texttt{Gemini-3-Pro-Think} model, the same problems and student solutions from \nospacehumands, and the same Full-style prompt (problem, marking scheme, sample solution). The only difference is the inference procedure: vanilla CoT corresponds to a single grading call (with $16$ attempts for the MajorityVote), whereas \method runs the UCB procedure with $c=0.25$, $8$ arms (one per candidate grade in $\{0,\ldots,7\}$), and $16$ oracle pulls. The comparison is over the $48$ datapoints.

\paragraph{Aggregate metrics.}
\method improves every aggregate metric. The biggest shift is in Bias: CoT is consistently optimistic (mean signed error $+1.383$ grade points), whereas \method is closest to zero (mean signed error $+0.340$). The QWK lift of $+0.171$ is the largest improvement we observe from any prompt-engineering or aggregation (MajorityVote, Best-of-N) methods in this paper.

\begin{center}
\footnotesize
\renewcommand{\arraystretch}{1.1}
\begin{tabular}{l|c|c|c|c|c|c}
\toprule
\textbf{Method} & EM & MAE & RMSE & Bias & Pearson $r$ & QWK \\
\midrule
Vanilla CoT (single attempt)         & $0.362$ & $1.660$ & $2.502$ & $+1.352$ & $0.511$ & $0.449$ \\
Vanilla CoT ($16$-shot majority vote) & $0.340$ & $1.681$ & $2.497$ & $+1.383$ & $0.515$ & $0.454$ \\
\method\ (UCB, $c=0.25$)             & $\mathbf{0.447}$ & $\mathbf{1.277}$ & $\mathbf{1.957}$ & $\mathbf{+0.340}$ & $\mathbf{0.633}$ & $\mathbf{0.625}$ \\
\bottomrule
\end{tabular}
\end{center}

\noindent
Two observations follow. First, the $16$-shot MajorityVote variant is no better (and slightly worse on EM) than the single-attempt baseline. The 16 CoT samples agree with each other on $38$ of $48$ datapoints, so voting cannot denoise a structural bias. Second, the gain in instance-level MAE is asymmetric across the human-grade range. Vanilla CoT is monotonically lenient on the lower band of the scale (mean signed error $+2.60$ at human grade $0$, $+2.67$ at human grade $2$) and slightly lenient on the upper band, with perfect proofs graded correctly. \method better matches grades on the lower end (mean signed error drops to $+0.40$ at human grade $0$ and to $+1.00$ at human grade $2$), while introducing a small under-grading of about $-1$ point at human grades $4$, $5$, and $6$.

\paragraph{Two concrete cases.}
The two text boxes below illustrate the two failure modes vanilla CoT exhibits on this benchmark. Each box reports the predicted grade and the per-instance MAE for both methods, and excerpts representative free-form text from the model in each role. The grading CoT was compressed for better readability.

\begin{tcolorbox}[
    colback=gray!2!white,
    colframe=gray!75!black,
    fonttitle=\bfseries,
    coltitle=white,
    title={Case G.\ \texttt{Gemini-3-Pro-Think} on \humands P6.\ \textbf{Human grade $= 0$.}},
    label={box:case_G}]
\footnotesize

\textbf{Aggregate grading behaviour on this single (problem, candidate-solution) pair.}\\[2pt]
\textcolor{BrickRed}{\textbf{Vanilla CoT}} (16 attempts): mean grade $\mathbf{5.25}$, MajorityVote $6$, instance MAE against the human reference $\mathbf{6}$.\\
\textcolor{ForestGreen}{\textbf{\method\ (UCB, $c=0.25$)}}: selected grade $\mathbf{0}$, instance MAE against the human reference $\mathbf{0}$.\\
\textbf{Per-instance error reduction under \method: \textcolor{red!70!black}{$-6$~pts}.}

\vspace{0.4em}\hrule\vspace{0.4em}

\begin{multicols}{2}

\textcolor{BrickRed}{\textbf{Vanilla CoT (representative response).}}\\[2pt]
The model correctly penalises the foundational definitional error on the rubric line for "Defining a graph" but then awards full credit on the two downstream items. A representative justification reads: "\textit{\textbf{Despite the initial setup error}, the solution correctly treats the removed `bubbles' as intervals in this step and rigorously shows how the remaining people can merge their intervals to maintain the condition.}" The samples cluster tightly, since the bias lives at the level of how the rubric template is applied rather than at the level of token-sampling noise.

\columnbreak

\textcolor{ForestGreen}{\textbf{\method\ (representative arm justifications).}}\\[2pt]
The grade-$0$ arm correctly justifies the cascading-failure argument: "\textit{because $G$ is defined between $n$ people and $m$ individual cupcakes \ldots\ the application of Hall's theorem here is meaningless. \ldots\ Applying a valid theorem to an invalid, broken graph earns no points.}" The grade-$6$ arm has to invoke an unsupported "\textit{the student implicitly corrects their definitional error mid-proof}" move, which the source text does not support; the oracle assigns full reward to the grade-$0$ arm and minimal reward to the high-grade arms.
\end{multicols}

\vspace{-0.3em}
\textbf{What this case shows.} The vanilla-CoT failure mode is \emph{atomistic rubric application}: the model sums per-rubric credit independently and ignores the cascading consequence of the foundational error. \method\ structurally counters this and correctly points out the accumulation of errors. The CoT samples cluster tightly, so any sampling-based ensemble (MajorityVote, Best-of-N) cannot escape the bias.
\end{tcolorbox}

\begin{tcolorbox}[
    colback=gray!2!white,
    colframe=gray!75!black,
    fonttitle=\bfseries,
    coltitle=white,
    title={Case H.\ \texttt{Gemini-3-Pro-Think} on \humands P5.\ \textbf{Human grade $= 5$.}},
    label={box:case_H}]
\footnotesize

\textbf{Aggregate grading behaviour on this single (problem, candidate-solution) pair.}\\[2pt]
\textcolor{BrickRed}{\textbf{Vanilla CoT}} (16 attempts): mean grade $\mathbf{7.00}$ with zero variance across attempts, instance MAE against the human reference $\mathbf{2}$.\\
\textcolor{ForestGreen}{\textbf{\method\ (UCB, $c=0.25$)}}: selected grade $\mathbf{5}$, instance MAE against the human reference $\mathbf{0}$.\\
\textbf{Per-instance error reduction under \method: \textcolor{red!70!black}{$-2$~pts}.}

\vspace{0.4em}\hrule\vspace{0.4em}

\begin{multicols}{2}

\textcolor{BrickRed}{\textbf{Vanilla CoT (representative response).}}\\[2pt]
The CoT samples agree exactly on the maximum grade. One of them notices the off-by-one in the binomial reindexing step but labels it "\textit{a minor typographical error \ldots\ a transcription slip rather than a conceptual misunderstanding}". Once the issue is labelled a typo, the rubric cell auto-fills with full credit. The spread across attempts is zero, so no sampling-based ensemble can recover the human grade.

\columnbreak

\textcolor{ForestGreen}{\textbf{\method\ (representative arm justifications).}}\\[2pt]
The grade-$7$ arm reads the off-by-one as "\textit{clearly a transcription slip}". The grade-$5$ arm reads the same fact as "\textit{a critical algebraic error \ldots\ the inductive hypothesis cannot be directly applied, leaving a significant logical gap in the proof}". The oracle returns full reward for both readings (each is plausible in isolation); UCB tie-breaking returns the lower arm id, which corresponds to the stricter (and correct) human grade.
\end{multicols}

\vspace{-0.3em}
\textbf{What this case shows.} The vanilla-CoT failure mode here is \emph{over-charitable interpretation}: a real algebraic error is reframed as a typo, lifting the grade to the maximum. Because the CoT samples agree for the most part, the failure is deterministic, not stochastic, and no sampling ensemble can correct it. \method produces two equally plausible defences and lets the algorithm choose a better grading arm.
\end{tcolorbox}

\paragraph{Hypothesis enumeration via arms.} Vanilla CoT is a free generation: the model decides on the spot what the grade should be, writing a chain of thought that can drift toward a conclusion which is then rationalised against the rubric. The arm-generation phase of \method\ does the opposite. It runs $8$ separate calls, each of which assumes the answer is grade $k$ and produces the strongest possible defence of grade $k$. Low-grade defences end up surprisingly compelling because they do not have to wave away foundational errors (Case G). The model is now arguing against its own default rather than rationalising it.

\paragraph{Oracle as adversarial verifier.} The oracle is a separate model call that scores each arm's defence on a Likert scale, mapped to a $[0,1]$ reward. The oracle does not have to commit to a number; it just has to ask whether each defence holds up against the source. That framing is much easier for the model than open-ended grading. In Case~H the oracle returns full reward for both the lenient and the strict reading of the off-by-one, since both are individually plausible.

\paragraph{UCB tie-breaking is a strict-grading prior.} When several arms tie at maximum reward, UCB returns the arm with the lowest id. Because arm ids equal candidate grades, "lowest id" means "lowest grade". When the evidence is ambiguous between "give credit" and "do not give credit", a professional grader defaults to the latter (the proof should remove ambiguity). In Case~H this tie-breaker alone is what recovers the correct human grade.

\paragraph{Why self-consistency cannot close the gap.} A natural counterpoint is that the gain might be variance reduction and that a simpler "sample CoT more times and apply MajorityVote" strategy would close it. The aggregate table refutes this: the $16$-shot MajorityVote variant is slightly worse than a single sample, and the 16 samples agree with each other on the large majority of cells. Token-level sampling on a deterministic prompt with strong prompt-side bias does not produce diverse enough gradings to escape the bias. The diversity in \method comes from the prompt side (eight different "assume answer $= k$" prompts), not from token-level sampling.

In summary, the dominant LLM grading failure on this benchmark is over-grading bias, not stochastic noise. Variance-reduction methods (sampling and voting) do not address bias because the bias is deterministic. 

\section{Inclusion of USAMO 2025 samples in the training set}

To further validate the effect of \nospacesynthmethod-generated samples on the quality of judging after supervised fine-tuning (SFT), we conducted an additional experiment where we fine-tuned a base model without USAMO 2025 samples and tested it exclusively on USAMO 2025. This experiment is designed to see whether the synthetic samples can generalize to previously unseen math problems. Table~\ref{tab:sft-experiment-without-usamo-2025} reports results across thre metrics: QWK, RMSE and Bias, and under two settings: Full and Blind. We observe that all metrics improve after finetuning and highlights that the synthetic data may help with generalization on grading problems, even when the test problems come from a completely different competition.

\begin{table*}[t!]
    \centering
    \caption{SFT results with excluded USAMO 2025 datapoints}
    \label{tab:sft-experiment-without-usamo-2025}
\begin{footnotesize}
\begin{tabular}{lc*{7}{c}}
  \toprule
  & \multicolumn{3}{c}{\textbf{Full Setting}} & \multicolumn{3}{c}{\textbf{Blind Setting}} \\
  \cmidrule(lr){2-4}\cmidrule(lr){5-7}
  & QWK ($\uparrow$) & RMSE ($\downarrow$) & Bias ($\approx 0$) & QWK ($\uparrow$) & RMSE ($\downarrow$) & Bias ($\approx 0$)\\
  \midrule
  Qwen3.5-9B (Base) & 0.02 & 4.00 & 3.23 & 0.06 & 2.52 & 0.63 \\
  Qwen3.5-9B (SFT) & 
  \textbf{0.23} & \textbf{2.19} & \textbf{0.17} & \textbf{0.08} & \textbf{2.32} & \textbf{-0.52} \\
  \bottomrule
\end{tabular}
\end{footnotesize}
\end{table*}


\section{Effect of changing the generator model in \synthmethod pipeline}

To further extend our analysis of \nospacesynthmethod, we swapped the base Claude Opus 4.6 model for the smaller and less capable DeepSeek V4 Flash. The results are presented in Table~\ref{tab:sft-experiment-dsv4f-base}. To exclude the effects of memorization and overfitting, we excluded the USAMO 2025 samples. Our tests suggest that using both models improves the performance of the underlying weaker 9B model. In Blind settings Flash model exhibits better performance, while Opus is generally better under Full setting evaluation. We emphasize that using both models lead to a bump in performance comapred to the base Qwen-3.5-9B model.

\begin{table*}[t!]
    \centering
    \caption{Comparative performance of base and fine-tuned models on \synthmethod synthetic samples coming from DeepSeek-V4-Flash (DSV4F) model and larger Claude Opus 4.6 (Opus-4.6). Best results are \textbf{bolded}.}
    \label{tab:sft-experiment-dsv4f-base}
\begin{footnotesize}
\begin{tabular}{lc*{7}{c}}
  \toprule
  & \multicolumn{3}{c}{\textbf{Full Setting}} & \multicolumn{3}{c}{\textbf{Blind Setting}} \\
  \cmidrule(lr){2-4}\cmidrule(lr){5-7}
  & QWK ($\uparrow$) & RMSE ($\downarrow$) & Bias ($\approx 0$) & QWK ($\uparrow$) & RMSE ($\downarrow$) & Bias ($\approx 0$)\\
  \midrule
  Qwen3.5-9B (Base) & 0.02 & 4.00 & 3.23 & 0.06 & 2.52 & 0.63 \\
  Qwen3.5-9B (SFT-Opus-4.6) & 
  \textbf{0.35} & \textbf{2.08} & \textbf{0.27} & 0.07 & 2.40 & 0.39 \\
  Qwen3.5-9B (SFT-DSV4F) & 
  0.23 & 2.18 & \textbf{0.27} & \textbf{0.17} & \textbf{2.29} & \textbf{-0.23} \\
  \midrule
  Qwen3.5-27B (Base) & 0.36 & 2.98 & 1.54 & -0.11 & 4.12 & -3.40 \\
  Qwen3.5-27B (SFT-Opus-4.6) & \textbf{0.47} & 2.10 & 0.77 & \textbf{0.06} & 2.50 & \textbf{0.21} \\
  Qwen3.5-27B (SFT-DSV4F) & 0.40 & \textbf{1.93} & \textbf{0.33} & \textbf{0.06} & \textbf{2.42} & 0.27 \\
  \bottomrule
\end{tabular}
\end{footnotesize}
\end{table*}


\section{Benchmarking \synthmethod SFT pipeline against ProofBench}

Table~\ref{tab:sft-experiment-proofbench} presents additional evaluation results on ProofBench~\cite{DBLP:journals/corr/abs-2510-13888}, a dataset of LLM-generated proofs for competition-level mathematics paired with expert human grade annotations. For this experiment, we fine-tuned two models (Qwen3.5-9B and Qwen3.5-27B) on synthetic samples generated by our \synthmethod pipeline, mirroring the setup described in the main text. We strictly excluded any overlap between the ProofBench dataset and our training set. This ensures that performance improvements stem from genuine learned grading capabilities rather than data contamination or overfitting. We report results for both the Full setting (which provides access to the markscheme and golden solution) and the Blind setting (which provides neither).

We observe that despite being fine-tuned exclusively on synthetic data, the models achieve correlation levels with ground-truth labels that match or exceed baselines, alongside a general reduction in overall error and bias. This consistent improvement demonstrates that our proposed markscheme-aligned synthetic error injection is both a robust evaluation proxy and an effective method for generating training data that aligns with expert human judgment. While the distributions of synthetic and naturally occurring errors inherently differ, our results reveal a functional overlap. Because the fine-tuned base model shows improved performance on an entirely independent problem set with a different group of expert annotations, it is clear that \nospacesynthmethod's synthetic errors generalize to real-world mistakes and unseen grading tasks. Furthermore, our generation pipeline provides fine-grained control over the exact type and location of injected mistakes, offering a targeted approach to further refine LLM grading capabilities.

\begin{table*}[t!]
    \centering
    \caption{Comparative performance of base and fine-tuned models on \synthmethod synthetic samples on ProofBench benchmark. Best results are \textbf{bolded}.}
    \label{tab:sft-experiment-proofbench}
\begin{footnotesize}
\begin{tabular}{lc*{7}{c}}
  \toprule
  & \multicolumn{3}{c}{\textbf{Full Setting}} & \multicolumn{3}{c}{\textbf{Blind Setting}} \\
  \cmidrule(lr){2-4}\cmidrule(lr){5-7}
  & QWK ($\uparrow$) & RMSE ($\downarrow$) & Bias ($\approx 0$) & QWK ($\uparrow$) & RMSE ($\downarrow$) & Bias ($\approx 0$)\\
  \midrule
  Qwen3.5-9B (Base) & 0.70 & 2.29 & 1.11 & 0.47 & 3.16 & 1.98 \\
  Qwen3.5-9B (SFT)  & \textbf{0.79} & \textbf{1.50} & \textbf{0.15} & \textbf{0.53} & \textbf{2.32} & \textbf{0.93} \\
  \midrule
  Qwen3.5-27B (Base) & 0.80 & 1.81 & 0.41 & \textbf{0.49} & 3.08 & 1.04 \\
  Qwen3.5-27B (SFT)  & \textbf{0.81} & \textbf{1.44} & \textbf{-0.12} & 0.48 & \textbf{2.36} & \textbf{0.28} \\
  \bottomrule
\end{tabular}
\end{footnotesize}
\end{table*}


\section{Effect of Multi-Armed Bandit algorithm on grading}\label{app:effect-of-mab}

\begin{table*}[t!]
    \centering
    \caption{Effect of the multi-armed bandit (UCB) algorithm on grading performance. Baselines include a single-model approach and a uniform sampling strategy, where each arm is pulled exactly $r$ times before a final decision is made. Best results are highlighted in \textbf{bold}.}
    \label{tab:effect-of-mab}
\begin{footnotesize}
\begin{tabular}{lc*{3}{c}}
  \toprule
  & QWK ($\uparrow$) & RMSE ($\downarrow$) & Bias ($\approx 0$) \\
  \midrule
  Qwen3.5-27B (model only) & 0.30 & 3.04 & 1.81 \\
  Qwen3.5-27B (uniform with $r=1$) & 0.40 & 2.78 & -1.48 \\
  Qwen3.5-27B (uniform with $r=5$) & 0.56 & 2.25 & -1.26 \\
  Qwen3.5-27B (UCB) & \textbf{0.57} & \textbf{2.15} & \textbf{-0.98} \\
  \bottomrule
\end{tabular}
\end{footnotesize}
\end{table*}


To isolate the performance gains of the Multi-Armed Bandit (MAB) algorithm from the underlying grade justification pipeline, we conducted an ablation study replacing \nospacemethod's Upper Confidence Bound (UCB) selection policy with a uniform sampling baseline. In the "Uniform" setting, the model generates a justification for each candidate grade, an oracle assigns a confidence score to each, and the system selects the highest-scoring grade. Unlike \nospacemethod, which adaptively allocates its computational budget based on continuous oracle feedback, this baseline samples every arm a fixed number of times (e.g., $r$), entirely removing the UCB exploration-exploitation dynamics. This setting also corresponds to A/B testing. For comparison, we also included a standard baseline where the model directly outputs a grade using MajorityVote@16.

As presented in Table~\ref{tab:effect-of-mab}, the results indicate that the justification-and-scoring mechanism alone significantly improves baseline performance, and increasing the uniform sampling budget further amplifies these gains. However, the complete MABGrader framework remains superior. By dynamically balancing exploration and exploitation, the UCB algorithm converges on the optimal grade much more reliably under similar budget constraints.

\goodbreak\newpage

\section{Prompt templates}
\label{app:templates}

This section reproduces the six prompt templates referenced throughout the paper. The first one is the rubric grading prompt sourced from MathArena~\cite{DBLP:journals/corr/abs-2505-23281}, which we use as our Full-template grading prompt in the human-set evaluation. We include it here for completeness so that our experimental setup can be reproduced from a single document. The next two are ProofAutoGrader and ProofGrader templates under Full settings. 

The remaining three templates implement the three stages of the \synthmethod synthetic-data pipeline (Section~\ref{sec:ds construction}). The error-injection prompt drives the generator that produces a first-pass candidate student solution targeted at a specified score profile and includes a worked USAMO-style example so that the generator has a clear reference for what calibrated student errors look like. The critique prompt drives an independent grader that audits the candidate against the markscheme and outputs a JSON-formatted misalignment report with revision guidance for each rubric criterion. The refinement prompt prompts the LLM again, conditioned on both the candidate solution and the critique above, and produces a minimally-edited revision that is better aligned with the target score profile on each rubric criterion.

\bigskip
\noindent\rule{\linewidth}{0.8pt}\par\nopagebreak
\noindent\textbf{MathArena rubric grading prompt template (\emph{Full} setting).}\par\nopagebreak
\noindent\rule{\linewidth}{0.4pt}\par\nopagebreak
{\footnotesize
\begin{verbatim}
# Instruction

You are an expert mathematician that grades solutions of high-school
olympiad-level problems. You will be given a mathematical problem,
as well as a grading scheme that you should adhere to. Your task is
to accurately grade a solution according to that grading scheme.

# Problem and Scheme

## Problem statement

{problem}

## Grading scheme

{grading_scheme}

# Example

Here is how an example grading looks:

## Example solution:

{sample_solution}

## Example grading:

{sample_grading}

# Evaluation

Produce a scoring using the same format for the following solution.
Reason carefully about the contents of the solution and make sure
that all steps of the proof are included and rigorous.

## Formatting instructions

Follow a JSON-style formatting as the above example, namely:

- The categories should be in the same order.
- The awarded number of points for each category should follow
  after the 'points' key.
- The reasoning for each decision should be explained after
  the 'desc' key.
- Output only the JSON response.
- JSON response should be wrapped in ```json ```.
- Do not output triple slashes or other characters that cause
  JSON decode errors.

Here is the solution you should grade:

{solution}
\end{verbatim}
}\par\noindent\rule{\linewidth}{0.8pt}
\bigskip

\bigskip
\noindent\rule{\linewidth}{0.8pt}\par\nopagebreak
\noindent\textbf{ProofAutoGrader rubric grading prompt template (\emph{Full} setting).}\par\nopagebreak
\noindent\rule{\linewidth}{0.4pt}\par\nopagebreak
{\footnotesize
\begin{verbatim}
You are an expert grader for the International Mathematics Olympiad (IMO). Your task is to
evaluate a proposed solution strictly and rigorously. Keep in mind the standards at the IMO
are extremely high: only arguments that are logically sound, complete, and precise should 
be rewarded.


### General Scoring Rubric
Scores are assigned on a 0-7 scale. The general guidelines are:


**7 Points (Correct):** The solution is complete, correct, and fully rigorous. If the
submission contains incorrect attempts or lines of reasoning but ultimately presents a
complete and correct solution, it should still be awarded full points; the presence of
earlier, discarded work does not detract from the final correct proof.

**6 Points (Almost Correct):** The solution is almost correct with a sound core argument,
but contains minor errors in calculation or small gaps in logic. Missing proofs for major
components, unjustified claims, or sketchy arguments are **not** eligible for 6 points.

**1 Point (Partial Progress):** The solution demonstrates substantial progress explicitly
mentioned in the grading guidelines. Initial observations, reformulating the problem
without making substantive headway, or proving partial results not mentioned in the
grading guidelines are generally **not** eligible for this score.

**0 Points (Incorrect):** The solution doesn't make substantial progress that is a
key step in the full solution or is fundamentally flawed. All partial progress
without key results or lacking rigor also fall in this category.


### Input Data and Interpretation
You are provided with the following:

**Problem Statement:** The IMO problem.

**Ground Truth Solution:** A reference solution. Assume this solution is correct.
It demonstrates one valid approach.

**Specific Grading Guidelines:** Criteria for awarding credit for this specific problem.
These guidelines take precedence over the General Scoring Rubric, especially for
partial credit.

**Proposed Solution:** The student submission.


### Evaluation Process
You must follow this structured process:

**Analyze References:** Meticulously read and understand the problem and
Ground Truth Solution check the Specific Grading Guidelines. Identify the key
steps for a complete solution and the criteria for partial credit.

**Step-by-Step Verification:** Verify the logical validity and rigor of every step.
Identify all flaws, gaps, assumptions, and errors. **Make sure you fully understand
every piece of logic behind each step of the proposed solution, you must be careful
for solutions that 'pretend' to be correct.**

**Assess Progress:** Determine the extent of non-trivial progress made.

**Score Determination:** Compare the findings against the Specific Grading
Guidelines and the General Rubric to determine the final score.


### Output Requirements
You must provide your final score in the format `<points>N</points>`. Ensure
the `<points>` block is used **only once**, as your answer will be parsed
based on the first `<points> </points>` block that appears in your whole response.

----------------------------------------------------------

**PROBLEM STATEMENT**
{problem}

**GROUND-TRUTH SOLUTION**
{sample_solution}

**SPECIFIC GRADING GUIDELINES**
{grading_scheme}

**PROPOSED SOLUTION**
{solution}

----------------------------------------------------------

Present your detailed thought process and formal justification based
on the scoring rubric and grading guidelines, and finally present your
final score in the format below.

- `<points>N</points>`

\end{verbatim}
}\par\noindent\rule{\linewidth}{0.8pt}
\bigskip

\bigskip
\noindent\rule{\linewidth}{0.8pt}\par\nopagebreak
\noindent\textbf{ProofGrader rubric grading prompt template (\emph{Full} setting).}\par\nopagebreak
\noindent\rule{\linewidth}{0.4pt}\par\nopagebreak
{\footnotesize
\begin{verbatim}
You are an **expert math proof grader**. You are judging the correctness of
an LLM-generated proof for a math problem.

### Input

Your input will consist of:
* **Problem Statement**: A mathematical problem that the proof is attempting to solve.

* **Reference Solution**: A correct solution or proof provided for reference. This is
**not necessarily the only valid solution**. If the problem requires a final numeric
or algebraic answer, this section contains the correct answer, which should be the
only accepted final answer (though alternative reasoning paths are valid).

* **Marking Scheme**: A problem-specific grading rubric (0-7 scale) with checkpoints,
zero-credit items, and deductions. You must
follow this scheme when assigning points.

* **Proof Solution**: The proof that you need to evaluate. This proof may contain
errors, omissions, or unclear steps. The proof
was generated by another language model.

### Task

Analyze the proof carefully. 

* Compare the proof against the reference solution and the marking scheme.
* Award points according to the marking scheme’s checkpoints, zero-credit items,
and deductions.
* Identify logical errors, incorrect steps, or unclear reasoning.
* Give a score between 0 and 7 with a brief overall assessment.
* Show clearly how the score was derived:
* Which checkpoints were earned (with awarded points).
* Any zero-credit items or deductions applied.
* How the subtotal leads to the final score (0-7).

### Output Format

Respond with **only** well-formed XML using the structure below. Do not include
any extra text or Markdown.

**Requirements:**
- ‘<score>‘ must be an integer in [0, 7].
- ‘<assessment>‘ must be a **detailed analysis** that explains
your reasoning step-by-step and provides a clear **rationale for
the score**. Reference specific claims/lines if present.
- ‘<errors>‘ must be a list of specific issues (empty if score = 7).
Example output:
<score>0</score>
<assessment>The proof shows a good understanding of the main idea
but has some unclear reasoning and minor mistakes...</assessment>
<errors>
1. specific error 1,
2. specific error 2,
...
</errors>
----------------------------------------------------------
**Problem Statement**
{problem}

**Reference Solution**
{sample_solution}

**Marking Scheme**
{grading_scheme}

**Proof Solution**
{solution}
\end{verbatim}
}\par\noindent\rule{\linewidth}{0.8pt}
\bigskip

\bigskip
\noindent\rule{\linewidth}{0.8pt}\par\nopagebreak
\noindent\textbf{\synthmethod{}error-injection prompt template (one-shot).}\par\nopagebreak
\noindent\rule{\linewidth}{0.4pt}\par\nopagebreak
{\scriptsize
\begin{verbatim}
You are an expert mathematics professor and synthetic data
generator. Your task is to generate a realistic, natural-sounding
student solution to a given advanced mathematics problem.

You will be provided with:
1. The Problem
2. The Markscheme (Grading Scheme)
3. A Correct Reference Solution
4. A Score Profile specifying exactly how many points the student
   earns on each rubric criterion

Your objective is to write a complete candidate proof that earns
**EXACTLY** the points specified in the score profile on each
criterion.

Rules:

1. **Full-marks criteria**: The relevant proof sections must be
   mathematically sound and rigorous. You may freely copy or closely
   paraphrase corresponding parts of the reference solution --
   students who understand a technique often reproduce it in very
   similar language.

2. **Deducted criteria -- error calibration**: For criteria where
   points are LOST, introduce a *believable* student error
   proportional to the size of the deduction:
   - **Minor deduction** (e.g., losing 1 of 3 points): a small gap
     -- an off-by-one index, an edge case left unchecked, a minor
     algebraic slip, a missing quantifier, or an unjustified but
     plausible assertion.
   - **Moderate deduction** (e.g., losing 2 of 3 points): a
     significant logical gap -- wrong case analysis, an unverified
     hypothesis used as if proven, a crucial intermediate step
     hand-waved or skipped.
   - **Full deduction** (losing all points on a criterion): a
     fundamental omission or conceptual error -- the entire
     argument for that part is missing, a theorem is invoked with
     unmet hypotheses, or the approach simply does not address
     the criterion.

3. **Error believability**: Every injected error must be one a
   competent but imperfect student would plausibly make. Ask
   yourself: "What would a student who understood the main ideas
   but rushed or overlooked something actually write?" Avoid random
   deletions or nonsensical mistakes.

4. **Coherence**: The proof must read as a single, unified student
   attempt -- not a patchwork of correct and incorrect pieces
   stitched together. Correct and incorrect sections must flow
   naturally into each other with consistent transitions, notation,
   and writing style throughout.

5. **Copying policy**: You may copy or closely paraphrase the
   reference solution, especially for full-marks sections. For
   deducted sections, base your writing on the reference's
   structure but introduce the targeted errors organically. Do NOT
   egregiously copy-paste the entire reference and merely delete
   lines to simulate errors.

---

### Worked Example

Below is a complete example demonstrating the expected
input-to-output mapping.

<example_problem>
Let $k$ and $d$ be positive integers. Prove that there exists a
positive integer $N$ such that for every odd integer $n>N$, the
digits in the base-$2n$ representation of $n^k$ are all greater
than $d$.
</example_problem>

<example_markscheme>
[
    {
        "desc": "Uses the base-2n representation of $n^k$ to find
                 a closed form expression of each digit
                 $a_i=\left\lfloor \frac{n^k}{(2n)^i}\right\rfloor
                 mod\ (2n)$",
        "points": 1,
        "title": "Initial closed form expression for each digit"
    },
    {
        "desc": "Transforms the digit expression into a easily
                 boundable one by interchanging the floor and
                 modulo operations.",
        "points": 3,
        "title": "Interchanging the floor and mod"
    },
    {
        "desc": "Correctly bounds the aforementioned expression,
                 usually by showing the numerator is divisible by
                 n, so that
                 $a_i>=\left\lfloor \frac{n}{2^i}\right\rfloor$.",
        "points": 2,
        "title": "Bounding the expression"
    },
    {
        "desc": "Chooses a provably sufficiently large $N$. The
                 point is not awarded if it has not proven that
                 n^k has at most k digits.",
        "points": 1,
        "title": "Finishing"
    }
]
</example_markscheme>

<example_correct_solution>
We claim one such $N$ is $\boxed{N=2^{k-1}(d+1)}$. Begin by
letting $a_i$ be the digit at the $i$th position from the right
(0-indexed) of $n^k$ in base $2n$; i.e., $a_0$ is the units digit.
Obviously, $a_i=0$ for all $i\ge k$, since $n^k<(2n)^k$. We WTS
$a_i>d$ for each $i=0,1,\dots,k-1$, and observe the following:
\[ a_i=\left\lfloor\frac{n^k}{(2n)^i}\right\rfloor\text{ mod }2n
     =\left\lfloor\frac{n^{k-i}}{2^i}\right\rfloor\text{ mod }2n
     =\left\lfloor\frac{n^{k-i}\text{ mod }2^{i+1}n}{2^i}\right
     \rfloor. \]
Note that the numerator is a positive multiple of $n$, since 1)
$k-i>0\implies n\mid\gcd(n^{k-i},2^{i+1}n)$ and 2) ($n$ is odd)
$2^{i+1}n\nmid n^{k-i}$. Thus,
\[ a_i=\left\lfloor\frac{n^{k-i}\text{ mod }2^{i+1}n}{2^i}\right
     \rfloor>d \quad\iff\quad
   n^{k-i}\text{ mod }2^{i+1}n\ge n\ge 2^i(d+1), \]
which is true since $n>N=2^{k-1}(d+1)$ and $k-1\ge i$. $\square$
</example_correct_solution>

<example_score_profile>
1. "Initial closed form expression for each digit" (1 pts)
   -> Awarded 1/1  [FULL MARKS]
   Uses the base-2n representation of $n^k$ to find a closed form
   expression of each digit
   $a_i=\left\lfloor \frac{n^k}{(2n)^i}\right\rfloor mod\ (2n)$
2. "Interchanging the floor and mod" (3 pts)
   -> Awarded 3/3  [FULL MARKS]
   Transforms the digit expression into a easily boundable one by
   interchanging the floor and modulo operations.
3. "Bounding the expression" (2 pts) -> Awarded 2/2  [FULL MARKS]
   Correctly bounds the aforementioned expression, usually by
   showing the numerator is divisible by n, so that
   $a_i>=\left\lfloor \frac{n}{2^i}\right\rfloor$.
4. "Finishing" (1 pts) -> Awarded 0/1  [LOSE 1/1 points]
   Chooses a provably sufficiently large $N$. The point is not
   awarded if it has not proven that n^k has at most k digits.

Total score: 6/7
</example_score_profile>

**Why this output works**: Criteria 1-3 earn full marks, so the
student closely follows the reference. Criterion 4 loses its point
because the student omits the digit-count bound "$a_i=0$ for all
$i\ge k$ since $n^k<(2n)^k$" -- a plausible oversight by a student
who focused on the algebraic core and forgot to justify why only
$k$ digits matter:

# Candidate Solution
We claim one such $N$ is $\boxed{N=2^{k-1}(d+1)}$. Begin by letting
$a_i$ be the digit at the $i$th position from the right (0-indexed)
of $n^k$ in base $2n$; i.e., $a_0$ is the units digit. We WTS
$a_i>d$ for each $i=0,1,\dots,k-1$, and observe the following:
$$ a_i=\left\lfloor\frac{n^k}{(2n)^i}\right\rfloor\text{ mod }2n
      =\left\lfloor\frac{n^{k-i}}{2^i}\right\rfloor\text{ mod }2n
      =\left\lfloor\frac{n^{k-i}\text{ mod }2^{i+1}n}{2^i}\right
      \rfloor. $$
Note that the numerator is a positive multiple of $n$, since 1)
$k-i>0\implies n\mid\gcd(n^{k-i},2^{i+1}n)$ and 2) ($n$ is odd)
$2^{i+1}n\nmid n^{k-i}$. Thus,
$$ a_i=\left\lfloor\frac{n^{k-i}\text{ mod }2^{i+1}n}{2^i}\right
      \rfloor>d \quad\iff\quad
   n^{k-i}\text{ mod }2^{i+1}n\ge n\ge 2^i(d+1), $$
which is true since $n>N=2^{k-1}(d+1)$ and $k-1\ge i$.
\hfill $\square$

# Grader's Notes
- **Finishing (0/1)**: The student never shows that $n^k$ has at
  most $k$ digits in base $2n$ (i.e., omits "$a_i=0$ for all
  $i\ge k$ since $n^k<(2n)^k$"). Without this, the claim that it
  suffices to check only $i=0,\dots,k-1$ is unjustified, so $N$ is
  not proven sufficiently large.

---

Now generate a candidate student solution for the following:

<problem>
{problem}
</problem>

<markscheme>
{grading_scheme}
</markscheme>

<correct_solution>
{correct_solution}
</correct_solution>

<score_profile>
{score_profile}
</score_profile>

Write the candidate proof below. Use standard LaTeX formatting for
all mathematics. After the proof, provide a brief "Grader's Notes"
section: for each deducted criterion, state the criterion name, the
score awarded, the specific error introduced, and why it warrants
that deduction.

# Candidate Solution
[The student's complete proof]

# Grader's Notes
[For each deducted criterion: criterion name (awarded/max), the
specific error, and why it matches the deduction]
\end{verbatim}
}\par\noindent\rule{\linewidth}{0.8pt}
\bigskip

\bigskip
\noindent\rule{\linewidth}{0.8pt}\par\nopagebreak
\noindent\textbf{\synthmethod{}critique prompt template.}\par\nopagebreak
\noindent\rule{\linewidth}{0.4pt}\par\nopagebreak
{\footnotesize
\begin{verbatim}
You are a rigorous mathematics examiner auditing a synthetic
student solution. A generator produced a candidate solution
designed to earn a specific TARGET score profile. Your task is
to independently grade the candidate and verify that a real
grader, using the markscheme below, would award exactly the
target scores.

You have access to the correct reference solution as the ground
truth for mathematical correctness.

<problem>
{problem}
</problem>

<markscheme>
{grading_scheme}
</markscheme>

<correct_solution>
{correct_solution}
</correct_solution>

<target_score_profile>
{score_profile}
</target_score_profile>

<candidate_solution>
{candidate_solution}
</candidate_solution>

Instructions:

1. **Grade each criterion independently.** For each rubric item,
   read the candidate solution and determine the ACTUAL points it
   would earn based solely on the markscheme and mathematical
   content. Use the reference solution as the correctness benchmark.

2. **Compare actual vs. target.** Set "aligned" to true only if
   actual_score == target_score for that criterion.

3. **For misaligned criteria, provide specific and actionable
   revision guidance:**
   - If the candidate earns **MORE** than the target (error too
     mild or absent): state what is currently correct that should
     be broken, and suggest a specific, *believable* student error
     -- something a competent student could plausibly make. Do not
     suggest random deletions.
   - If the candidate earns **LESS** than the target (error too
     severe): state what is currently wrong and how to partially
     or fully fix it so the student earns the target partial credit.

4. **Error calibration reference** (use this to judge whether error
   severity matches the deduction):
   - Minor deduction (e.g., 1 of 3 points lost): small gap --
     off-by-one, unchecked edge case, missing quantifier,
     unjustified assertion.
   - Moderate deduction (e.g., 2 of 3 points lost): significant
     logical gap -- skipped step, unverified hypothesis, wrong
     case analysis.
   - Full deduction (all points on a criterion lost): fundamental
     omission -- entire argument missing, theorem invoked with
     unmet hypotheses, approach doesn't address the criterion.

5. **Coherence check.** Also assess whether the proof reads as a
   natural, unified student attempt. If it feels like a patchwork
   of correct and incorrect pieces, or if errors seem artificially
   inserted rather than organically arising, flag this in the
   revision guidance.

6. **Copying policy.** Close paraphrasing of the reference solution
   is acceptable for full-marks sections -- do NOT flag similarity
   to the reference as a problem. Students who master a technique
   often reproduce it in very similar language.

Output ONLY a JSON object (no markdown fences, no surrounding text)
in this exact format:

{
  "criteria": [
    {
      "criterion_index": 1,
      "title": "criterion title",
      "target_score": 0,
      "actual_score": 0,
      "aligned": true,
      "feedback": "What the candidate does for this criterion,
                   what the target requires, and (if misaligned)
                   a specific revision to make."
    }
  ],
  "overall_aligned": true,
  "revision_guidance": "High-level summary of all revisions needed,
                        including any coherence issues. Empty
                        string if fully aligned."
}
\end{verbatim}
}\par\noindent\rule{\linewidth}{0.8pt}
\bigskip

\bigskip
\noindent\rule{\linewidth}{0.8pt}\par\nopagebreak
\noindent\textbf{\synthmethod{}refinement prompt template.}\par\nopagebreak
\noindent\rule{\linewidth}{0.4pt}\par\nopagebreak
{\footnotesize
\begin{verbatim}
You are an expert mathematics professor revising a synthetic
student solution. A critique found that some rubric criteria do
NOT earn the intended number of points. Your task is to revise
the solution so each criterion earns EXACTLY the target score.

<problem>
{problem}
</problem>

<markscheme>
{grading_scheme}
</markscheme>

<correct_solution>
{correct_solution}
</correct_solution>

<target_score_profile>
{score_profile}
</target_score_profile>

<current_candidate>
{candidate_solution}
</current_candidate>

<critique>
{critique}
</critique>

Revision rules:

1. **Preserve aligned criteria.** For criteria already matching the
   target score, keep those proof sections as close to unchanged as
   possible. Do not inadvertently break what already works.

2. **Address each critique point specifically.** For each misaligned
   criterion, follow the critique's feedback:
   - If the candidate earns **MORE** than the target: introduce a
     larger or more fundamental error to bring the score down. The
     error must be *believable* -- something a competent but
     imperfect student would plausibly write.
   - If the candidate earns **LESS** than the target: soften or
     partially fix the error so the student earns the intended
     partial credit. Use the reference solution to restore the
     relevant argument.

3. **Error calibration:**
   - Minor deduction (e.g., 1 of 3 points lost): small gap --
     off-by-one, unchecked edge case, missing quantifier,
     unjustified assertion.
   - Moderate deduction (e.g., 2 of 3 points lost): significant
     logical gap -- skipped step, unverified hypothesis, wrong
     case analysis.
   - Full deduction (all points on a criterion lost): fundamental
     omission -- entire argument missing, theorem misapplied,
     approach doesn't address the criterion.

4. **Coherence.** The revised proof must read as a single, unified
   student attempt. After changing one section, re-read the
   surrounding paragraphs to ensure transitions, notation, and
   writing style remain consistent throughout.

5. **Copying policy.** You may freely copy or closely paraphrase
   the reference solution for full-marks sections -- students who
   master a technique often reproduce it in similar language. For
   deducted sections, base your revision on the reference's
   structure but weave in errors organically. Do NOT simply delete
   lines from the reference to simulate mistakes.

6. **Minimal edits.** Make the smallest changes that fix the
   misalignments. Rewriting large correct sections increases the
   risk of introducing unintended new errors or breaking aligned
   criteria.

Output format:

# Candidate Solution
[The revised proof]

# Grader's Notes
[For each deducted criterion: criterion name (awarded/max), the
specific error, and why it matches the deduction]
\end{verbatim}
}\par\noindent\rule{\linewidth}{0.8pt}
\bigskip

\clearpage

\newpage


\goodbreak\newpage\newpage

\section*{NeurIPS Paper Checklist}

\begin{enumerate}

\item {\bf Claims}
    \item[] Question: Do the main claims made in the abstract and introduction accurately reflect the paper's contributions and scope?
    \item[] Answer: \answerYes{} 
    \item[] Justification: Each claim is supported by empirical evaluation across a broad range of language models.
    \item[] Guidelines:
    \begin{itemize}
        \item The answer \answerNA{} means that the abstract and introduction do not include the claims made in the paper.
        \item The abstract and/or introduction should clearly state the claims made, including the contributions made in the paper and important assumptions and limitations. A \answerNo{} or \answerNA{} answer to this question will not be perceived well by the reviewers. 
        \item The claims made should match theoretical and experimental results, and reflect how much the results can be expected to generalize to other settings. 
        \item It is fine to include aspirational goals as motivation as long as it is clear that these goals are not attained by the paper. 
    \end{itemize}

\item {\bf Limitations}
    \item[] Question: Does the paper discuss the limitations of the work performed by the authors?
    \item[] Answer: \answerYes{} 
    \item[] Justification: We provide limitations and future work analysis of our work in the Appendix.
    \item[] Guidelines:
    \begin{itemize}
        \item The answer \answerNA{} means that the paper has no limitation while the answer \answerNo{} means that the paper has limitations, but those are not discussed in the paper. 
        \item The authors are encouraged to create a separate ``Limitations'' section in their paper.
        \item The paper should point out any strong assumptions and how robust the results are to violations of these assumptions (e.g., independence assumptions, noiseless settings, model well-specification, asymptotic approximations only holding locally). The authors should reflect on how these assumptions might be violated in practice and what the implications would be.
        \item The authors should reflect on the scope of the claims made, e.g., if the approach was only tested on a few datasets or with a few runs. In general, empirical results often depend on implicit assumptions, which should be articulated.
        \item The authors should reflect on the factors that influence the performance of the approach. For example, a facial recognition algorithm may perform poorly when image resolution is low or images are taken in low lighting. Or a speech-to-text system might not be used reliably to provide closed captions for online lectures because it fails to handle technical jargon.
        \item The authors should discuss the computational efficiency of the proposed algorithms and how they scale with dataset size.
        \item If applicable, the authors should discuss possible limitations of their approach to address problems of privacy and fairness.
        \item While the authors might fear that complete honesty about limitations might be used by reviewers as grounds for rejection, a worse outcome might be that reviewers discover limitations that aren't acknowledged in the paper. The authors should use their best judgment and recognize that individual actions in favor of transparency play an important role in developing norms that preserve the integrity of the community. Reviewers will be specifically instructed to not penalize honesty concerning limitations.
    \end{itemize}

\item {\bf Theory assumptions and proofs}
    \item[] Question: For each theoretical result, does the paper provide the full set of assumptions and a complete (and correct) proof?
    \item[] Answer: \answerNA{} 
    \item[] Justification: The paper does not include theoretical results.
    \item[] Guidelines:
    \begin{itemize}
        \item The answer \answerNA{} means that the paper does not include theoretical results. 
        \item All the theorems, formulas, and proofs in the paper should be numbered and cross-referenced.
        \item All assumptions should be clearly stated or referenced in the statement of any theorems.
        \item The proofs can either appear in the main paper or the supplemental material, but if they appear in the supplemental material, the authors are encouraged to provide a short proof sketch to provide intuition. 
        \item Inversely, any informal proof provided in the core of the paper should be complemented by formal proofs provided in appendix or supplemental material.
        \item Theorems and Lemmas that the proof relies upon should be properly referenced. 
    \end{itemize}

    \item {\bf Experimental result reproducibility}
    \item[] Question: Does the paper fully disclose all the information needed to reproduce the main experimental results of the paper to the extent that it affects the main claims and/or conclusions of the paper (regardless of whether the code and data are provided or not)?
    \item[] Answer: \answerYes{} 
    \item[] Justification: Yes, we discuss all parameters for verification, grading, dataset construction, and inference, and provide training model cards for each SFT experiment.
    \item[] Guidelines:
    \begin{itemize}
        \item The answer \answerNA{} means that the paper does not include experiments.
        \item If the paper includes experiments, a \answerNo{} answer to this question will not be perceived well by the reviewers: Making the paper reproducible is important, regardless of whether the code and data are provided or not.
        \item If the contribution is a dataset and\slash or model, the authors should describe the steps taken to make their results reproducible or verifiable. 
        \item Depending on the contribution, reproducibility can be accomplished in various ways. For example, if the contribution is a novel architecture, describing the architecture fully might suffice, or if the contribution is a specific model and empirical evaluation, it may be necessary to either make it possible for others to replicate the model with the same dataset, or provide access to the model. In general. releasing code and data is often one good way to accomplish this, but reproducibility can also be provided via detailed instructions for how to replicate the results, access to a hosted model (e.g., in the case of a large language model), releasing of a model checkpoint, or other means that are appropriate to the research performed.
        \item While NeurIPS does not require releasing code, the conference does require all submissions to provide some reasonable avenue for reproducibility, which may depend on the nature of the contribution. For example
        \begin{enumerate}
            \item If the contribution is primarily a new algorithm, the paper should make it clear how to reproduce that algorithm.
            \item If the contribution is primarily a new model architecture, the paper should describe the architecture clearly and fully.
            \item If the contribution is a new model (e.g., a large language model), then there should either be a way to access this model for reproducing the results or a way to reproduce the model (e.g., with an open-source dataset or instructions for how to construct the dataset).
            \item We recognize that reproducibility may be tricky in some cases, in which case authors are welcome to describe the particular way they provide for reproducibility. In the case of closed-source models, it may be that access to the model is limited in some way (e.g., to registered users), but it should be possible for other researchers to have some path to reproducing or verifying the results.
        \end{enumerate}
    \end{itemize}

\item {\bf Open access to data and code}
    \item[] Question: Does the paper provide open access to the data and code, with sufficient instructions to faithfully reproduce the main experimental results, as described in supplemental material?
    \item[] Answer: \answerYes{} 
    \item[] Justification: The codebase is attached with the submission.
    \item[] Guidelines:
    \begin{itemize}
        \item The answer \answerNA{} means that paper does not include experiments requiring code.
        \item Please see the NeurIPS code and data submission guidelines (\url{https://neurips.cc/public/guides/CodeSubmissionPolicy}) for more details.
        \item While we encourage the release of code and data, we understand that this might not be possible, so \answerNo{} is an acceptable answer. Papers cannot be rejected simply for not including code, unless this is central to the contribution (e.g., for a new open-source benchmark).
        \item The instructions should contain the exact command and environment needed to run to reproduce the results. See the NeurIPS code and data submission guidelines (\url{https://neurips.cc/public/guides/CodeSubmissionPolicy}) for more details.
        \item The authors should provide instructions on data access and preparation, including how to access the raw data, preprocessed data, intermediate data, and generated data, etc.
        \item The authors should provide scripts to reproduce all experimental results for the new proposed method and baselines. If only a subset of experiments are reproducible, they should state which ones are omitted from the script and why.
        \item At submission time, to preserve anonymity, the authors should release anonymized versions (if applicable).
        \item Providing as much information as possible in supplemental material (appended to the paper) is recommended, but including URLs to data and code is permitted.
    \end{itemize}

\item {\bf Experimental setting/details}
    \item[] Question: Does the paper specify all the training and test details (e.g., data splits, hyperparameters, how they were chosen, type of optimizer) necessary to understand the results?
    \item[] Answer: \answerYes{} 
    \item[] Justification: Yes, we provide both Datasheet and model cards for evaluation, data collection and training.
    \item[] Guidelines:
    \begin{itemize}
        \item The answer \answerNA{} means that the paper does not include experiments.
        \item The experimental setting should be presented in the core of the paper to a level of detail that is necessary to appreciate the results and make sense of them.
        \item The full details can be provided either with the code, in appendix, or as supplemental material.
    \end{itemize}

\item {\bf Experiment statistical significance}
    \item[] Question: Does the paper report error bars suitably and correctly defined or other appropriate information about the statistical significance of the experiments?
    \item[] Answer: \answerYes{} 
    \item[] Justification: We report mean and variance of individual grading with 16 independent runs. Error bars are reported for each conducted experiment in the Appendix of the work. 
    \item[] Guidelines:
    \begin{itemize}
        \item The answer \answerNA{} means that the paper does not include experiments.
        \item The authors should answer \answerYes{} if the results are accompanied by error bars, confidence intervals, or statistical significance tests, at least for the experiments that support the main claims of the paper.
        \item The factors of variability that the error bars are capturing should be clearly stated (for example, train/test split, initialization, random drawing of some parameter, or overall run with given experimental conditions).
        \item The method for calculating the error bars should be explained (closed form formula, call to a library function, bootstrap, etc.)
        \item The assumptions made should be given (e.g., Normally distributed errors).
        \item It should be clear whether the error bar is the standard deviation or the standard error of the mean.
        \item It is OK to report 1-sigma error bars, but one should state it. The authors should preferably report a 2-sigma error bar than state that they have a 96\% CI, if the hypothesis of Normality of errors is not verified.
        \item For asymmetric distributions, the authors should be careful not to show in tables or figures symmetric error bars that would yield results that are out of range (e.g., negative error rates).
        \item If error bars are reported in tables or plots, the authors should explain in the text how they were calculated and reference the corresponding figures or tables in the text.
    \end{itemize}

\item {\bf Experiments compute resources}
    \item[] Question: For each experiment, does the paper provide sufficient information on the computer resources (type of compute workers, memory, time of execution) needed to reproduce the experiments?
    \item[] Answer: \answerYes{} 
    \item[] Justification: We disclose every model accessed via API. 
    \item[] Guidelines:
    \begin{itemize}
        \item The answer \answerNA{} means that the paper does not include experiments.
        \item The paper should indicate the type of compute workers CPU or GPU, internal cluster, or cloud provider, including relevant memory and storage.
        \item The paper should provide the amount of compute required for each of the individual experimental runs as well as estimate the total compute. 
        \item The paper should disclose whether the full research project required more compute than the experiments reported in the paper (e.g., preliminary or failed experiments that didn't make it into the paper). 
    \end{itemize}
    
\item {\bf Code of ethics}
    \item[] Question: Does the research conducted in the paper conform, in every respect, with the NeurIPS Code of Ethics \url{https://neurips.cc/public/EthicsGuidelines}?
    \item[] Answer: \answerYes{} 
    \item[] Justification: All sourced data is in accordance with ethical standards.
    \item[] Guidelines:
    \begin{itemize}
        \item The answer \answerNA{} means that the authors have not reviewed the NeurIPS Code of Ethics.
        \item If the authors answer \answerNo, they should explain the special circumstances that require a deviation from the Code of Ethics.
        \item The authors should make sure to preserve anonymity (e.g., if there is a special consideration due to laws or regulations in their jurisdiction).
    \end{itemize}

\item {\bf Broader impacts}
    \item[] Question: Does the paper discuss both potential positive societal impacts and negative societal impacts of the work performed?
    \item[] Answer: \answerNA{} 
    \item[] Justification: Our work discusses fairness in mathematical verification. We do not anticipate negative societal impacts.
    \item[] Guidelines:
    \begin{itemize}
        \item The answer \answerNA{} means that there is no societal impact of the work performed.
        \item If the authors answer \answerNA{} or \answerNo, they should explain why their work has no societal impact or why the paper does not address societal impact.
        \item Examples of negative societal impacts include potential malicious or unintended uses (e.g., disinformation, generating fake profiles, surveillance), fairness considerations (e.g., deployment of technologies that could make decisions that unfairly impact specific groups), privacy considerations, and security considerations.
        \item The conference expects that many papers will be foundational research and not tied to particular applications, let alone deployments. However, if there is a direct path to any negative applications, the authors should point it out. For example, it is legitimate to point out that an improvement in the quality of generative models could be used to generate Deepfakes for disinformation. On the other hand, it is not needed to point out that a generic algorithm for optimizing neural networks could enable people to train models that generate Deepfakes faster.
        \item The authors should consider possible harms that could arise when the technology is being used as intended and functioning correctly, harms that could arise when the technology is being used as intended but gives incorrect results, and harms following from (intentional or unintentional) misuse of the technology.
        \item If there are negative societal impacts, the authors could also discuss possible mitigation strategies (e.g., gated release of models, providing defenses in addition to attacks, mechanisms for monitoring misuse, mechanisms to monitor how a system learns from feedback over time, improving the efficiency and accessibility of ML).
    \end{itemize}
    
\item {\bf Safeguards}
    \item[] Question: Does the paper describe safeguards that have been put in place for responsible release of data or models that have a high risk for misuse (e.g., pre-trained language models, image generators, or scraped datasets)?
    \item[] Answer: \answerNA{} 
    \item[] Justification: Every benchmark comes with a root source of the problem to direct the readers to original releases. Other than that we do not anticipate misuse risks.
    \item[] Guidelines:
    \begin{itemize}
        \item The answer \answerNA{} means that the paper poses no such risks.
        \item Released models that have a high risk for misuse or dual-use should be released with necessary safeguards to allow for controlled use of the model, for example by requiring that users adhere to usage guidelines or restrictions to access the model or implementing safety filters. 
        \item Datasets that have been scraped from the Internet could pose safety risks. The authors should describe how they avoided releasing unsafe images.
        \item We recognize that providing effective safeguards is challenging, and many papers do not require this, but we encourage authors to take this into account and make a best faith effort.
    \end{itemize}

\item {\bf Licenses for existing assets}
    \item[] Question: Are the creators or original owners of assets (e.g., code, data, models), used in the paper, properly credited and are the license and terms of use explicitly mentioned and properly respected?
    \item[] Answer: \answerYes{} 
    \item[] Justification: We cite and explicitly mention every sourced data and templates in the main text and Appendix.
    \item[] Guidelines:
    \begin{itemize}
        \item The answer \answerNA{} means that the paper does not use existing assets.
        \item The authors should cite the original paper that produced the code package or dataset.
        \item The authors should state which version of the asset is used and, if possible, include a URL.
        \item The name of the license (e.g., CC-BY 4.0) should be included for each asset.
        \item For scraped data from a particular source (e.g., website), the copyright and terms of service of that source should be provided.
        \item If assets are released, the license, copyright information, and terms of use in the package should be provided. For popular datasets, \url{paperswithcode.com/datasets} has curated licenses for some datasets. Their licensing guide can help determine the license of a dataset.
        \item For existing datasets that are re-packaged, both the original license and the license of the derived asset (if it has changed) should be provided.
        \item If this information is not available online, the authors are encouraged to reach out to the asset's creators.
    \end{itemize}

\item {\bf New assets}
    \item[] Question: Are new assets introduced in the paper well documented and is the documentation provided alongside the assets?
    \item[] Answer: \answerYes{} 
    \item[] Justification: Yes, every datapoint comes with a marking scheme, candidate solutions and corresponding justifications.
    \item[] Guidelines:
    \begin{itemize}
        \item The answer \answerNA{} means that the paper does not release new assets.
        \item Researchers should communicate the details of the dataset\slash code\slash model as part of their submissions via structured templates. This includes details about training, license, limitations, etc. 
        \item The paper should discuss whether and how consent was obtained from people whose asset is used.
        \item At submission time, remember to anonymize your assets (if applicable). You can either create an anonymized URL or include an anonymized zip file.
    \end{itemize}

\item {\bf Crowdsourcing and research with human subjects}
    \item[] Question: For crowdsourcing experiments and research with human subjects, does the paper include the full text of instructions given to participants and screenshots, if applicable, as well as details about compensation (if any)? 
    \item[] Answer: \answerNA{} 
    \item[] Justification: The work does not use crowd sourcing.
    \item[] Guidelines:
    \begin{itemize}
        \item The answer \answerNA{} means that the paper does not involve crowdsourcing nor research with human subjects.
        \item Including this information in the supplemental material is fine, but if the main contribution of the paper involves human subjects, then as much detail as possible should be included in the main paper. 
        \item According to the NeurIPS Code of Ethics, workers involved in data collection, curation, or other labor should be paid at least the minimum wage in the country of the data collector. 
    \end{itemize}

\item {\bf Institutional review board (IRB) approvals or equivalent for research with human subjects}
    \item[] Question: Does the paper describe potential risks incurred by study participants, whether such risks were disclosed to the subjects, and whether Institutional Review Board (IRB) approvals (or an equivalent approval/review based on the requirements of your country or institution) were obtained?
    \item[] Answer: \answerNA{} 
    \item[] Justification: The work does not use crowd sourcing.
    \item[] Guidelines:
    \begin{itemize}
        \item The answer \answerNA{} means that the paper does not involve crowdsourcing nor research with human subjects.
        \item Depending on the country in which research is conducted, IRB approval (or equivalent) may be required for any human subjects research. If you obtained IRB approval, you should clearly state this in the paper. 
        \item We recognize that the procedures for this may vary significantly between institutions and locations, and we expect authors to adhere to the NeurIPS Code of Ethics and the guidelines for their institution. 
        \item For initial submissions, do not include any information that would break anonymity (if applicable), such as the institution conducting the review.
    \end{itemize}

\item {\bf Declaration of LLM usage}
    \item[] Question: Does the paper describe the usage of LLMs if it is an important, original, or non-standard component of the core methods in this research? Note that if the LLM is used only for writing, editing, or formatting purposes and does \emph{not} impact the core methodology, scientific rigor, or originality of the research, declaration is not required.
    \item[] Answer: \answerYes{} 
    \item[] Justification: We explicitly describe the LLM uses in grading mathemtical problems. Every experiment comes with a detailed explanation of base models, templates and other relevant parameters.
    \item[] Guidelines:
    \begin{itemize}
        \item The answer \answerNA{} means that the core method development in this research does not involve LLMs as any important, original, or non-standard components.
        \item Please refer to our LLM policy in the NeurIPS handbook for what should or should not be described.
    \end{itemize}

\end{enumerate}

\end{document}